\documentclass{article} 
\usepackage[svgnames,dvipsnames,table]{xcolor}         

\usepackage{amsmath,amsfonts,bm}

\def\eqref#1{equation~\ref{#1}}

\def\1{\bm{1}}

\DeclareMathAlphabet{\mathsfit}{\encodingdefault}{\sfdefault}{m}{sl}
\SetMathAlphabet{\mathsfit}{bold}{\encodingdefault}{\sfdefault}{bx}{n}

\newcommand{\E}{\mathbb{E}}

\newcommand{\Var}{\mathrm{Var}}

\usepackage{booktabs}       
\usepackage[hyphens]{url}
\usepackage[colorlinks=true, citecolor=Navy, breaklinks=true]{hyperref}       
\usepackage{nicefrac}       
\usepackage{caption}
\usepackage{subcaption}
\usepackage{graphicx}
\usepackage{wrapfig}
\usepackage[most]{tcolorbox}
\usepackage{multirow}
\usepackage[capitalize,noabbrev]{cleveref}
\usepackage[inline]{enumitem}
\usepackage{mathtools}
\usepackage{amsthm}
\usepackage{pifont}
\usepackage{cereb}

\usepackage[textsize=tiny]{todonotes}

\crefname{figure}{Fig.}{Figs.}
\Crefname{figure}{Fig.}{Figs.}
\crefname{equation}{Eq.}{Eqs.}
\Crefname{equation}{Eq.}{Eqs.}
\crefname{section}{Sec.}{Secs.}
\Crefname{section}{Sec.}{Secs.}
\crefname{subsection}{Sec.}{Secs.}
\Crefname{subsection}{Sec.}{Secs.}
\crefname{subsubsection}{Sec.}{Secs.}
\Crefname{subsubsection}{Sec.}{Secs.}

\theoremstyle{definition}

\definecolor{taborange}{RGB}{235,115,12}
\definecolor{tabblue}{RGB}{31,119,180}
\definecolor{tabgreen}{RGB}{44,160,44}
\definecolor{tabred}{RGB}{214,39,40}

\usepackage{tikz}
\usepackage{pgfplots}
\usetikzlibrary{arrows.meta,shapes.geometric}

\definecolor{cborange}{RGB}{255,127,0} 
\definecolor{cbblue}{RGB}{31,119,180}  

\usepackage{enumitem}
\setlist[itemize]{left=0.5em}
\setlist[enumerate]{left=0.5em}

\newcommand{\tpp}{\mathrm{TPP}}

\newcommand{\leftfig}{\text{\emph{left}}}
\newcommand{\middlefig}{\text{\emph{middle}}}
\newcommand{\rightfig}{\text{\emph{right}}}
\newcommand{\Leftfig}{\text{\emph{Left}}}
\newcommand{\Middlefig}{\text{\emph{Middle}}}
\newcommand{\Rightfig}{\text{\emph{Right}}}

\newcommand{\tenx}{\mbox{10}\times}
\newcommand{\dtoz}{\mbox{D2Z}}
\newcommand{\mup}{\mbox{$\mu$P}}
\newcommand{\titer}{\tau_\mathrm{iter}}
\newcommand{\tepochwang}{\tau_\mathrm{epoch}}
\newcommand{\tepoch}{\tau}
\newcommand{\tema}{\tepoch}  
\newcommand{\constant}{\mbox{\emph{Constant}}}
\newcommand{\linear}{\mbox{\emph{Linear}}}

\newcommand{\dmodel}{d_{\mathrm{model}}}
\newcommand{\dproxy}{d_{\mathrm{proxy}}}
\newcommand{\nlayers}{n_{\mathrm{layers}}}
\newcommand{\dffn}{d_{\mathrm{ffn}}}
\newcommand{\dhead}{d_{\mathrm{head}}}

\newcommand{\hateta}{\tilde{\eta}}
\newcommand{\trainfrac}{\hat{t}}
\newcommand{\cbs}{B_{\mathrm{crit}}}
\newcommand{\bcrit}{\cbs}  
\newcommand{\bopt}{B_{\mathrm{opt}}}
\newcommand{\lhat}{\hat{L}}
\newcommand{\nopt}{N_{\mathrm{opt}}}
\newcommand{\dopt}{D_{\mathrm{opt}}}
\newcommand{\copt}{C_{\mathrm{opt}}}

\newcommand{\rstar}{r}
\newcommand{\nconst}{A}
\newcommand{\dconst}{B}
\DeclareMathOperator{\powerlaw}{PowerLaw}

\newcommand{\bconst}{b_{\text{const}}}
\newcommand{\qconst}{q_{\text{const}}}
\newcommand{\bexp}{b_{\text{exp}}}
\newcommand{\qexp}{q_{\text{exp}}}
\newcommand{\ellhat}{\hat{\ell}}

\newcommand{\maxlrdetail}{1.62\mbox{e-}02}
\newcommand{\tstar}{T^{\star}}

\tcbset{textmarker/.style={%
  enhanced,
  parbox=false,boxrule=0mm,boxsep=0mm,arc=0mm,
  outer arc=0mm,left=6mm,right=3mm,top=4pt,bottom=3pt,
  toptitle=1mm,bottomtitle=1mm,oversize,
  left skip=0mm,
  right skip=8mm
}}

\newcounter{fcounter}
\crefname{fcounter}{Finding}{Findings}

\newcounter{kcounter}
\newcommand\takeaway[1]{
        \refstepcounter{kcounter}\vspace{2pt}
        \begin{tcolorbox}[colback=green!10!white,colframe=green!80!black,boxsep=1pt,left=2pt,right=2pt,top=1pt,bottom=1pt]\noindent{\textbf{\sffamily Key takeaway \arabic{kcounter}}: \sffamily #1}
        \end{tcolorbox}\vspace{0pt}
}
\crefname{kcounter}{Takeaway}{Takeaways}

\newtcolorbox{hypothesisBox}{textmarker,
    borderline west={6pt}{0pt}{blue},
    colback=blue!10!white}
\newcounter{hcounter}
\crefname{hcounter}{Hypothesis}{Hypotheses}

\renewcommand{\cite}[1]{\PackageError{MyPackage}{Do not use \string\cite\space with natbib. Use \string\citet\space or \string\citep}{See the natbib package documentation for explanation.}}

\makeatletter
\AtBeginDocument{%
  \@ifpackageloaded{cleveref}{%

    \providecommand\cref@appendix@setup{%
      \crefname{appendix}{Appendix}{Appendices}%
      \Crefname{appendix}{Appendix}{Appendices}%
    }%

    \AddToHook{cmd/appendix/before}{%
      \cref@appendix@setup
      \@ifundefined{chapter}{%
        \crefalias{section}{appendix}%
      }{%
        \crefalias{chapter}{appendix}%
      }%
      \crefalias{subsection}{appendix}%
      \crefalias{subsubsection}{appendix}%
    }%

  }{}
}
\makeatother

\title{Scaling with Collapse: Efficient and Predictable Training of LLM Families}

\author{
    \hspace{-4pt} Shane Bergsma  \ \ \ Bin Claire Zhang \ \ \ Nolan Dey \ \ \ Shaheer Muhammad \ \ \ Gurpreet Gosal \\
    \textbf{Joel Hestness} \ \ \ \\
    \normalfont Cerebras Systems \\
    \texttt{\{shane.bergsma,joel\}@cerebras.net}
}

\begin{document}
\maketitle

\begin{abstract}
Effective LLM training depends on predictable scaling of key
quantities---such as final loss and optimal hyperparameters---with
model and dataset size.
\citet{qiu2025scaling} recently showed
that this predictability can extend beyond scalars: whole training
loss curves can \emph{collapse} onto a universal trajectory after a
simple normalization.  What remains unclear is whether this phenomenon
persists for LLM families trained under \emph{practical scaling
recipes}, where width, depth, learning rate, batch size, and weight
decay are scaled jointly.  We show that it does: loss curves collapse
across scales precisely when optimization hyperparameters are set
optimally for the given data budget, in accordance with recent
empirical scaling laws.  Collapse therefore emerges as a signature of
compute-efficient training.  We demonstrate two applications at scale:
(1)~deviation-from-collapse provides a sensitive, early diagnostic of
training pathologies, and (2)~predictability of collapsed curves
enables early stopping in large-scale hyperparameter tuning.  Finally,
we train a competitive LLM family, \emph{Celerity}, using these
insights, establishing collapse as an effective tool for developing
efficient LLMs.
All experiments were run on Cerebras CS-3 systems.

\end{abstract}

\begin{figure}[ht]
  \centering
  \begin{minipage}{0.33\textwidth}
    \includegraphics[trim={0.32cm 0.42cm 0.264cm 0.33cm}, clip, width=\linewidth]{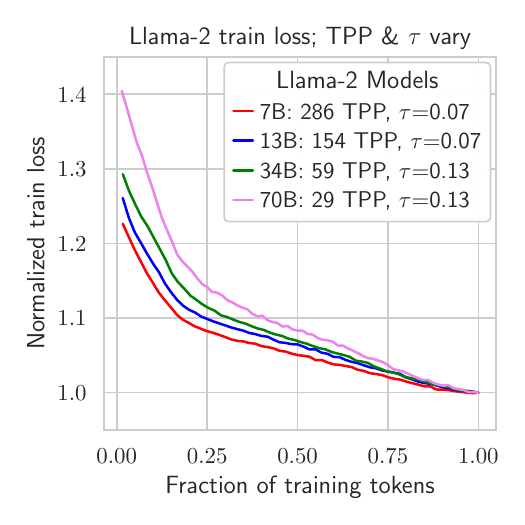}
  \end{minipage}\hfill
  \begin{minipage}{0.33\textwidth}
    \includegraphics[trim={0.32cm 0.42cm 0.264cm 0.33cm}, clip, width=\linewidth]{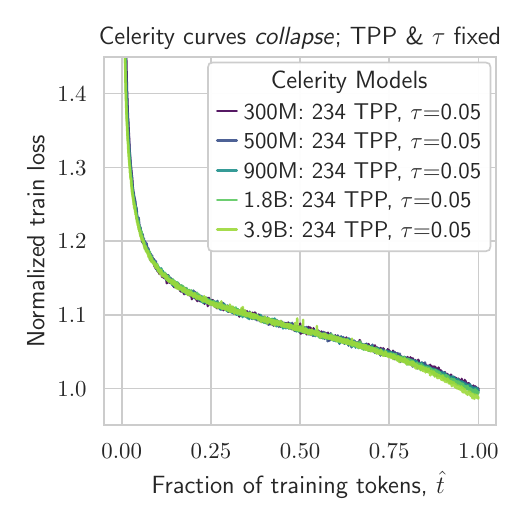}
  \end{minipage}\hfill
  \begin{minipage}{0.33\textwidth}
    \includegraphics[trim={0.32cm 0.42cm 0.264cm 0.33cm}, clip, width=\linewidth]{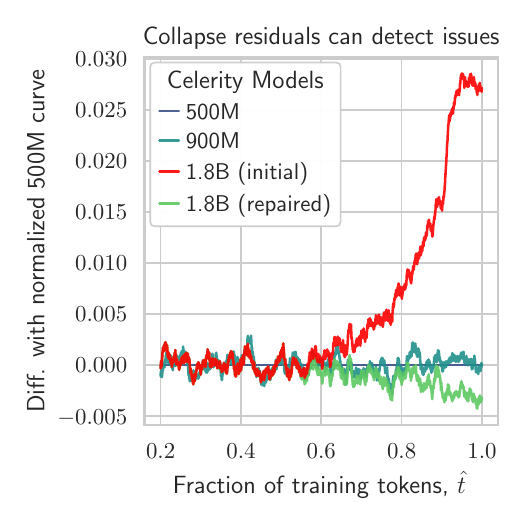}
  \end{minipage}
  \caption{$\Leftfig$: Prior LLM families like Llama-2 train at
    varying tokens-per-parameter (TPP; $D/N$) and AdamW timescale
    $\tema$; training loss curves do \emph{not} collapse.
    $\Middlefig$: Fixing TPP and setting $\tema$ optimally for that
    TPP, Celerity loss curves \emph{do} collapse.
    $\Rightfig$: Deviations from collapse allow precise identification
    (and earlier repair) of numerics issues in large-scale training
    runs.
    \label{fig:hook}}
\end{figure}


\section{Introduction}\label{sec:intro}
Scaling up pre-training has emerged as the primary route to improving
LLM performance~\citep{brown2020language,achiam2023gpt}. Yet once we
reach frontier scales, opportunities for direct experimentation
disappear~\citep{xiao2024rethinking}. How then can we train
effectively at those scales---what size of model should we use, and
how should we set hyperparameters? Encouragingly, recent work has
revealed that several quantities are remarkably \emph{predictable} as
we scale deep learning.  These include model performance as a function
of model and dataset
size~\citep{hestness2017scalinglaws,kaplan2020scaling}, as well as
hyperparameters under maximal update parameterization ($\mup$), which
enables optimal base learning rates and initializations to
approximately transfer across widths~\citep{yang2022mup}.  In this
paper we build on this trajectory of predictability: we show that, at
LLM scale, training loss curves (TLCs) from different model
sizes \emph{collapse} onto a single universal curve after a simple
normalization---provided models are trained with a particular
hyperparameter-scaling recipe.

\citet{qiu2025scaling} only recently demonstrated this striking
regularity in TLCs, showing collapse when training with $\mup$ on
small-scale autoregressive tasks. As their validation was limited to
small models trained with vanilla Adam~\citep{kingma2014adam}, without
weight decay, they explicitly call for tests at larger scales with
\emph{practical scaling ladders} that co-scale width, depth, batch
size, and weight decay. Our work addresses this gap, showing that
collapse persists in full-scale LLM families.

While modeling LLM loss is an active research topic
(\cref{sec:related}), the ability to predict TLCs has
great \emph{practical} value.  For example, human judgment is now
required to decide whether training has \emph{recovered} from a loss
spike---or whether rewinding/restarting is
needed~\citep{chowdhery2022palm,zhang2022opt}.  Other subjective
signals, such as a gradual
upward \emph{trend}~\citep{zhang2022logbook}, can also trigger
interventions. Yet criteria remain vague: \citet{touvron2023llama2}
report Llama-2 TLCs ``did not show any sign of saturation,'' but how
to recognize saturation is unclear.  If TLCs collapse across sizes,
practitioners can compare in-progress training to a universal
reference, monitor residuals, and extrapolate final loss from partial
trajectories. Teams already rely on TLCs in this way, often without a
principled account of what governs TLC shape; for example, Falcon's
final LR was chosen by simply continuing the run performing best after
warmup~\citep{almazrouei2023falcon}.

In this paper we show that the essential condition for collapse under
$\mup$ is that the LR schedule, tokens-per-parameter ratio (TPP), and
AdamW timescale $\tema$~\citep{wang2024how} are held fixed across
model sizes. This reflects a deeper regularity: prior work showed that
optimal $\tema$ depends only on TPP~\citep{bergsma2025power}. Thus,
scaling across fixed TPP with $\tema$ chosen optimally guarantees
collapse, and collapse emerges as a robust marker of compute-efficient
and stable pre-training. When $\tema$ is mis-scaled---as in the Llama-2
family (\cref{fig:hook}, \leftfig)---normalized curves fail to align.

\begin{figure}[ht]
  \centering
    \begin{tikzpicture}
      \node[anchor=south west, inner sep=0] (img) {\includegraphics[trim={0.6cm 0.0cm 0.9cm 0.0cm}, width=0.52\columnwidth]{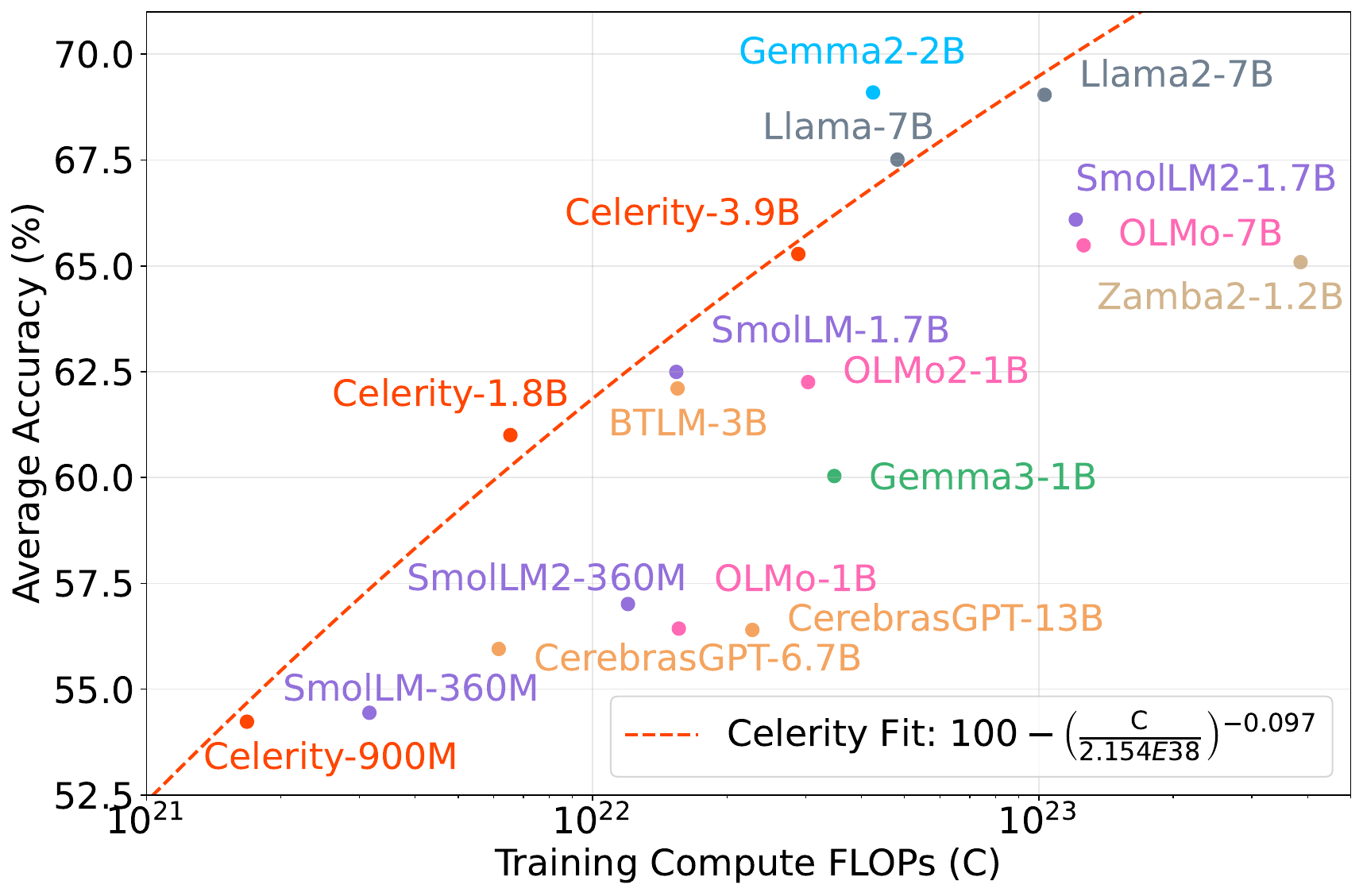}};
      \begin{scope}[x={(img.south east)}, y={(img.north west)}]
        \draw[->, line width=0.8pt, >=Latex]
        (0.29,0.76) -- (0.19,0.86)
        node[above, font=\small, xshift=-1pt, yshift=1pt] {better};
      \end{scope}
    \end{tikzpicture}
    \caption{Celerity is at the compute-efficiency frontier. (Average
      accuracy on tasks \emph{arc-c, arc-e, boolq, hellaswag, piqa,
      siqa, winogrande}; see \cref{tab:all_evals}.)}
    \label{fig:celerity_compute1}   
\end{figure}

We introduce \emph{Celerity} as the first LLM family trained with both
optimal $\tema$ scaling and demonstrable TLC collapse
(\cref{fig:hook}, $\middlefig$). Effective parameterization, including
tuning and transferring $\tema$, helped Celerity land on the
compute-efficiency frontier for open models of its scale
(\cref{fig:celerity_compute1}). Meanwhile, deviations from collapse
provided a sensitive diagnostic of training issues: in our 1.8B run, a
numerical instability became evident from collapse residuals
(\cref{fig:hook}, $\rightfig$) well before the raw TLC showed an
upward trend (\cref{fig:celerity}, $\rightfig$).  Celerity
exemplifies \emph{scaling with collapse}: efficient, predictable
training, across scales and throughout the run.

In summary, our main contributions are:

\begin{itemize}
\item Identifying the key factors influencing loss curve shape under $\mup$:
      the LR schedule, the TPP ratio, and the AdamW timescale $\tema$,
      and explaining shape dependence on these quantities
      (\cref{sec:shape}).
\item Demonstrating that when $\tema$ is set optimally for a given TPP,
      TLCs \emph{collapse} across model scales, providing a signature
      of compute-efficient training (\cref{sec:shape}).
\item Introducing the \emph{Celerity} family, the first large-scale LLMs
      trained in a collapse regime (\cref{sec:celerity}).
\item Proposing a simple functional form for normalized TLCs, and
      showing that fitting this form on small-scale training runs
      enables early stopping in large-scale hyperparameter tuning       
      (\cref{sec:tuning}).
\end{itemize}

\section{Background}\label{sec:background}
Training loss curves (TLCs) for different model sizes typically differ
in scale, duration and final loss. Yet after a simple normalization,
they can align closely, or \emph{collapse}, onto a common trajectory.
Collapse emerges only when three controls are matched across
scales. The tokens-per-parameter ratio (TPP) determines how much data
each parameter sees and thus affects the normalized pace of
improvement; the AdamW timescale $\tau$ governs how long the optimizer
``remembers'' past gradients, shaping the bias--variance trade-off
over training; and the learning-rate schedule phases early bias
reduction against late variance suppression. When TPP and $\tau$ are
chosen consistently across model sizes and the LR schedule is fixed,
the resulting normalized TLCs become approximately scale-invariant. We
now formalize these quantities; \cref{tab:symbols} summarizes key
symbols.

\paragraph{TPP.}

The TPP ratio is equal to number of training tokens $D$ divided by the
model size $N$.  This simple quantity plays a surprisingly profound
role in compute-efficient LLM training and TLC
shape.  \citet{hoffmann2022empirical} investigated, for a given
compute budget $C$, how to allocate $D$ and $N$ in order to minimize
loss.  They found optimal $D$ and $N$ scale roughly equally as $C$
increases, with the optimal $D/N$ ratio relatively constant at around
20 TPP (\cref{sec:compression}).  Replication studies have found
similar results~\citep{besiroglu2024chinchilla,porian2024resolving},
and 20 TPP has emerged as a rule-of-thumb for compute-optimal
training~\citep{dey2023cerebras,zhang2024how}.

\paragraph{$\mup$.}

$\mup$~\citep{yang2020feature} and related parameterizations for
depth~\citep{bordelon2023depthwise,yang2023tensor,dey2025dont} seek to
achieve consistent, stable training dynamics as networks scale up.
Moreover, with $\mup$, base hyperparameters can be tuned on a
small \emph{proxy} model and then transferred to larger scales.
Given the width of the proxy model, $d_p$, and target, $d_t$, $\mup$
prescribes scaling factors to apply to the base LR, initial weight
variance, and other base HPs.

$\mup$ is increasingly used in LLM
training~\citep{dey2023cerebras,dey2023btlm3b8k,sengupta2023jais,shen2024power,hu2024minicpm}.
Moreover, recent work has shown that, when using $\mup$, other
important aspects of training may \emph{decouple} from model size,
including optimal batch size (scaling primarily in the total number of
tokens~\citep{zhang2024how,bergsma2025power}), and optimal AdamW
timescale/weight decay (scaling primarily in
TPP~\citep{bergsma2025power}).
%

\paragraph{Supercollapse.}

Using $\mup$, \citet{qiu2025scaling} observed that TLCs for different
model sizes, despite varying widely over compute and absolute loss,
appear to follow a consistent shape.  This motivated them to affinely
rescale the curves to the normalized loss $\ell$ given by:
\begin{equation}\label{eq:qiu}
\ell(\trainfrac, N, \omega) = 
                 (L(\trainfrac \cdot \tstar(N), N, \omega) - \lhat)/
                 (L(\tstar(N), N, \omega) - \lhat)
\end{equation}
where $\omega$ is the random seed, $\trainfrac$ is the fraction of
training completed (what \citet{qiu2025scaling} refer to as
\emph{normalized compute}), $N$ is the number of model parameters, and
$\tstar(N)$ is the corresponding compute-optimal number of training
steps, estimated from a power law fit.  $\lhat$ is an offset, which
they subsequently set to the estimated irreducible loss of their power
law.

Training compute-optimally under $\mup$ (on small-scale autoregressive
tasks, e.g., predicting chess moves), \citet{qiu2025scaling} showed
TLCs collapse under this normalization---indeed, they
\emph{super}collapse, meaning they differ by less than the noise
from inter-run variation.
They further show that collapse arises naturally in
constant-learning-rate models where loss obeys typical neural power
laws, while extending the theory to arbitrary LR schedules via a
theoretical model of quadratic loss.

\paragraph{The AdamW EMA and its timescale.}

AdamW updates at step $t$ can be expressed in terms of learning rate
$\eta$ and weight decay $\lambda$ as: $\theta_t = (1 -
\eta\lambda)\theta_{t-1} - \eta \frac{\hat{m}_t}{\sqrt{\hat{v}_t} +
  \epsilon}$, where $\hat{m}_t$ and $\hat{v}_t$ are bias-corrected
EMAs of gradients and squared gradients,
respectively~\citep{kingma2014adam}.  \citet{wang2024how} observed
that AdamW parameters $\theta_t$ can also be viewed as an EMA---of
weight \emph{updates}. That is, the standard EMA form \mbox{$y_t = (1
- \alpha)y_{t-1} + \alpha x_t$} matches AdamW
when \mbox{$y_t=\theta_t$}, \mbox{$\alpha=\eta\lambda$},
and \mbox{$x_t=-\frac{1}{\lambda}\frac{\hat{m}_t}{\sqrt{\hat{v}_t}
+ \epsilon}$}. The \emph{timescale} $\titer = \nicefrac{1}{\alpha} =
\nicefrac{1}{\eta\lambda}$ represents the approximate number of
iterations over which updates are averaged.
When expressed in epochs as $\tepochwang = \titer/M$, where $M$ is the
number of iterations per epoch, \citet{wang2024how} found the optimal
$\tepochwang$ (swept by varying $\lambda$) \emph{remains stable} under
model and dataset scaling on image tasks.

Since LLM pre-training typically uses a single epoch, we follow
\citet{bergsma2025power} in defining a
normalized timescale $\tepoch = \titer/T$, where $T$ is the total
number of optimization steps. As $T = D/B$ (total tokens/batch size):
\begin{equation}
\tema = 1/(\eta\lambda T) = B/(\eta\lambda D).
\end{equation}
In contrast with the results
in \citet{wang2024how}, \citet{bergsma2025power} did \emph{not} find
optimal $\tema$ to remain stable in LLM training, but instead to
decrease as a (scale-invariant) power law in TPP.\@



\begin{table}[ht]
  \centering
  \caption{Core quantities used throughout the paper.\label{tab:symbols}}
  \vspace{-3pt} 
  \small
  \begin{tabular}{ll}
    \toprule
    Symbol & Meaning \\
    \midrule
    $N$ & Number of model parameters \\
    $D$ & Total number of training tokens \\
    $B$ & Batch size (tokens per optimization step) \\
    $T = D/B$ & Total number of optimization steps \\
    $\trainfrac = t/T$ & Fraction of training completed \\
    $\tpp = D/N$ & Tokens-per-parameter ratio (TPP) \\
    \addlinespace
    $L(t)$ & Training loss at step $t$ \\
    $\ell(\trainfrac)$ & Normalized training loss curve (TLC) \\
    \addlinespace
    $\eta$ & Learning rate \\
    $\lambda$ & Weight decay coefficient \\
    $\titer = 1/(\eta\lambda)$ & AdamW timescale (in steps) \\
    $\tema = \titer/T = 1/(\eta\lambda T) = B/(\eta\lambda D)$ & Normalized AdamW timescale \\
    \bottomrule
  \end{tabular}
\end{table}

\section{What factors modulate training curve shape?}\label{sec:shape}
\paragraph{Experimental setup.}

We use a GPT2-like LLM~\citep{radford2019gpt2}, with ALiBi
embeddings~\citep{press2022alibi} and SwiGLU~\citep{shazeer2020glu}.
We train on SlimPajama~\citep{cerebras2023slimpajama}.
Models are trained with AdamW and $\mup$.  We use a linear
decay-to-zero LR schedule,
context length of 2048, and the GPT2 vocabulary.
Full architecture and other details are in
\cref{sec:experimental_details}.

We plot $\ell$ vs.\ \emph{training fraction} $\trainfrac = t/T =
tB/D$, with step count $t$, total steps $T$, batch size $B$, and
dataset size $D$.
To reduce noise in small-$B$ settings, we \emph{post hoc} aggregate
losses $\ell(\trainfrac)$ using a moving-average filter over a window
of 100 steps, smoothing curves without altering the underlying
trajectory.
We also consistently found simply \emph{dividing by the final training
loss} (i.e., $\lhat = 0$ in \cref{eq:qiu}) resulted in optimal
alignment across scales, so use this for all curves.

\paragraph{Finding: $\tema$ modulates TLC shape.}

\begin{figure}[ht]
  \centering
  \begin{minipage}{0.33\textwidth}
    \includegraphics[trim={0.3cm 0.42cm 0.264cm 0.3cm}, clip, width=\linewidth]{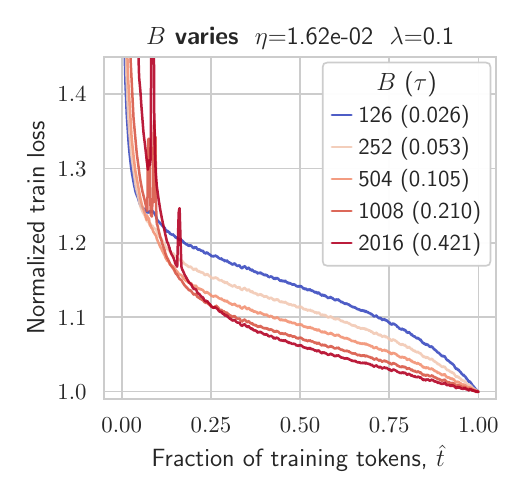}
  \end{minipage}\hfill
  \begin{minipage}{0.33\textwidth}
    \includegraphics[trim={0.3cm 0.42cm 0.264cm 0.3cm}, clip, width=\linewidth]{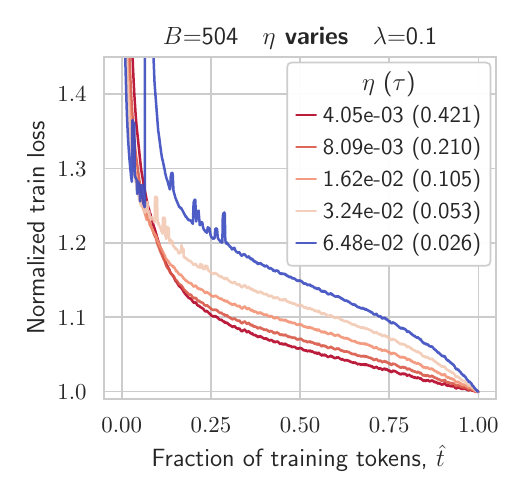}
  \end{minipage}\hfill
  \begin{minipage}{0.33\textwidth}
    \includegraphics[trim={0.3cm 0.42cm 0.264cm 0.3cm}, clip, width=\linewidth]{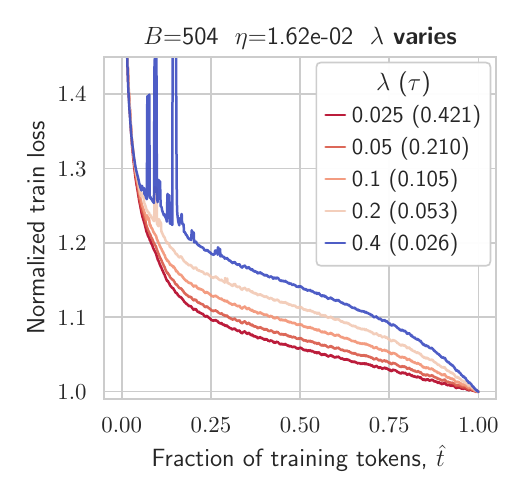}
  \end{minipage}
  \caption{\textbf{AdamW timescale $\tema$ modulates TLC shape (610M,
      80TPP)}: Sweeping $\eta$ ($\leftfig$), $\lambda$ ($\middlefig$),
    or $B$ ($\rightfig$) produces matching variations in normalized
    TLCs when $\tema$ varies identically.\label{fig:tema}}
\end{figure}

\cref{fig:tema} shows normalized TLCs for 610M models
trained to 80~TPP, sweeping either learning rate $\eta$, weight decay
$\lambda$, or batch size $B$ in each subplot. Across hyperparameters,
TLCs with matching $\tema$ exhibit very similar shapes, reflecting
consistent timescale control. Similar patterns hold across other
scales and dataset sizes.
Generally, as $\tema$ increases, TLCs drop more early and less later.
This is also a function of the LR schedule: when we switch to using a
$\constant$ LR, there is no final drop, lower-$\tema$ TLCs are lower
throughout (appendix \cref{fig:lrsched}).

\paragraph{Finding: TPP modulates TLC shape.}

\begin{figure}[ht]
  \centering
  \begin{minipage}{0.33\textwidth}
    \includegraphics[trim={0.3cm 0.42cm 0.264cm 0.3cm}, clip, width=\linewidth]{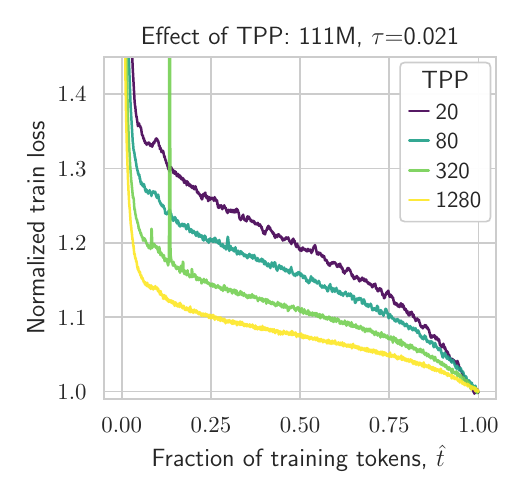}
  \end{minipage}\hfill
  \begin{minipage}{0.33\textwidth}
    \includegraphics[trim={0.3cm 0.42cm 0.264cm 0.3cm}, clip, width=\linewidth]{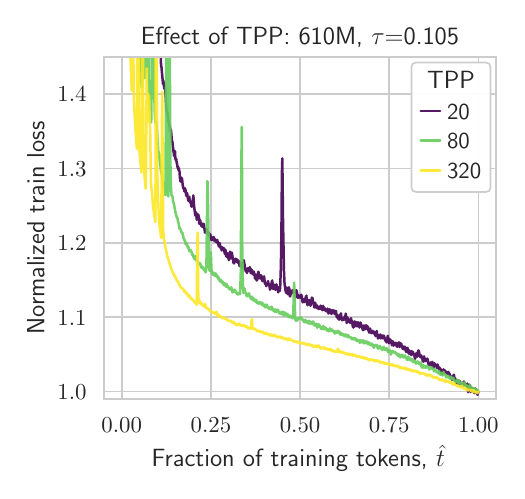}
  \end{minipage}\hfill
  \begin{minipage}{0.33\textwidth}
    \includegraphics[trim={0.3cm 0.42cm 0.264cm 0.3cm}, clip, width=\linewidth]{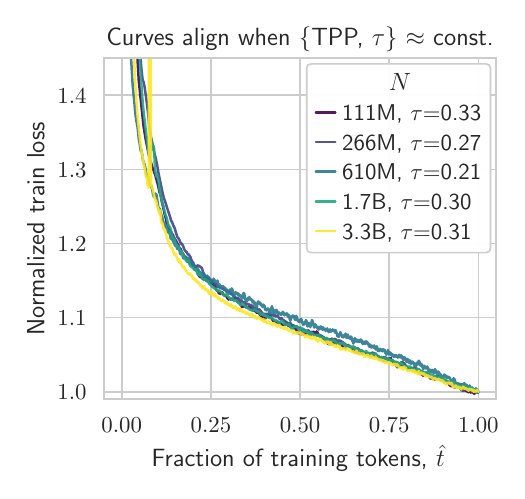}
  \end{minipage}
  \caption{\textbf{TPP modulates TLC shape.}
    Fixing $\tau$ for 111M ($\leftfig$) \& 610M ($\middlefig$) while
    \emph{increasing} TPP, curves shift \emph{down}.  When $\tema$
    $\approx$ const.\ and TPP also fixed (at 20), curves roughly
    collapse ($\rightfig$).\label{fig:tpp}}
\end{figure}

We now fix $\tema$ and test increasing \emph{TPP}, finding TLCs drop
earlier and flatten for longer (\cref{fig:tpp}, $\leftfig$,
$\middlefig$; see also Llama-2 for $\tau=0.07$ and $\tau=0.13$
in \cref{fig:hook}, $\leftfig$).  Intuitively, relative to length of
training, higher TPP drops loss more at the beginning and then obtains
diminishing returns later on.
In \cref{fig:tpp}, $\rightfig$, TLC shape is quite similar across
model scales at the same TPP (when $\tema$ is roughly equal), showing
TPP's shaping effect is \emph{scale-invariant} (scaling from 111M to
3.3B at fixed TPP represents a $1000\times$ increase in training
FLOPs).

\paragraph{Using bias and variance to explain the effect of $\tema$.}

Classical analyses of stochastic optimization
(e.g., \citet{bottou2018optimization,dangelo2024why}) decompose loss
into two components: a \emph{bias} term, reflecting dependence on
initial weights, and a \emph{variance} term, reflecting noise from
stochastic gradients. Early in training bias dominates; later,
variance determines the attainable loss floor.

In our setting, the AdamW timescale $\tema$ controls this trade-off
through the EMA over weight updates (\cref{sec:background}). A smaller
$\tema$ corresponds to a shorter memory: updates depend mainly on
recent gradients, yielding rapid bias reduction but a higher variance
floor. A larger $\tema$ averages over more past gradients, reducing
variance more effectively but slowing early progress. Empirically,
TLCs reflect this pacing, with smaller $\tema$ producing faster early
descent (e.g., left panel of appendix~\cref{fig:lrsched}).

This intuition is formalized in \cref{sec:app_tema} via a noisy
quadratic model:
\begin{equation}
\label{eq:tema_main}
\mathbb{E}[L(\trainfrac)]
= \frac{h\,\sigma_x^2}{4\,\tema}\Big(1-e^{-2\trainfrac/\tema}\Big)
\;+\; \frac{h}{2}\,e^{-2\trainfrac/\tema}\,\mathbb{E}[\theta(0)^2],
\end{equation}
where the first term approaches a variance floor $\propto 1/\tema$ and
the second is an exponentially decaying bias term. Smaller $\tema$
thus yields faster initial decay but a higher floor, while larger
$\tema$ yields slower initial decay but a lower floor, matching the
observed \emph{fast-then-flatten} behavior under constant LR.
With LR decay, $\eta_t\lambda$ decreases and the instantaneous
timescale $\tema_t=1/(\eta_t\lambda T)$ \emph{increases}, enhancing
late-stage variance suppression and steepening final drop
(e.g., \cref{fig:lrsched}: more LR decay, more drop).

\emph{Scale invariance.}
After normalizing by the final loss, the curvature factor $h$
cancels (\cref{sec:app_tema}). Provided residual bias at
end-of-training is negligible relative to the variance floor, the
normalized TLC depends only on $\tema$ and $\trainfrac$. Thus, at
matched $\tema$, normalized TLCs collapse across scales.

\paragraph{Explaining effect of TPP.}

TPP affects TLCs via power laws. With a \emph{constant} LR, every step
is the endpoint of a shorter run, trained to
$\trainfrac \cdot \tpp$. \citet{qiu2025scaling} note $L(\trainfrac)$
therefore \emph{follows the same power law as final loss of a run
fully trained to that effective budget}.  Normalizing by the projected
loss at \emph{total} TPP removes dependence on model and dataset size,
so $\ell(\trainfrac)$ depends only on $\trainfrac$ and total TPP
(\cref{sec:app_tpp}).  Higher-TPP curves analytically decay faster and
level off sooner.  LR schedules deform the curves, but deformation is
also scale invariant given consistent curvature of the loss landscape
across model sizes under $\mup$~\citep{noci2024super}.


\takeaway{
TLC shape is governed by three scale-invariant controls: (1)~AdamW
timescale $\tau$ (bias--variance trade-off), (2)~tokens-per-parameter
ratio (TPP), which sets the relative pace of improvement, and
(3)~learning-rate schedule, which phases early and late loss
reduction. When matched across model sizes, normalized TLCs follow the
same trajectory---that is, they collapse.}

\section{Celerity: A compute-efficient model family with collapse}\label{sec:celerity}
We have established that collapse arises when $\tema$ and TPP are held
fixed across model sizes.  Meanwhile, prior work has shown optimal
$\tema$ to depend only on TPP~\citep{bergsma2025power}. Here, we
introduce a model family, \emph{Celerity}, trained at fixed TPP and
with $\tema$ chosen optimally for that TPP, i.e., a regime where
collapse emerges naturally as a consequence of good training.

\paragraph{Compute vs.\ parameter efficiency.}

A key question for Celerity is which TPP to use: $\approx{20}$ is
compute-optimal (\cref{sec:background}), while higher TPP means
greater \emph{parameter efficiency} (fewer parameters to obtain same
loss).
For small, inference-ready models, parameter efficiency is paramount,
but such models are usually
distilled~\citep{tunstall2023zephyr,wang2025distilqwen2} rather than
pre-trained at high-TPP\@.
\emph{Our main interest is developing pre-training strategies for very large models}.
As models scale, the relative importance of compute-efficiency
increases---indeed, public families often have declining TPP as size
increases~\citep{touvron2023llama,biderman2023pythia}.

\begin{figure}
  \centering
    \begin{tikzpicture}
      \node[anchor=south west, inner sep=0] (img) {\includegraphics[width=0.38\columnwidth,trim={0.26cm 0.26cm 0.27cm 0.26cm},clip]{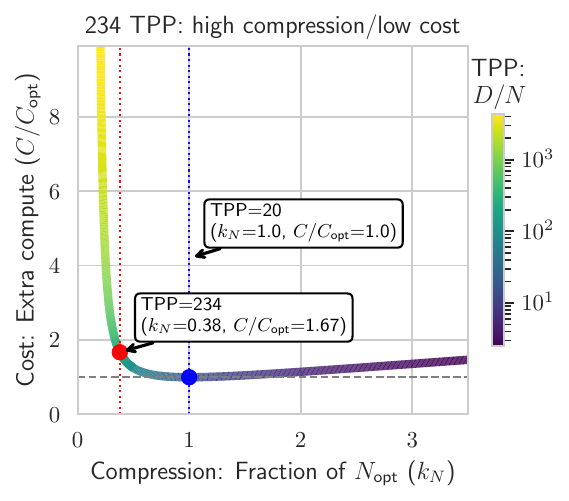}};
      \begin{scope}[x={(img.south east)}, y={(img.north west)}]
        \draw[->, line width=0.8pt, >=Latex]
        (0.8,0.9) -- (0.7,0.8)
        node[below, font=\small, xshift=-1pt, yshift=1pt] {better};
      \end{scope}
    \end{tikzpicture}
    \caption{Expected iso-loss compute vs.\ compress trade-off as TPP varies.\label{fig:compression}}
\end{figure}

Yet even for the largest models, parameter efficiency remains
valuable, e.g., when generating distillation logits or synthetic data.
To choose Celerity's TPP, we analyze this trade-off:
\cref{sec:compression} derives an expression for the extra
compute required to compress a model to a fraction of the
\emph{size when training compute optimally}---while maintaining the same loss.
This expression leverages power law fits from prior work.
\cref{fig:compression} plots the trade-off, where a TPP ratio
of 234 is estimated to achieve a 62\% reduction in parameters with
only a 67\% increase in total FLOPs (relative to 20 TPP). This is a
responsible balance point, near what has been called
the \emph{critical model size}---the point where further increasing
compute obtains massively-diminishing returns in parameter
efficiency~\citep{devries2023chinchilla_analysis} (e.g.,
doubling \emph{our} FLOPs, to 3.34$\times$ compute-optimal, reduces
$N$ by only a further $\approx 11\%$).
  
Even if the ultimate goal is a ``herd'' of models at varying TPP, such
as Llama-2 in \cref{fig:hook}, $\leftfig$, there are advantages to
training different ``bands'' within the herd, e.g., 7B, 13B, 34B,
70B \emph{all} at 29 TPP:
\begin{itemize}
\item \emph{Tuning}: you can
      fine-tune $\tema$ at a smaller scale and zero-shot transfer to
      larger models.
\item \emph{Diagnostics}: Because TLCs collapse, deviations provide
      an early warning of training issues.
\item \emph{Cost}: Fixed-TPP bands are cheap (e.g., 10$\times$ lower $N$
      $\rightarrow$ 10$\times$ smaller $D$ $\rightarrow$ 100$\times$
      less compute).
\end{itemize}

\paragraph{Philosophy.}

Celerity aims to advance general LLM capabilities using public
pre-training corpora and fully-open, consistent methods---rather than
targeting specific benchmarks.  In contrast, the majority of LLMs now
\emph{anneal on training subsets of downstream
benchmarks}~\citep{dubey2024llama,achiam2023gpt}, or inject special
high-quality math~\citep{olmo2024}, code~\citep{zhang2024map}, or
instruction~\citep{hu2024minicpm} data during a
late-stage \emph{mid-training} process.
Since these practices make evaluation
problematic~\citep{dominguez2024training}, Celerity can serve as a
comparison for models trained without (or prior to) applying such
techniques.

\paragraph{Experimental details.}

\begin{table}
  \centering
  \caption{Architecture of Celerity\label{tab:celerity_model_arch}}
  \small
  \begin{tabular}{lccccc}
    \toprule
    Celerity: & 300M & 500M & 900M & 1.8B & 3.9B \\
    \midrule
    Hidden Dim & 640 & 896 & 1152 & 1536 & 2048 \\ 
    Num Heads & 10 & 14 & 9 & 12 & 16 \\
    Head Size & 64 & 64 & 128 & 128 & 128 \\
    Layers & 13 & 17 & 23 & 30 & 40 \\
    \midrule
    Batch Size & 176 & 240 & 336 & 464 & 672\\
    \midrule
    Vocabulary & \multicolumn{5}{c}{Llama-3 (size 128256)} \\
    Embeddings & \multicolumn{5}{c}{ALiBi, Untied} \\
    Seq Length & \multicolumn{5}{c}{8192} \\
    \midrule
    Non-linearity & \multicolumn{5}{c}{Squared ReLU} \\
    FFN Mult & \multicolumn{5}{c}{8$\times$} \\
    Norm Type & \multicolumn{5}{c}{Pre-Layer Normalization, $\epsilon=10^{-5}$} \\
    \midrule
    LR Schedule & \multicolumn{5}{c}{Peak: 0.15, linear decay-to-zero} \\
    LR warmup & \multicolumn{5}{c}{min(10\% of total tokens, 375M tokens)} \\
    \bottomrule
  \end{tabular}
\end{table}

Celerity pre-trains in bands of 20, 80, and 234 TPP, each spanning
300M--3.9B models (\cref{tab:celerity_model_arch});
see \cref{sec:celerity_details} for further details. Key enablers of
Celerity's reliable, efficient training include:

\begin{itemize}
\item \emph{Data}: emphasizing (open) educational, math, and
      coding data throughout training
      (appendix \cref{tab:celerity_datasets}); this outperformed
      training on the general SlimPajama dataset
      (\cref{tab:celerity_data_comp}).
\item \emph{Parameterization}: Using CompleteP, which enables hyperparameter transfer
      over width \emph{and} depth, was more efficient/reliable than
      $\mup$ (\cref{fig:parameterization}).
\item \emph{Optimization}: LR, $\tema$, batch size tuned small,
      transferred via scaling rules.
\end{itemize}

\paragraph{Evaluation results.}

Appendix \cref{tab:all_evals} provides full downstream evaluation
results for Celerity and other public models tested on seven common
downstream tasks.
\cref{fig:celerity_compute1} shows that Celerity models form the accuracy/compute Pareto frontier up to our
largest training budget.
Against BTLM~\citep{dey2023btlm3b8k}---trained before task-specific
data annealing became standard---Celerity achieves comparable accuracy
with 75\% fewer training FLOPs.
%
%
Extrapolation via a fitted power law in compute (dashed line in plot)
suggests smooth scaling and continued competitiveness.
For comparison with distilled models, we count only \emph{student}
FLOPs in \cref{fig:celerity_compute1}.  Including \emph{teacher} FLOPs
(forward passes), or the cost of teacher training, strengthens
Celerity further (appendix \cref{fig:celerity_compute2}).

In terms of parameter efficiency, Celerity is weaker than high-TPP
families (\cref{fig:celerity_parameter1,fig:celerity_parameter2}),
meaning such models save FLOPs at inference.
However, beyond the importance of studying compute efficiency for
hyper-scale training, there is strong motivation to train and study
compute-efficient smaller models: growing evidence suggests some
models may be counter-productively (even
\emph{catastrophically}) overtrained, making them harder to
fine-tune~\citep{springer2025overtrained} and
quantize~\citep{kumar2024scaling}. Compute-efficient alternatives
therefore serve both as a principled baseline for understanding
scaling and as a practical fallback when high-TPP models prove
brittle.

\paragraph{Collapse results.}

In \cref{sec:shape}, we normalized training loss curves by dividing by
the final loss value, $L(T)$ (\cref{eq:qiu}). To use collapse as a
diagnostic \emph{during} training, we need a way to normalize when
$L(T)$ is still unknown. We explored two strategies and use
early-align in our experiments:
\begin{itemize}
\item \emph{Estimate}: extrapolate $L(T)$ from a power law fit at lower scales.
\item \emph{Early-align}: choose $L(T)$ so $\ell(t)$
best aligns with the smallest-scale curve over 25-50\% portion.
\end{itemize}

\begin{figure}[ht]
  \centering
  \begin{minipage}{0.33\textwidth}
    \includegraphics[trim={0.3cm 0.42cm 0.264cm 0.3cm}, clip, width=\linewidth]{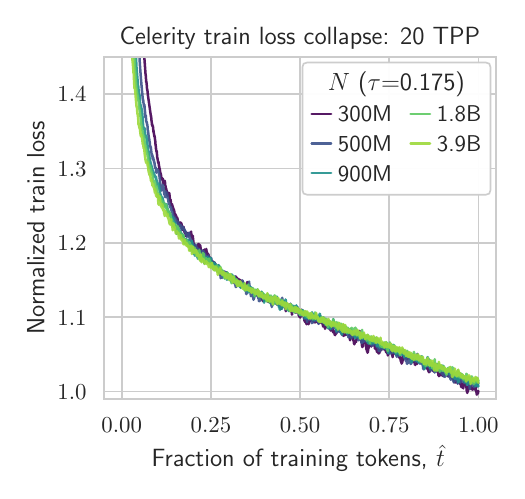}
  \end{minipage}\hfill
  \begin{minipage}{0.33\textwidth}
    \includegraphics[trim={0.3cm 0.42cm 0.264cm 0.3cm}, clip, width=\linewidth]{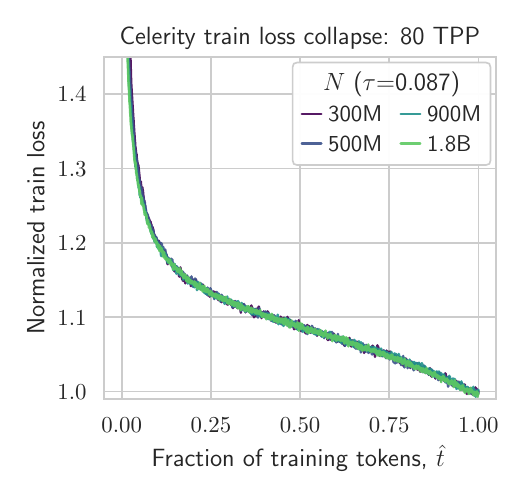}
  \end{minipage}\hfill
  \begin{minipage}{0.33\textwidth}
    \includegraphics[trim={0.3cm 0.42cm 0.264cm 0.3cm}, clip, width=\linewidth]{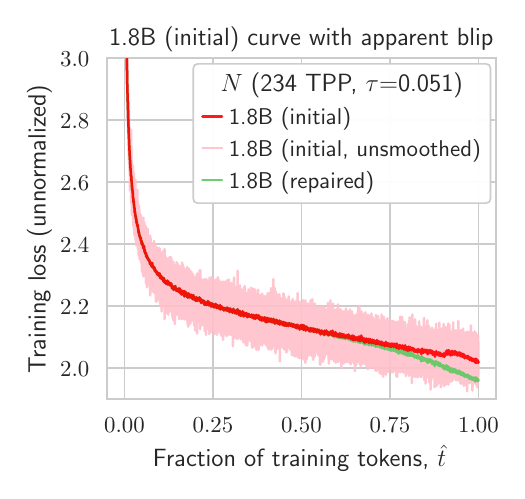}
  \end{minipage}
  \caption{\textbf{Collapse in Celerity models.}
    Celerity 20 TPP ($\leftfig$) and 80 TPP ($\middlefig$) models
    exhibit collapse.  $\Rightfig$: smoothing helps detect blip in
    loss near the end of initial 1.8B run (red curve), but divergence
    can be detected much earlier using collapse residuals
    (\cref{fig:hook}, $\rightfig$).\label{fig:celerity}}
\end{figure}

\cref{fig:celerity} shows normalized curves. Collapse is tight at
80 TPP ($\middlefig$). At 20 TPP ($\leftfig$), we see small early
deviations, which we attribute to differing LR warmup proportions
(\cref{tab:celerity_model_arch}). At 234 TPP, divergences appear late
in training for larger models (\cref{fig:hook}, $\middlefig$).
Investigating, we find loss improves disproportionately on training
data, while held-out data remains aligned with projections.

\paragraph{Collapse for monitoring.}

\cref{fig:celerity} ($\rightfig$) shows the unnormalized TLC for our
original 1.8B, 234 TPP run. Smoothing helps reveal a sudden rise in
training loss, but only after 90\% of training. Without a collapse
reference, it would be impossible to see that problems began much
earlier. By comparing against the 500M TLC reference (\cref{fig:hook},
$\rightfig$), we pinpoint divergence starting near 60\%. Knowing this
timing was crucial: we did not waste effort investigating late-stage
data redundancy, and instead realized the problem coincided with a job
restart under a new compute allocation.

The collapse reference was also essential for debugging: by running
ablations with different batch sizes and measuring divergence from the
reference, we confirmed the anomaly arose from a numerical issue in a
loss kernel triggered only at specific microbatch sizes. After fixing
the kernel and restarting from before the divergence, training tracked
the reference TLC closely (\cref{fig:hook}).


\takeaway{
Celerity trains models in fixed-TPP bands at the optimal $\tau$ for
each band. This produces collapse across scales, enabling
hyperparameter transfer, early detection of training issues, and
reliable cross-scale comparisons. The 234-TPP band lies on the
compute--accuracy frontier while using $\approx$62\% fewer parameters
than compute-optimal training at equal loss.}

\section{Collapse enables early stopping in hyperparameter tuning}\label{sec:tuning}
Training to completion is expensive. If normalized TLCs behave
predictably, can we stop earlier and still recover the final loss?  We
show collapse enables principled early stopping in tuning, and
introduce a predictive model---fit at small scales, and re-used to
extrapolate large-scale TLCs.

\paragraph{Role of $\tema$ in tuning.}

\begin{figure}[ht]
  \centering
  \begin{minipage}{0.6735\textwidth}
    \begin{minipage}{0.49\textwidth}
      \includegraphics[trim={0.3cm 0.42cm 0.264cm 0.3cm}, clip, width=\linewidth]{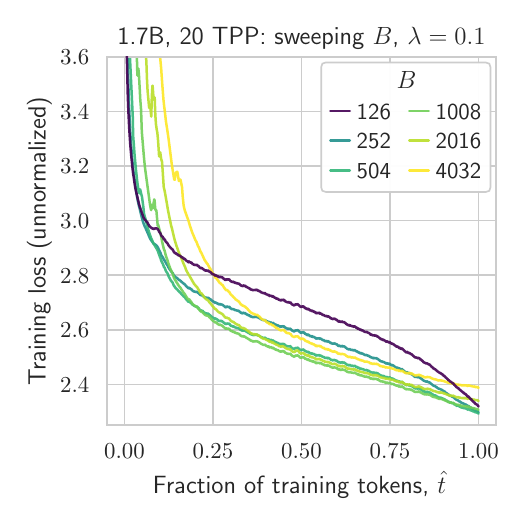}
    \end{minipage}\hfill
    \begin{minipage}{0.49\textwidth}
      \includegraphics[trim={0.3cm 0.42cm 0.264cm 0.3cm}, clip, width=\linewidth]{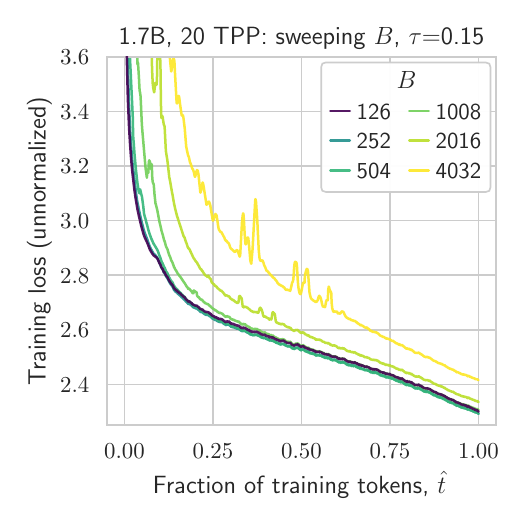}
    \end{minipage}
  \end{minipage}
  \caption{\textbf{Predictability of batch size sweeps.}  $\Leftfig$:
    When fixing weight decay $\lambda$ in batch size sweeps (standard
    practice), normalized TLCs \emph{cross}, making final loss hard to
    predict from partial results.  $\Rightfig$: Fixing $\tema$ instead
    (by adjusting $\lambda$), TLCs maintain ordering, enabling early
    stopping.\label{fig:hptuning}}
\end{figure}

Recent work tunes learning rate $\eta$ and batch size $B$ at smaller
scales and extrapolates via power
laws~\citep{hu2024minicpm,bi2024deepseek,porian2024resolving}. These
studies typically fix weight decay $\lambda$, which unintentionally
varies $\tema$---and hence TLC shape. As
\cref{fig:hptuning} ($\leftfig$) shows, when $\tema$ varies, mid-training
loss is a poor predictor of final outcomes during a batch size
sweep. In contrast, when $\tema$ is fixed during tuning (by adjusting
$\lambda$), the ordering of $B$-specific curves is preserved
throughout training ($\rightfig$); runs can be stopped early (e.g., at
25\%) while still reliably identifying the best batch size.

There are, however, cases where $\tema$ must vary. For example,
\citet{bergsma2025power} found optimal $\tema$ was no longer
constant once $B>\bcrit$, potentially requiring retuning of $\lambda$.


\paragraph{Exploiting collapse for early stopping.}

Suppose we wish to find the optimal setting of a hyperparameter (HP)
in large-scale training. Naively, we could sweep across settings of
the HP and train a large-scale model to completion at each setting;
the lowest final loss would identify the best choice. Alternatively,
rather than training to completion, we propose a procedure to infer
the final loss values from partial training runs. We do this by
exploiting the collapse phenomenon as follows:

\begin{enumerate}
\item For each large-scale setting in the sweep, first identify the corresponding \emph{TLC controls}, i.e, the LR schedule, $\tema$, and TPP.
\item Train a smaller (e.g., 100M-parameter) model for each unique combination of TLC controls.
\item Normalize these small-run loss curves to obtain
the \emph{normalized universal TLCs} ($\ell(\trainfrac)$)
corresponding to each set of controls.
\item Perform \emph{partial} training runs (e.g., to 30\% of tokens) at each HP setting.
\item For each partial large-scale loss curve, use its corresponding
universal TLC to \emph{predict} the final loss value.  Do this by
determining the divisor $L(T)$ that maximally \emph{aligns} the
partial curve with the same segment of the corresponding universal
TLC. These divisors thus act both as normalizing constants \emph{and},
by construction, as calibrated \emph{extrapolations} of the final
loss.
\item Select the optimal hyperparameter setting corresponding to the lowest predicted
final loss.
\end{enumerate}

\paragraph{Predicting the normalized universal TLCs.}

In some cases, we may already have many small-scale runs, but not loss
curves for each of the \emph{specific} TLC controls corresponding to
our large-scale sweep. In this situation, we hypothesize that
a \emph{parametric surrogate model} can generate high-fidelity
normalized universal TLCs as a function of $\tema$ and TPP. This
allows us to obtain the normalized TLC shapes required for Step 5 of
the above procedure, \emph{without} training small-models at
exactly-matching controls. Fitting such a surrogate lets us leverage
our broader TLC dataset, and enables estimation of normalized TLC
shapes that go beyond trained regimes (see \cref{sec:optimal} for an
example).

For the surrogate $\ell(\trainfrac)$ model, we experimented with
several functional forms and ablations on our 111M-scale data,
focusing on:
\begin{equation}\label{eq:prediction}
  \ellhat(\trainfrac) = ((1 + \epsilon_1)/(\trainfrac + \epsilon_1))^m
                        + b \cdot (\eta(\trainfrac) + \epsilon_2)^q
\end{equation}
The first term captures power-law improvement in training fraction
(\cref{sec:app_tpp}) while the second term modulates this by the LR
schedule $\eta(\trainfrac)$, reflecting how variance suppression is
phased over training (\cref{sec:app_tema}).
$m$, $b$, $q$, $\epsilon_1$ and $\epsilon_2$ are fit parameters.
We divide $\ellhat(\trainfrac)$ by its final value so that $\ellhat(1)
= 1.0$.
Fixing $\epsilon_1=0.001$ and $\epsilon_2=0.1$ avoids large swings at
$\ellhat(0)$ and $\ellhat(1)$.

In practice, we find $m$ can be fixed (we use 0.05).  Parameters $b$
and $q$ then vary systematically with $\tau$ and TPP, respectively,
which we capture with power laws:
\begin{equation}
b = \bconst \cdot (\tema)^{\bexp},\quad q = \qconst \cdot (\tpp)^{\qexp}
\end{equation}
Because $b$ and $q$ interact, jointly fitting their parameters would
require a $\mathcal{O}(g^4)$ grid search (with $g$ the grid
resolution). Instead, we alternate: fit $(\bconst,
\bexp)$ with fixed $q$, then fit $(\qconst,\qexp)$ with fixed $b$,
iterating to convergence. This reduces cost to $\mathcal{O}(g^2)$
while yielding stable fits.

\paragraph{Results: prediction.}

We fit the $b$ and $q$ power laws on 111M-scale data and evaluate
using mean absolute error (MAE) between $\ellhat$ and true $\ell$,
computed over $\trainfrac \in [0.2,1]$ (ignoring error around LR
warmup, when initial curves are noisy).  We report unweighted mean MAE
across all curves.

\begin{figure}[ht]
  \centering
  \begin{minipage}{0.33\textwidth}
    \input{tikz_figs/illustrate_3.3B_pred.tex}
    \captionof{figure}{\textbf{3.3B-scale predictions and true
        normalized TLCs.}}
    \label{fig:prediction}
  \end{minipage}\hfill
  \begin{minipage}{0.66\textwidth}
    \begin{minipage}{0.49\textwidth}
      \includegraphics[trim={0.3cm 0.42cm 0.264cm 0.3cm}, clip, width=\linewidth]{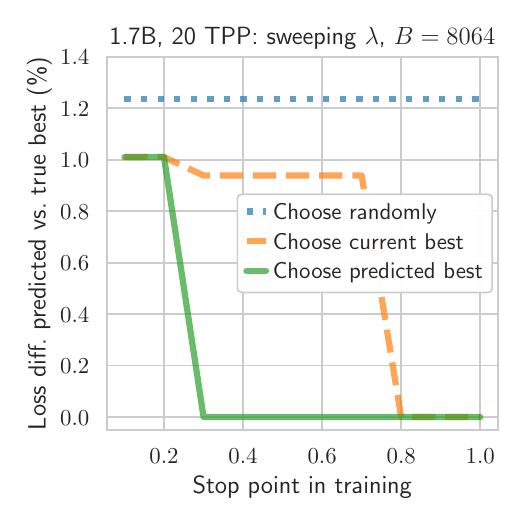}
    \end{minipage}\hfill
    \begin{minipage}{0.49\textwidth}
      \includegraphics[trim={0.3cm 0.42cm 0.264cm 0.3cm}, clip, width=\linewidth]{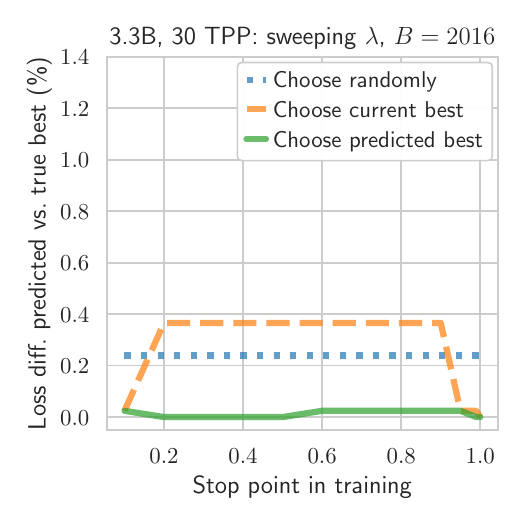}
    \end{minipage}
    \captionsetup{width=.93\linewidth}
    \captionof{figure}{\textbf{Early stopping works best with
        \emph{predicted} loss:} Tuning $\lambda$ in 1.7B ($\leftfig$)
      and 3.3B ($\rightfig$) models.}
    \label{fig:tuning_tests}
  \end{minipage}
\end{figure}

Results show that predictions are good: MAE is low and actually
improves with scale (appendix \cref{tab:fit_scaling}), likely because
(1)~larger datasets yield smoother TLCs, and (2)~fewer extreme
hyperparameter tests at larger scales.
\cref{fig:prediction} ($\leftfig$) shows an example:
predictions trained on 111M-scale TLCs (1000$\times$ fewer FLOPs)
closely match observed curves for a 3.3B model.

Estimating $b$ and $q$ as power laws reduces MAE by two-thirds
compared to using fixed values (\cref{tab:fit_ablation}), though error
remains $\approx$2$\times$ higher than an oracle fit of $b$ and
$q$ \emph{per curve}.  Adjusting for both $\tau$ and TPP is vital;
however, fitting $b$ and $q$ jointly on both did not improve further.

\paragraph{Results: tuning.}

We now test whether optimal LLM settings can be predicted from partial
training runs. At different stopping points in training, we choose a
setting as the best, and evaluate the gap between the chosen setting's
final loss and the true best setting. We compare the following
choices:
\begin{enumerate}
\item \emph{Random baseline}: randomly choose one setting as the best.
\item \emph{Current best}: choose the setting giving the best result at the stop point.
\item \emph{Predicted best}: Align partial TLCs with predicted $\ellhat$, choose lowest fitted normalizer $L(T)$.
\end{enumerate}

\cref{fig:tuning_tests} shows results for $\lambda$ sweeps at
1.7B/20TPP ($\leftfig$) and 3.3B/30TPP ($\rightfig$).
\emph{Predicted best} achieves negligible loss gaps when stopping
after just 30\% and 10\% of training, respectively. In contrast,
\emph{current best}---used in \citet{almazrouei2023falcon} for LR
tuning---succeeds initially at 3.3B but fails at 1.7B, showing it is
not a general solution. Further experiments are in
\cref{sec:app_early}.


\takeaway{
Collapse enables reliable early stopping. By aligning partial training
curves to a small-scale reference, we can predict final loss $L(T)$
and select hyperparameters after only 10--30\% of training,
substantially reducing tuning compute.}

\section{Related work}\label{sec:related}
\paragraph{Scaling laws and scale-stable dynamics.}
\emph{Neural scaling laws} relate loss (generally obtained from
\emph{separate} training runs) to growth in model, data, and
compute sizes, via power laws
\citep{hestness2017scalinglaws,kaplan2020scaling,henighan2020scaling,hoffmann2022empirical,caballero2022broken,alabdulmohsin2022revisiting}.
To ensure stable training as models scale, parameterizations such as
$\mup$ transfer base hyperparameters across sizes, and yield early
dynamics that are scale-stable
\citep{yang2022mup,vyas2023feature,kalra2023universal}, even
\emph{super-consistent} (in curvature) \citep{noci2024super}.
Observing \emph{suboptimal LRs} under $\mup$ as \emph{data} scales,
recent work has proposed decreasing the LR as a function of
$D$~\citep{shen2024power,bjorck2024scaling}; \citet{bergsma2025power}
unify these techniques as forms of $\tema$ adjustment.
\citet{qiu2025scaling} show that, for compute-optimal ladders, TLCs
collapse after normalization. We build on these threads at LLM scale
while co-scaling width, depth, batch size, and weight decay,
identifying new controls that govern TLC collapse.

\paragraph{LLM loss-curve prediction.}
While \citet{kaplan2020scaling} fit a simple power law to TLCs, recent
papers make loss prediction explicitly
LR-dependent~\citep{tissue2024scaling,luo2025multi,schaipp2025surprising,qiu2025scaling,hong2025provable}.
Complementary to these, we take a timescale-centric view: AdamW
implements an EMA over updates, and the normalized timescale $\tema$
(jointly set by LR, weight decay, and batch size) acts to control an
\emph{implicit batch size}, one that trades bias reduction vs.\ variance
suppression and thereby shapes TLCs. In a noisy-quadratic model
(\cref{sec:app_tema}), we derive an expression for training loss under
a constant LR, and explain why decaying schedules invert the ordering
of TLCs across $\tema$, with deformations remaining scale-invariant
once normalized.

\paragraph{Early stopping, HPO, and monitoring.}
Early-termination and HPO methods extrapolate TLCs or prune trials
\citep{swersky2014freeze,domhan2015speeding,jaderberg2017population,zela2018towards,li2018hyperband,choi2018difficulty,akiba2019optuna},
but typically require many short runs and are not tailored to LLM
pre-training regimes.  Our approach leverages \emph{collapse itself}:
fit a small-scale predictor of normalized TLCs, align in-progress
curves to infer $L(T)$, and select winners by 10-30\% of training.
Operationally, large-scale reports document spikes and divergences
\citep{chowdhery2022palm,zhang2022opt,wortsman2023small,molybog2023theory};
we show collapse residuals provide a quantitative, scale-normalized
early-warning signal and a practical aid for debugging.

\section{Conclusion}\label{sec:conclusion}
At LLM scale, \emph{normalized training loss curves collapse across
model sizes} when three controls align: the AdamW timescale $\tema$,
the tokens-per-parameter ratio (TPP), and the learning-rate
schedule. Empirically, $\tema$ (bias--variance smoothing) and TPP
(power-law improvement rate) set TLC shape, while the schedule phases
these effects. Fixing TPP and setting $\tema$ optimally for that TPP
yields alignment across $\sim$100M--3.9B parameters in our
experiments.

We instantiate this in \textbf{Celerity}: fixed TPP with optimal
$\tema$ produces tight collapse and competitive
accuracy. \emph{Collapse residuals} surface issues early, localize
their onset, and enable safer restarts. A simple predictor for
normalized TLCs (fit at small scale) supports \emph{early stopping} in
HPO: by 10--30\% of training we can select winners and estimate
$L(T)$, saving tuning compute.  For \$1B runs, collapse provides a
valuable reference trajectory: keeping training on track, every step
of the way.

\bibliography{bib_tloss}

@misc{abadi2015tf,
  title = {{TensorFlow}: Large-Scale Machine Learning on Heterogeneous Distributed Systems},
  author = {Mart{\'{\i}}n Abadi and Ashish Agarwal and Paul Barham and Eugene Brevdo and Zhifeng Chen and Craig Citro and Gregory S. Corrado and Andy Davis and Jeffrey Dean and Matthieu Devin and others},
  year = 2016,
  volume = {abs/1603.04467}
}

@inproceedings{alibi,
  title = {Train Short, Test Long: Attention with Linear Biases Enables Input Length Extrapolation},
  author = {Ofir Press and Noah Smith and Mike Lewis},
  year = 2022,
  booktitle = {International Conference on Learning Representations}
}

@misc{chowdhery2022palm,
  title = {{PaLM}: Scaling Language Modeling with Pathways},
  author = {Aakanksha Chowdhery and Sharan Narang and Jacob Devlin and Maarten Bosma and Gaurav Mishra and Adam Roberts and Paul Barham and Hyung Won Chung and Charles Sutton and Sebastian Gehrmann and others},
  year = 2022,
  eprint = {2204.02311},
  archiveprefix = {arXiv},
  primaryclass = {cs.CL}
}

@misc{biderman2023pythia,
  title = {{Pythia}: A Suite for Analyzing Large Language Models Across Training and Scaling},
  author = {Stella Biderman and Hailey Schoelkopf and Quentin Anthony and Herbie Bradley and Kyle O'Brien and Eric Hallahan and Mohammad Aflah Khan and Shivanshu Purohit and USVSN Sai Prashanth and Edward Raff and Aviya Skowron and Lintang Sutawika and Oskar van der Wal},
  year = 2023,
  eprint = {2304.01373},
  archiveprefix = {arXiv},
  primaryclass = {cs.CL}
}

@misc{eval-harness,
  title = {A Framework for Few-shot Language Model Evaluation},
  author = {Gao, Leo and Tow, Jonathan and Biderman, Stella and Black, Sid and DiPofi, Anthony and Foster, Charles and Golding, Laurence and Hsu, Jeffrey and McDonell, Kyle and Muennighoff, Niklas and Phang, Jason and Reynolds, Laria and Tang, Eric and Thite, Anish and Wang, Ben and Wang, Kevin and Zou, Andy},
  year = 2021,
  howpublished = {Zenodo}
}

@misc{rae2022scaling,
  title = {Scaling Language Models: Methods, Analysis \& Insights from Training {Gopher}},
  author = {Jack W. Rae and Sebastian Borgeaud and Trevor Cai and Katie Millican and Jordan Hoffmann and Francis Song and John Aslanides and Sarah Henderson and Roman Ring and Susannah Young and others},
  year = 2022,
  eprint = {2112.11446},
  archiveprefix = {arXiv},
  primaryclass = {cs.CL}
}

@inproceedings{hellaswag,
  title = {{HellaSwag}: Can a Machine Really Finish Your Sentence?},
  author = {Zellers, Rowan  and Holtzman, Ari  and Bisk, Yonatan  and Farhadi, Ali  and Choi, Yejin},
  year = 2019,
  booktitle = {Proceedings of the 57th Annual Meeting of the Association for Computational Linguistics}
}

@misc{hestness2017scalinglaws,
  title = {Deep Learning Scaling is Predictable, Empirically},
  author = {Joel Hestness and Sharan Narang and Newsha Ardalani and Gregory Diamos and Heewoo Jun and Hassan Kianinejad and Md. Mostofa Ali Patwary and Yang Yang and Yanqi Zhou},
  year = 2017,
  eprint = {1712.00409},
  archiveprefix = {arXiv},
  primaryclass = {cs.LG}
}

@article{hoffmann2022empirical,
  title={An empirical analysis of compute-optimal large language model training},
  author={Hoffmann, Jordan and Borgeaud, Sebastian and Mensch, Arthur and Buchatskaya, Elena and Cai, Trevor and Rutherford, Eliza and de Las Casas, Diego and Hendricks, Lisa Anne and Welbl, Johannes and Clark, Aidan and others},
  journal={Advances in Neural Information Processing Systems},
  volume={35},
  year={2022}
}

@misc{touvron2023llama,
  title = {{LLaMA}: Open and Efficient Foundation Language Models},
  author = {Hugo Touvron and Thibaut Lavril and Gautier Izacard and Xavier Martinet and Marie-Anne Lachaux and Timothée Lacroix and Baptiste Rozière and Naman Goyal and Eric Hambro and Faisal Azhar and Aurelien Rodriguez and Armand Joulin and Edouard Grave and Guillaume Lample},
  year = 2023,
  eprint = {2302.13971},
  archiveprefix = {arXiv},
  primaryclass = {cs.CL}
}

@inproceedings{piqa,
  title = {{PIQA}: Reasoning about Physical Commonsense in Natural Language},
  author = {Yonatan Bisk and Rowan Zellers and Ronan Le Bras and Jianfeng Gao and Yejin Choi},
  year = 2020,
  booktitle = {Thirty-Fourth AAAI Conference on Artificial Intelligence}
}

@misc{radford2019gpt2,
  title = {Language Models are Unsupervised Multitask Learners},
  author = {Alec Radford and Jeffrey Wu and Rewon Child and David Luan and Dario Amodei and Ilya Sutskever},
  year = 2019
}

@article{shazeer2020glu,
  title={{GLU} variants improve transformer},
  author={Shazeer, Noam},
  journal={arXiv preprint arXiv:2002.05202},
  year={2020}
}

@article{fedus2022switch,
  title={Switch transformers: Scaling to trillion parameter models with simple and efficient sparsity},
  author={Fedus, William and Zoph, Barret and Shazeer, Noam},
  journal={Journal of Machine Learning Research},
  volume={23},
  number={120},
  pages={1--39},
  year={2022}
}

@article{raffel2020exploring,
  title = {Exploring the Limits of Transfer Learning with a Unified Text-to-Text Transformer},
  author = {Raffel, Colin and Shazeer, Noam and Roberts, Adam and Lee, Katherine and Narang, Sharan and Matena, Michael and Zhou, Yanqi and Li, Wei and Liu, Peter J.},
  year = 2020,
  journal = {Journal of Machine Learning Research}
}

@inproceedings{vaswani2017attention,
  title = {Attention Is All You Need},
  author = {Vaswani, Ashish and Shazeer, Noam and Parmar, Niki and Uszkoreit, Jakob and Jones, Llion and Gomez, Aidan N and Kaiser, {\L}ukasz and Polosukhin, Illia},
  year = 2017,
  booktitle = {Advances in Neural Information Processing Systems}
}

@article{winogrande,
  title = {{WinoGrande}: An Adversarial Winograd Schema Challenge at Scale},
  author = {Sakaguchi, Keisuke and Bras, Ronan Le and Bhagavatula, Chandra and Choi, Yejin},
  year = 2021,
  journal = {Communications of the ACM}
}

@inproceedings{yang2022mup,
  title = {Tuning Large Neural Networks via Zero-Shot Hyperparameter Transfer},
  author = {Yang, Greg and Hu, Edward and Babuschkin, Igor and Sidor, Szymon and Liu, Xiaodong and Farhi, David and Ryder, Nick and Pachocki, Jakub and Chen, Weizhu and Gao, Jianfeng},
  year = 2021,
  booktitle = {Advances in Neural Information Processing Systems}
}

@misc{zhang2022opt,
  title = {{OPT}: Open Pre-trained Transformer Language Models},
  author = {Susan Zhang and Stephen Roller and Naman Goyal and Mikel Artetxe and Moya Chen and Shuohui Chen and Christopher Dewan and Mona Diab and Xian Li and Xi Victoria Lin and Todor Mihaylov and Myle Ott and Sam Shleifer and Kurt Shuster and Daniel Simig and Punit Singh Koura and Anjali Sridhar and Tianlu Wang and Luke Zettlemoyer},
  year = 2022,
  eprint = {2205.01068},
  archiveprefix = {arXiv},
  primaryclass = {cs.CL}
}

@article{touvron2023llama2,
  title={{LLaMA~2}: Open foundation and fine-tuned chat models},
  author={Touvron, Hugo and Martin, Louis and Stone, Kevin and Albert, Peter and Almahairi, Amjad and Babaei, Yasmine and Bashlykov, Nikolay and Batra, Soumya and Bhargava, Prajjwal and Bhosale, Shruti and others},
  journal={arXiv preprint arXiv:2307.09288},
  year={2023}
}

@article{bottou2018optimization,
  title={Optimization methods for large-scale machine learning},
  author={Bottou, L{\'e}on and Curtis, Frank E and Nocedal, Jorge},
  journal={SIAM review},
  volume={60},
  number={2},
  pages={223--311},
  year={2018},
  publisher={SIAM}
}

@article{brown2020language,
  title={Language models are few-shot learners},
  author = {Brown, Tom and Mann, Benjamin and Ryder, Nick and Subbiah, Melanie and Kaplan, Jared D and Dhariwal, Prafulla and Neelakantan, Arvind and Shyam, Pranav and Sastry, Girish and Askell, Amanda and others},
  journal={Advances in Neural Information Processing Systems},
  volume={33},
  pages={1877--1901},
  year={2020}
}

@article{lecun1989optimal,
  title={Optimal brain damage},
  author={LeCun, Yann and Denker, John and Solla, Sara},
  journal={Advances in Neural Information Processing Systems},
  volume={2},
  year={1989}
}

@inproceedings{shazeer2018adafactor,
  title={{Adafactor}: Adaptive learning rates with sublinear memory cost},
  author={Shazeer, Noam and Stern, Mitchell},
  booktitle={International Conference on Machine Learning},
  pages={4596--4604},
  year={2018},
  organization={PMLR}
}

@article{yang2020feature,
  title={Feature learning in infinite-width neural networks},
  author={Yang, Greg and Hu, Edward J},
  journal={arXiv preprint arXiv:2011.14522},
  year={2020}
}

@article{dey2023cerebras,
  title={Cerebras-{GPT}: Open compute-optimal language models trained on the {Cerebras} wafer-scale cluster},
  author={Dey, Nolan and Gosal, Gurpreet and Khachane, Hemant and Marshall, William and Pathria, Ribhu and Tom, Marvin and Hestness, Joel},
  journal={arXiv preprint arXiv:2304.03208},
  year={2023}
}

@article{almazrouei2023falcon,
  title={The {Falcon} series of open language models},
  author={Almazrouei, Ebtesam and Alobeidli, Hamza and Alshamsi, Abdulaziz and Cappelli, Alessandro and Cojocaru, Ruxandra and Debbah, M{\'e}rouane and Goffinet, {\'E}tienne and Hesslow, Daniel and Launay, Julien and Malartic, Quentin and others},
  journal={arXiv preprint arXiv:2311.16867},
  year={2023}
}

@article{wang2024how,
  title={How to set {AdamW}'s weight decay as you scale model and dataset size},
  author={Wang, Xi and Aitchison, Laurence},
  journal={arXiv preprint arXiv:2405.13698},
  year={2024}
}

@article{kingma2014adam,
  title={Adam: A method for stochastic optimization},
  author={Kingma, Diederik P and Ba, Jimmy},
  journal={arXiv preprint arXiv:1412.6980},
  year={2014}
}

@article{lepikhin2020gshard,
  title={Gshard: Scaling giant models with conditional computation and automatic sharding},
  author={Lepikhin, Dmitry and Lee, HyoukJoong and Xu, Yuanzhong and Chen, Dehao and Firat, Orhan and Huang, Yanping and Krikun, Maxim and Shazeer, Noam and Chen, Zhifeng},
  journal={arXiv preprint arXiv:2006.16668},
  year={2020}
}

@article{muennighoff2023scaling,
  title={Scaling data-constrained language models},
  author={Muennighoff, Niklas and Rush, Alexander and Barak, Boaz and Le Scao, Teven and Tazi, Nouamane and Piktus, Aleksandra and Pyysalo, Sampo and Wolf, Thomas and Raffel, Colin A},
  journal={Advances in Neural Information Processing Systems},
  volume={36},
  year={2023}
}

@article{sengupta2023jais,
  title={{Jais} and {Jais}-chat: Arabic-centric foundation and instruction-tuned open generative large language models},
  author={Sengupta, Neha and Sahu, Sunil Kumar and Jia, Bokang and Katipomu, Satheesh and Li, Haonan and Koto, Fajri and Marshall, William and Gosal, Gurpreet and Liu, Cynthia and Chen, Zhiming and others},
  journal={arXiv preprint arXiv:2308.16149},
  year={2023}
}

@article{henighan2020scaling,
  title={Scaling laws for autoregressive generative modeling},
  author={Henighan, Tom and Kaplan, Jared and Katz, Mor and Chen, Mark and Hesse, Christopher and Jackson, Jacob and Jun, Heewoo and Brown, Tom B and Dhariwal, Prafulla and Gray, Scott and others},
  journal={arXiv preprint arXiv:2010.14701},
  year={2020}
}

@article{bi2024deepseek,
  title={{DeepSeek} {LLM}: Scaling open-source language models with longtermism},
  author={Bi, Xiao and Chen, Deli and Chen, Guanting and Chen, Shanhuang and Dai, Damai and Deng, Chengqi and Ding, Honghui and Dong, Kai and Du, Qiushi and Fu, Zhe and others},
  journal={arXiv preprint arXiv:2401.02954},
  year={2024}
}

@article{tissue2024scaling,
  title={Scaling Law with Learning Rate Annealing},
  author={Tissue, Howe and Wang, Venus and Wang, Lu},
  journal={arXiv preprint arXiv:2408.11029},
  year={2024}
}

@article{hagele2024scaling,
  title={Scaling Laws and Compute-Optimal Training Beyond Fixed Training Durations},
  author={H{\"a}gele, Alexander and Bakouch, Elie and Kosson, Atli and Allal, Loubna Ben and Von Werra, Leandro and Jaggi, Martin},
  journal={arXiv preprint arXiv:2405.18392},
  year={2024}
}

@article{defazio2024road,
  title={The Road Less Scheduled},
  author={Defazio, Aaron and Yang, Xingyu {({Alice})} and Mehta, Harsh and Mishchenko, Konstantin and Khaled, Ahmed and Cutkosky, Ashok},
  journal={arXiv preprint arXiv:2405.15682},
  year={2024}
}

@article{hu2024minicpm,
  title={{MiniCPM}: Unveiling the potential of small language models with scalable training strategies},
  author={Hu, Shengding and Tu, Yuge and Han, Xu and He, Chaoqun and Cui, Ganqu and Long, Xiang and Zheng, Zhi and Fang, Yewei and Huang, Yuxiang and Zhao, Weilin and others},
  journal={arXiv preprint arXiv:2404.06395},
  year={2024}
}

@article{defazio2023optimal,
  title={Optimal linear decay learning rate schedules and further refinements},
  author={Defazio, Aaron and Cutkosky, Ashok and Mehta, Harsh and Mishchenko, Konstantin},
  journal={arXiv preprint arXiv:2310.07831},
  year={2023}
}

@article{shen2024power,
  title={Power Scheduler: A Batch Size and Token Number Agnostic Learning Rate Scheduler},
  author={Shen, Yikang and Stallone, Matthew and Mishra, Mayank and Zhang, Gaoyuan and Tan, Shawn and Prasad, Aditya and Soria, Adriana Meza and Cox, David D and Panda, Rameswar},
  journal={arXiv preprint arXiv:2408.13359},
  year={2024}
}

@article{wortsman2023small,
  title={Small-scale proxies for large-scale transformer training instabilities},
  author={Wortsman, Mitchell and Liu, Peter J and Xiao, Lechao and Everett, Katie and Alemi, Alex and Adlam, Ben and Co-Reyes, John D and Gur, Izzeddin and Kumar, Abhishek and Novak, Roman and others},
  journal={arXiv preprint arXiv:2309.14322},
  year={2023}
}

@misc{cerebras2023slimpajama,
author = {Soboleva, Daria and Al-Khateeb, Faisal and Myers, Robert and Steeves, Jacob R and Hestness, Joel and Dey, Nolan},
title = {{SlimPajama}: A {627B} token cleaned and deduplicated version of {RedPajama}},
year = 2023,
howpublished = {\specialhref{https://www.cerebras.net/blog/slimpajama-a-627b-token-cleaned-and-deduplicated-version-of-redpajama}{blue}{Web page}}
}

@misc{dey2023btlm3b8k,
      title={{BTLM-3B-8K}: {7B} Parameter Performance in a {3B} Parameter Model}, 
      author={Nolan Dey and Daria Soboleva and Faisal Al-Khateeb and Bowen Yang and Ribhu Pathria and Hemant Khachane and Shaheer Muhammad and Zhiming and Chen and Robert Myers and Jacob Robert Steeves and Natalia Vassilieva and Marvin Tom and Joel Hestness},
      year={2023},
      eprint={2309.11568},
      archivePrefix={arXiv},
      primaryClass={cs.AI}
}

@inproceedings{press2022alibi,
title={Train Short, Test Long: Attention with Linear Biases Enables Input Length Extrapolation},
author={Ofir Press and Noah Smith and Mike Lewis},
booktitle={International Conference on Learning Representations},
year={2022},
}

@article{loshchilov2016sgdr,
  title={{SGDR}: Stochastic gradient descent with warm restarts},
  author={Loshchilov, Ilya and Hutter, Frank},
  journal={arXiv preprint arXiv:1608.03983},
  year={2016}
}

@article{dubey2024llama,
  title={The {Llama~3} herd of models},
  author={Dubey, Abhimanyu and Jauhri, Abhinav and Pandey, Abhinav and Kadian, Abhishek and Al-Dahle, Ahmad and Letman, Aiesha and Mathur, Akhil and Schelten, Alan and Yang, Amy and Fan, Angela and others},
  journal={arXiv preprint arXiv:2407.21783},
  year={2024}
}

@inproceedings{gupta2018shampoo,
  title={Shampoo: Preconditioned stochastic tensor optimization},
  author={Gupta, Vineet and Koren, Tomer and Singer, Yoram},
  booktitle={International Conference on Machine Learning},
  pages={1842--1850},
  year={2018},
  organization={PMLR}
}

@article{zhang2024how,
  title={How Does Critical Batch Size Scale in Pre-training?},
  author={Zhang, Hanlin and Morwani, Depen and Vyas, Nikhil and Wu, Jingfeng and Zou, Difan and Ghai, Udaya and Foster, Dean and Kakade, Sham},
  journal={arXiv preprint arXiv:2410.21676},
  year={2024}
}

@article{mccandlish2018empirical,
  title = {An Empirical Model of Large-Batch Training},
  author = {Sam McCandlish and Jared Kaplan and Dario Amodei and others},
  journal = {arXiv preprint arXiv:1812.06162},
  year = 2018,
}

@article{kaplan2020scaling,
  title = {Scaling Laws for Neural Language Models},
  author = {Jared Kaplan and Sam McCandlish and Tom Henighan and Tom B. Brown and Benjamin Chess and Rewon Child and Scott Gray and Alec Radford and Jeffrey Wu and Dario Amodei},
  journal = {arXiv preprint arXiv:2001.08361},
  year = 2020,
}

@article{bjorck2024scaling,
  title={Scaling Optimal {LR} Across Token Horizons},
  author={Bjorck, Johan and Benhaim, Alon and Chaudhary, Vishrav and Wei, Furu and Song, Xia},
  journal={arXiv preprint arXiv:2409.19913},
  year={2024}
}

@article{porian2024resolving,
  title={Resolving discrepancies in compute-optimal scaling of language models},
  author={Porian, Tomer and Wortsman, Mitchell and Jitsev, Jenia and Schmidt, Ludwig and Carmon, Yair},
  journal={arXiv preprint arXiv:2406.19146},
  year={2024}
}

@article{besiroglu2024chinchilla,
  title={{Chinchilla} Scaling: A replication attempt},
  author={Besiroglu, Tamay and Erdil, Ege and Barnett, Matthew and You, Josh},
  journal={arXiv preprint arXiv:2404.10102},
  year={2024}
}

@article{krajewski2024scaling,
  title={Scaling laws for fine-grained mixture of experts},
  author={Krajewski, Jakub and Ludziejewski, Jan and Adamczewski, Kamil and Pi{\'o}ro, Maciej and Krutul, Micha{\l} and Antoniak, Szymon and Ciebiera, Kamil and Kr{\'o}l, Krystian and Odrzyg{\'o}{\'z}d{\'z}, Tomasz and Sankowski, Piotr and Cygan, Marek and Jaszczur, Sebastian},
  journal={arXiv preprint arXiv:2402.07871},
  year={2024}
}

@article{zhang2019algorithmic,
  title={Which algorithmic choices matter at which batch sizes? insights from a noisy quadratic model},
  author={Zhang, Guodong and Li, Lala and Nado, Zachary and Martens, James and Sachdeva, Sushant and Dahl, George and Shallue, Chris and Grosse, Roger B},
  journal={Advances in neural information processing systems},
  volume={32},
  year={2019}
}

@article{schaipp2025surprising,
  title={The Surprising Agreement Between Convex Optimization Theory and Learning-Rate Scheduling for Large Model Training},
  author={Schaipp, Fabian and H{\"a}gele, Alexander and Taylor, Adrien and Simsekli, Umut and Bach, Francis},
  journal={arXiv preprint arXiv:2501.18965},
  year={2025}
}

@article{bordelon2023depthwise,
  title={Depthwise hyperparameter transfer in residual networks: Dynamics and scaling limit},
  author={Bordelon, Blake and Noci, Lorenzo and Li, Mufan Bill and Hanin, Boris and Pehlevan, Cengiz},
  journal={arXiv preprint arXiv:2309.16620},
  year={2023}
}

@article{yang2023tensor,
  title={Tensor programs {VI}: Feature learning in infinite-depth neural networks},
  author={Yang, Greg and Yu, Dingli and Zhu, Chen and Hayou, Soufiane},
  journal={arXiv preprint arXiv:2310.02244},
  year={2023}
}

@article{bergsma2025straight,
  title={Straight to zero: Why linearly decaying the learning rate to zero works best for {LLMs}},
  author={Bergsma, Shane and Dey, Nolan and Gosal, Gurpreet and Gray, Gavia and Soboleva, Daria and Hestness, Joel},
  journal={arXiv preprint arXiv:2502.15938},
  year={2025}
}

@article{yang2024qwen2_5,
  title={Qwen2.5 technical report},
  author={Yang, An and Yang, Baosong and Zhang, Beichen and Hui, Binyuan and Zheng, Bo and Yu, Bowen and Li, Chengyuan and Liu, Dayiheng and Huang, Fei and Wei, Haoran and others},
  journal={arXiv preprint arXiv:2412.15115},
  year={2024}
}

@article{yang2025qwen3,
  title={Qwen3 technical report},
  author={Yang, An and Li, Anfeng and Yang, Baosong and Zhang, Beichen and Hui, Binyuan and Zheng, Bo and Yu, Bowen and Gao, Chang and Huang, Chengen and Lv, Chenxu and others},
  journal={arXiv preprint arXiv:2505.09388},
  year={2025}
}

@article{dangelo2024why,
  title={Why do we need weight decay in modern deep learning?},
  author={D'Angelo, Francesco and Andriushchenko, Maksym and Varre, Aditya Vardhan and Flammarion, Nicolas},
  journal={Advances in Neural Information Processing Systems},
  volume={37},
  pages={23191--23223},
  year={2024}
}

@article{kosson2024analyzing,
  title={Analyzing \& Reducing the Need for Learning Rate Warmup in {GPT} Training},
  author={Kosson, Atli and Messmer, Bettina and Jaggi, Martin},
  journal={arXiv preprint arXiv:2410.23922},
  year={2024}
}

@article{vyas2024soap,
  title={{SOAP}: Improving and stabilizing shampoo using adam},
  author={Vyas, Nikhil and Morwani, Depen and Zhao, Rosie and Kwun, Mujin and Shapira, Itai and Brandfonbrener, David and Janson, Lucas and Kakade, Sham},
  journal={arXiv preprint arXiv:2409.11321},
  year={2024}
}

@article{bergsma2025power,
  title={Power Lines: Scaling Laws for Weight Decay and Batch Size in {LLM} Pre-training},
  author={Bergsma, Shane and Dey, Nolan and Gosal, Gurpreet and Gray, Gavia and Soboleva, Daria and Hestness, Joel},
  journal={arXiv preprint arXiv:2505.13738},
  year={2025}
}

@article{feng2024maximize,
  title={Maximize Your Data's Potential: Enhancing {LLM} Accuracy with Two-Phase Pretraining},
  author={Feng, Steven and Prabhumoye, Shrimai and Kong, Kezhi and Su, Dan and Patwary, Mostofa and Shoeybi, Mohammad and Catanzaro, Bryan},
  journal={arXiv preprint arXiv:2412.15285},
  year={2024}
}

@article{olmo2024,
  title={{2 {OLMo} 2 {Furious}}},
  author={OLMo, Team and Walsh, Pete and Soldaini, Luca and Groeneveld, Dirk and Lo, Kyle and Arora, Shane and Bhagia, Akshita and Gu, Yuling and Huang, Shengyi and Jordan, Matt and others},
  journal={arXiv preprint arXiv:2501.00656},
  year={2024}
}

@article{dey2025dont,
  title={Don't be lazy: {CompleteP} enables compute-efficient deep transformers},
  author={Dey, Nolan and Zhang, Bin Claire and Noci, Lorenzo and Li, Mufan and Bordelon, Blake and Bergsma, Shane and Pehlevan, Cengiz and Hanin, Boris and Hestness, Joel},
  journal={arXiv preprint arXiv:2505.01618},
  year={2025}
}

@article{muennighoff2024olmoe,
  title={{OLMoE}: Open mixture-of-experts language models},
  author={Muennighoff, Niklas and Soldaini, Luca and Groeneveld, Dirk and Lo, Kyle and Morrison, Jacob and Min, Sewon and Shi, Weijia and Walsh, Pete and Tafjord, Oyvind and Lambert, Nathan and others},
  journal={arXiv preprint arXiv:2409.02060},
  year={2024}
}

@article{team2025gemma,
  title={Gemma~3 technical report},
  author={Team, Gemma and Kamath, Aishwarya and Ferret, Johan and Pathak, Shreya and Vieillard, Nino and Merhej, Ramona and Perrin, Sarah and Matejovicova, Tatiana and Ram{\'e}, Alexandre and Rivi{\`e}re, Morgane and others},
  journal={arXiv preprint arXiv:2503.19786},
  year={2025}
}

@article{team2024gemma2,
  title={Gemma~2: Improving open language models at a practical size},
  author={Team, Gemma and Riviere, Morgane and Pathak, Shreya and Sessa, Pier Giuseppe and Hardin, Cassidy and Bhupatiraju, Surya and Hussenot, L{\'e}onard and Mesnard, Thomas and Shahriari, Bobak and Ram{\'e}, Alexandre and others},
  journal={arXiv preprint arXiv:2408.00118},
  year={2024}
}

@article{dominguez2024training,
  title={Training on the test task confounds evaluation and emergence},
  author={Dominguez-Olmedo, Ricardo and Dorner, Florian E and Hardt, Moritz},
  journal={arXiv preprint arXiv:2407.07890},
  year={2024}
}

@article{zhang2024map,
  title={{MAP-Neo}: Highly capable and transparent bilingual large language model series},
  author={Zhang, Ge and Qu, Scott and Liu, Jiaheng and Zhang, Chenchen and Lin, Chenghua and Yu, Chou Leuang and Pan, Danny and Cheng, Esther and Liu, Jie and Lin, Qunshu and others},
  journal={arXiv preprint arXiv:2405.19327},
  year={2024}
}

@article{achiam2023gpt,
  title={{GPT-4} Technical Report},
  author={Achiam, Josh and Adler, Steven and Agarwal, Sandhini and Ahmad, Lama and Akkaya, Ilge and Aleman, Florencia Leoni and Almeida, Diogo and Altenschmidt, Janko and Altman, Sam and Anadkat, Shyamal and others},
  journal={arXiv preprint arXiv:2303.08774},
  year={2023}
}

@article{springer2025overtrained,
  title={Overtrained language models are harder to fine-tune},
  author={Springer, Jacob Mitchell and Goyal, Sachin and Wen, Kaiyue and Kumar, Tanishq and Yue, Xiang and Malladi, Sadhika and Neubig, Graham and Raghunathan, Aditi},
  journal={arXiv preprint arXiv:2503.19206},
  year={2025}
}

@article{kumar2024scaling,
  title={Scaling laws for precision},
  author={Kumar, Tanishq and Ankner, Zachary and Spector, Benjamin F and Bordelon, Blake and Muennighoff, Niklas and Paul, Mansheej and Pehlevan, Cengiz and R{\'e}, Christopher and Raghunathan, Aditi},
  journal={arXiv preprint arXiv:2411.04330},
  year={2024}
}

@article{wen2024understanding,
  title={Understanding warmup-stable-decay learning rates: A river valley loss landscape perspective},
  author={Wen, Kaiyue and Li, Zhiyuan and Wang, Jason and Hall, David and Liang, Percy and Ma, Tengyu},
  journal={arXiv preprint arXiv:2410.05192},
  year={2024}
}

@article{roller2021hash,
  title={Hash layers for large sparse models},
  author={Roller, Stephen and Sukhbaatar, Sainbayar and Szlam, Arthur and Weston, Jason},
  journal={Advances in Neural Information Processing Systems},
  volume={34},
  pages={17555--17566},
  year={2021}
}

@misc{paster2023openwebmath,
      title={{OpenWebMath}: An Open Dataset of High-Quality Mathematical Web Text},
      author={Keiran Paster and Marco Dos Santos and Zhangir Azerbayev and Jimmy Ba},
      year={2023},
      eprint={2310.06786},
      archivePrefix={arXiv},
      primaryClass={cs.AI}
}

@misc{li2023starcoder,
      title={{StarCoder}: may the source be with you!}, 
      author={Raymond Li and Loubna Ben Allal and Yangtian Zi and Niklas Muennighoff and Denis Kocetkov and others},
      year={2023},
      eprint={2305.06161},
      archivePrefix={arXiv},
      primaryClass={cs.CL}
}

@misc{benallal2024cosmopedia,
  author = {Ben Allal, Loubna and Lozhkov, Anton and Penedo, Guilherme and Wolf, Thomas and von Werra, Leandro},
  title = {Cosmopedia},
  year = 2024,
  howpublished = {\specialhref{https://huggingface.co/datasets/HuggingFaceTB/cosmopedia}{blue}{Hugging Face}}
}

@article{allal2024smollm,
  title={{SmolLM}-blazingly fast and remarkably powerful},
  author={Allal, Loubna Ben and Lozhkov, Anton and Bakouch, Elie and von Werra, Leandro and Wolf, Thomas},
  journal={Hugging Face Blog},
  volume={16},
  year={2024}
}

@article{allal2025smollm2,
  title={{SmolLM2}: When {Smol} Goes Big--Data-Centric Training of a Small Language Model},
  author={Allal, Loubna Ben and Lozhkov, Anton and Bakouch, Elie and Bl{\'a}zquez, Gabriel Mart{\'\i}n and Penedo, Guilherme and Tunstall, Lewis and Marafioti, Andr{\'e}s and Kydl{\'\i}{\v{c}}ek, Hynek and Lajar{\'\i}n, Agust{\'\i}n Piqueres and Srivastav, Vaibhav and others},
  journal={arXiv preprint arXiv:2502.02737},
  year={2025}
}

@misc{lozhkov2024fineweb-edu,
    author       = { Lozhkov, Anton and Ben Allal, Loubna and von Werra, Leandro and Wolf, Thomas },  
    title        = {{FineWeb-Edu}: the Finest Collection of Educational Content }, 
    year         = 2024,  
    howpublished = {\specialhref{https://huggingface.co/datasets/HuggingFaceFW/fineweb-edu}{blue}{Hugging Face}}
}

@misc{gabarain2024ultra,
  author = {Sebastian Gabarain},
  title = {UltraTextbooks-2.0},
  year = 2024,
  howpublished = {\specialhref{https://huggingface.co/datasets/Locutusque/UltraTextbooks-2.0}{blue}{Hugging Face}}
}

@article{duchi2011adaptive,
  title={Adaptive subgradient methods for online learning and stochastic optimization.},
  author={Duchi, John and Hazan, Elad and Singer, Yoram},
  journal={Journal of Machine Learning Research},
  volume={12},
  number={7},
  year={2011}
}

@misc{kimi2025k2,
  title = {{Kimi K2}: Open agentic intelligence},
  author = {Kimi Team},
  year = {2025},
  url = {https://github.com/MoonshotAI/Kimi-K2/blob/main/tech_report.pdf}
}

@article{liu2023sophia,
  title={Sophia: A scalable stochastic second-order optimizer for language model pre-training},
  author={Liu, Hong and Li, Zhiyuan and Hall, David and Liang, Percy and Ma, Tengyu},
  journal={arXiv preprint arXiv:2305.14342},
  year={2023}
}

@article{noci2024super,
  title={Super consistency of neural network landscapes and learning rate transfer},
  author={Noci, Lorenzo and Meterez, Alexandru and Hofmann, Thomas and Orvieto, Antonio},
  journal={Advances in Neural Information Processing Systems},
  volume={37},
  pages={102696--102743},
  year={2024}
}

@article{qiu2025scaling,
  title={Scaling Collapse Reveals Universal Dynamics in Compute-Optimally Trained Neural Networks},
  author={Qiu, Shikai and Xiao, Lechao and Wilson, Andrew Gordon and Pennington, Jeffrey and Agarwala, Atish},
  journal={arXiv preprint arXiv:2507.02119},
  year={2025}
}

@article{vyas2023feature,
  title={Feature-learning networks are consistent across widths at realistic scales},
  author={Vyas, Nikhil and Atanasov, Alexander and Bordelon, Blake and Morwani, Depen and Sainathan, Sabarish and Pehlevan, Cengiz},
  journal={Advances in Neural Information Processing Systems},
  volume={36},
  pages={1036--1060},
  year={2023}
}

@article{kalra2023universal,
  title={Universal sharpness dynamics in neural network training: Fixed point analysis, edge of stability, and route to chaos},
  author={Kalra, Dayal Singh and He, Tianyu and Barkeshli, Maissam},
  journal={arXiv preprint arXiv:2311.02076},
  year={2023}
}

@misc{devries2023chinchilla_analysis,
  author = {De Vries, Harm},
  title = {Go smol or go home},
  howpublished = {\specialhref{https://www.harmdevries.com/post/model-size-vs-compute-overhead/}{blue}{Blog post}},
  year = {2023}
}

@article{luo2025multi,
  title={A multi-power law for loss curve prediction across learning rate schedules},
  author={Luo, Kairong and Wen, Haodong and Hu, Shengding and Sun, Zhenbo and Liu, Zhiyuan and Sun, Maosong and Lyu, Kaifeng and Chen, Wenguang},
  journal={arXiv preprint arXiv:2503.12811},
  year={2025}
}

@article{glorioso2024zamba2,
  title={The {Zamba2} suite: Technical report},
  author={Glorioso, Paolo and Anthony, Quentin and Tokpanov, Yury and Golubeva, Anna and Shyam, Vasudev and Whittington, James and Pilault, Jonathan and Millidge, Beren},
  journal={arXiv preprint arXiv:2411.15242},
  year={2024}
}

@article{xiao2024rethinking,
  title={Rethinking conventional wisdom in machine learning: From generalization to scaling},
  author={Xiao, Lechao},
  journal={arXiv preprint arXiv:2409.15156},
  year={2024}
}

@article{molybog2023theory,
  title={A theory on {Adam} instability in large-scale machine learning},
  author={Molybog, Igor and Albert, Peter and Chen, Moya and DeVito, Zachary and Esiobu, David and Goyal, Naman and Koura, Punit Singh and Narang, Sharan and Poulton, Andrew and Silva, Ruan and others},
  journal={arXiv preprint arXiv:2304.09871},
  year={2023}
}

@misc{zhang2022logbook,
   author = {Susan Zhang and Stephen Roller and Naman Goyal and Mikel Artetxe and Moya Chen and Shuohui Chen and Christopher Dewan and Mona Diab and Xian Li and others},
   title = {OPT-175 Logbook},
   year = 2022,
   howpublished = {\specialhref{https://github.com/facebookresearch/metaseq/blob/main/projects/OPT/chronicles/OPT175B_Logbook.pdf}{blue}{PDF}}
}

@inproceedings{domhan2015speeding,
  title={Speeding up automatic hyperparameter optimization of deep neural networks by extrapolation of learning curves.},
  author={Domhan, Tobias and Springenberg, Jost Tobias and Hutter, Frank and others},
  booktitle={IJCAI},
  volume={15},
  pages={3460--8},
  year={2015}
}

@article{swersky2014freeze,
  title={Freeze-thaw {Bayesian} optimization},
  author={Swersky, Kevin and Snoek, Jasper and Adams, Ryan Prescott},
  journal={arXiv preprint arXiv:1406.3896},
  year={2014}
}

@article{zela2018towards,
  title={Towards automated deep learning: Efficient joint neural architecture and hyperparameter search},
  author={Zela, Arber and Klein, Aaron and Falkner, Stefan and Hutter, Frank},
  journal={arXiv preprint arXiv:1807.06906},
  year={2018}
}

@article{jaderberg2017population,
  title={Population based training of neural networks},
  author={Jaderberg, Max and Dalibard, Valentin and Osindero, Simon and Czarnecki, Wojciech M and Donahue, Jeff and Razavi, Ali and Vinyals, Oriol and Green, Tim and Dunning, Iain and Simonyan, Karen and others},
  journal={arXiv preprint arXiv:1711.09846},
  year={2017}
}

@inproceedings{hong2025provable,
  title={On the Provable Separation of Scales in Maximal Update Parameterization},
  author={Hong, Letong and Wang, Zhangyang},
  booktitle={Forty-second International Conference on Machine Learning},
  year={2025}
}

@article{li2018hyperband,
  title={Hyperband: A novel bandit-based approach to hyperparameter optimization},
  author={Li, Lisha and Jamieson, Kevin and DeSalvo, Giulia and Rostamizadeh, Afshin and Talwalkar, Ameet},
  journal={Journal of Machine Learning Research},
  volume={18},
  number={185},
  pages={1--52},
  year={2018}
}

@inproceedings{akiba2019optuna,
  title={Optuna: A next-generation hyperparameter optimization framework},
  author={Akiba, Takuya and Sano, Shotaro and Yanase, Toshihiko and Ohta, Takeru and Koyama, Masanori},
  booktitle={Proceedings of the 25th ACM SIGKDD international conference on knowledge discovery \& data mining},
  pages={2623--2631},
  year={2019}
}

@article{caballero2022broken,
  title={Broken neural scaling laws},
  author={Caballero, Ethan and Gupta, Kshitij and Rish, Irina and Krueger, David},
  journal={arXiv preprint arXiv:2210.14891},
  year={2022}
}

@article{alabdulmohsin2022revisiting,
  title={Revisiting neural scaling laws in language and vision},
  author={Alabdulmohsin, Ibrahim M and Neyshabur, Behnam and Zhai, Xiaohua},
  journal={Advances in Neural Information Processing Systems},
  volume={35},
  pages={22300--22312},
  year={2022}
}

@inproceedings{choi2018difficulty,
  title={On the difficulty of {DNN} hyperparameter optimization using learning curve prediction},
  author={Choi, Daeyoung and Cho, Hyunghun and Rhee, Wonjong},
  booktitle={TENCON 2018-2018 IEEE Region 10 Conference},
  pages={0651--0656},
  year={2018},
  organization={IEEE}
}

@article{sculley2015hidden,
  title={Hidden technical debt in machine learning systems},
  author={Sculley, David and Holt, Gary and Golovin, Daniel and Davydov, Eugene and Phillips, Todd and Ebner, Dietmar and Chaudhary, Vinay and Young, Michael and Crespo, Jean-Francois and Dennison, Dan},
  journal={Advances in neural information processing systems},
  volume={28},
  year={2015}
}

@article{song2025through,
  title={Through the river: Understanding the benefit of schedule-free methods for language model training},
  author={Song, Minhak and Baek, Beomhan and Ahn, Kwangjun and Yun, Chulhee},
  journal={arXiv preprint arXiv:2507.09846},
  year={2025}
}

@article{ludziejewski2025joint,
  title={Joint {MoE} Scaling Laws: Mixture of Experts Can Be Memory Efficient},
  author={Ludziejewski, Jan and Pi{\'o}ro, Maciej and Krajewski, Jakub and Stefaniak, Maciej and Krutul, Micha{\l} and Ma{\l}a{\'s}nicki, Jan and Cygan, Marek and Sankowski, Piotr and Adamczewski, Kamil and Mi{\l}o{\'s}, Piotr and others},
  journal={arXiv preprint arXiv:2502.05172},
  year={2025}
}

@article{busbridge2025distillation,
  title={Distillation scaling laws},
  author={Busbridge, Dan and Shidani, Amitis and Weers, Floris and Ramapuram, Jason and Littwin, Etai and Webb, Russ},
  journal={arXiv preprint arXiv:2502.08606},
  year={2025}
}

@article{wang2025distilqwen2,
  title={{DistilQwen2.5}: Industrial Practices of Training Distilled Open Lightweight Language Models},
  author={Wang, Chengyu and Yan, Junbing and Yue, Yuanhao and Huang, Jun},
  journal={arXiv preprint arXiv:2504.15027},
  year={2025}
}

@article{tunstall2023zephyr,
  title={{Zephyr}: Direct distillation of {LM} alignment},
  author={Tunstall, Lewis and Beeching, Edward and Lambert, Nathan and Rajani, Nazneen and Rasul, Kashif and Belkada, Younes and Huang, Shengyi and Von Werra, Leandro and Fourrier, Cl{\'e}mentine and Habib, Nathan and others},
  journal={arXiv preprint arXiv:2310.16944},
  year={2023}
}

@article{meterez2025seesaw,
  title={Seesaw: Accelerating Training by Balancing Learning Rate and Batch Size Scheduling},
  author={Meterez, Alexandru and Morwani, Depen and Wu, Jingfeng and Oncescu, Costin-Andrei and Pehlevan, Cengiz and Kakade, Sham},
  journal={arXiv preprint arXiv:2510.14717},
  year={2025}
}
\bibliographystyle{cereb}

\newpage
\appendix
\section{Limitations and future directions}\label{sec:limitations}
Across our initial (\cref{sec:shape,sec:tuning}) and Celerity
(\cref{sec:celerity}) setups, we have tested collapse across two
distinct settings of architecture (and context length), dataset (and
vocabulary size), and parameterization.  We directly compare TLCs from
these two settings in \cref{sec:app_additional}, while also describing
further experiments in learning-rate schedule ($\constant$ vs.\
$\tenx$ vs.\ $\dtoz$), Adam $\beta_1$/$\beta_2$ parameters, and dense
vs.\ sparse mixture-of-expert (MoE) architectures.  However, in all
cases our results are established under single-epoch pre-training with
AdamW.\ The observed patterns may change under extreme TPP,
multi-epoch training, alternate optimizers/schedules, or heavy
mid-training data annealing/curricula.

\paragraph{Optimizers.}

We hypothesize the optimizer timescale will remain a primary control
of TLC shape for other optimizers with decoupled weight decay (e.g.,
Sophia~\citep{liu2023sophia}, MuonClip~\citep{kimi2025k2}) whose
update rules can be expressed in EMA form analogous to AdamW
(\cref{sec:background}).  Likewise, the $\tema$ perspective should
also hold when AdamW is applied in alternate weight bases, e.g., as in
SOAP~\citep{vyas2024soap}, where AdamW is applied in Shampoo's
eigenbasis~\citep{gupta2018shampoo}.
Extending a timescale analysis to optimizers without a natural EMA
form (e.g., Adagrad~\citep{duchi2011adaptive},
Adafactor~\citep{shazeer2018adafactor}, SGD variants) is an important
future direction.

\paragraph{Data curricula.}

Given the growing use of data curricula and late-stage data annealing
in LLM pre-training, it is valuable to study how shifts in data affect
TLC shape across scales.  Collapse may also \emph{inform} curriculum
design by serving as a transfer marker.  For example, observing
limited opportunities for experimentation at large
scale, \citet{feng2024maximize} experiment at smaller scales
with \emph{downsampled} datasets that simulate the repetition
occurring at larger sizes (due to limited high-quality
tokens). Consistency in TLC shape could serve as an indicator of
whether the downsample proportions reflect a consistent
overfitting/generalization trade-off across scales.  Collapse can thus
serve to confirm that smaller-scale settings provide suitable proxies
for optimizing data mixes and other settings.

\paragraph{Celerity extensions.}

Beyond choosing TPP (controlling placement on the cost/compression
curve; \cref{fig:compression}), we aim to understand which factors or
training strategies \emph{shift the curve itself}. For faster
inference, we are especially interested in the location of
the \emph{parameter wall}---the minimal capacity achieving a target
loss---and how architecture (dense vs.\ MoE), routing, and depth/width
changes affect collapse and efficiency.

We intentionally chose dense models for our initial Celerity series
because dense models have fewer confounding factors (e.g., routing
strategy, number of experts), making them simpler to study and build
upon.  In our own practice, algorithmic innovations are typically
validated on dense models first.  However, we are interested in
scaling MoE-variants of our Celerity series due to their documented
savings in training
compute~\citep{krajewski2024scaling,ludziejewski2025joint}.

\paragraph{Train loss vs.\ generalization.}

We focus on \emph{training} loss because (i) it is FLOPs-free to
monitor, (ii) in LLM pre-training it typically tracks validation under
stationary data, and (iii) it surfaces issues earlier (e.g.,
duplicated segments), enabling targeted intervention before held-out
degradation. Late-stage annealing and domain shift can decouple
train-loss collapse from downstream behavior. We study validation
collapse for Celerity in \cref{sec:val_loss}; future work should also
consider \emph{downstream-collapse}, and measure
train$\leftrightarrow$val$\leftrightarrow$downstream correlations
across schedules and data mixtures.

\paragraph{Predictive model and schedules.}

Both collapse itself, and our predictive model's ability to accurately
forecast normalized TLCs, is impaired by loss spikes and divergences,
which move the normalized curve away from the universal trajectory
(sometimes temporarily, sometimes for extended periods).  From one
perspective, this is a feature not a bug, as the resulting collapse
anomalies provide a useful mechanism for detecting training issues
(discussed further below).

Empirically, dividing by the final training loss ($\hat L{=}0$)
aligned curves best; future work will study why irreducible-loss
offsets, as in \citet{qiu2025scaling}, were not beneficial.  In terms
of our predictive model, next steps include factoring LR envelope vs.\
anneal-phase effects (cf.~\citep{tissue2024scaling,luo2025multi}),
adding uncertainty (e.g., seed bootstraps) and uncertainty-aware
early-stopping policies.  Given that in our experiments, the
parametric predictor was fit for one specific LR schedule ($\dtoz$),
we should also revisit whether $b(\tema)$, $q(\tpp)$, and possibly $m$
vary systematically across cosine~\citep{loshchilov2016sgdr}, inverse
square-root~\citep{vaswani2017attention,raffel2020exploring,shen2024power},
and warmup-stable-decay
(WSD)~\citep{hu2024minicpm,hagele2024scaling,wen2024understanding,song2025through}
schedules, and schedule-free schemes~\citep{defazio2024road}.

It would also be interesting to measure collapse when batch size
schedules are used.  Theoretically, we could maintain collapse by
adjusting weight decay whenever batch size changes, maintaining the
$\tema$ invariant.  We could compare such adjustments to adjustments
of LR, e.g., as in~\citet{meterez2025seesaw}.

\paragraph{Systems effects and making collapse a practice.}

Collapse residuals are sensitive to systems
effects---microbatching/accumulation, precision, kernel changes,
restarts---which can create artifacts or reveal true pathologies. To
pay down ``hidden technical debt''~\citep{sculley2015hidden}, we
advocate a lightweight \emph{collapse monitor}: log fraction-of-data
in addition to raw step count (easy to add in
TensorBoard~\citep{abadi2015tf}), as well as microbatch statistics and
restart boundaries; normalize online and alert when residuals exceed
policy thresholds. Treating collapse as an operational invariant
reduces configuration fragility and surfaces data/numerics issues
early.

\section{Explaining TLC shape: further details}

\subsection{Full experimental details}\label{sec:experimental_details}

\begin{table}
  \centering
  \caption{Model architectures used in \cref{sec:shape} and \cref{sec:tuning}.\label{tab:old_model_info}}
\begin{tabular}{@{}ccccc@{}}
\toprule
Model & $\dmodel$ & $\nlayers$ & $\dffn$ & $\dhead$ \\ \midrule
111M  & 768    & 10 &  2048 & 64           \\
266M  & 768    & 32 &  2048 & 64           \\
610M  & 2048   & 10 &  5461 & 64           \\
1.7B  & 2048   & 32 &  5461 & 64           \\
3.3B  & 2048   & 64 &  5461 & 64           \\
\bottomrule
\end{tabular}
\end{table}

\begin{table}
  \centering
  \caption{Models, tokens-per-parameter (TPP) and corresponding
    dataset sizes (in tokens), number of model variants trained (over
    LR schedule type, $\eta$, $\lambda$, $B$) for models used in
    \cref{sec:shape} and \cref{sec:tuning}.  In total, $\approx$600
    TLCs were analyzed.\label{tab:old_data_sizes}}
\begin{tabular}{@{}cccc@{}}
  \toprule
Model & TPP & $D$ & Variants trained \\ \midrule
111M  & 20  & 2.19B  & 74 \\
111M  & 80  & 8.76B  & 50 \\
111M  & 200 & 21.9B  & 28 \\
111M  & 320 & 35.0B  & 40 \\
111M  & 1280 & 140.1B & 11 \\
266M  & 20  & 5.31B  & 25 \\
266M  & 80  & 21.2B  & 19 \\
266M  & 320 & 85.0B  & 19 \\
266M  & 1280 & 339.8B & 3 \\
610M  & 20  & 12.1B  & 205 \\
610M  & 80  & 48.5B  & 53 \\
610M  & 200 & 121.3B & 14 \\
610M  & 320 & 194.1B & 5 \\
1.7B  & 20  & 34.3B  & 31 \\
1.7B  & 80  & 137.2B & 11 \\
1.7B  & 160 & 274.3B & 1 \\
1.7B  & 320 & 548.6B & 1 \\
3.3B  & 20  & 66.5B  & 2 \\
3.3B  & 23  & 76.5B  & 1 \\
3.3B  & 30  & 99.8B  & 5 \\
\bottomrule
\end{tabular}
\end{table}

In this section, we provide details on the model architecture
(\cref{tab:old_model_info}) and training data
(\cref{tab:old_data_sizes}) for models used in experiments in
\cref{sec:shape}, \cref{sec:tuning}, and elsewhere in the appendix.
Experimental details for the Celerity model series are in
\cref{sec:celerity_details}.

In total, $\approx$600 TLCs were analyzed for these experiments.
All such models were GPT2-style LLMs~\citep{radford2019gpt2}
with ALiBi~\citep{press2022alibi} embeddings and
SwiGLU~\citep{shazeer2020glu} non-linearity.
We use the AdamW optimizer.  Following standard practice, we do not
apply weight decay or bias to LayerNorm layers.
Default AdamW settings are $\beta_1 = 0.9$, $\beta_2 = 0.95$, and
$\epsilon = 1$e$-8$.
We report cross-entropy loss.
%
%
We parameterize with maximal update parameterization,
$\mup$~\citep{yang2022mup}, with hyperparameters set via proxy tuning,
as described below.
For a given TPP, all models have the exact same warmup phase: a linear
warmup of the learning rate from 0 to the maximum value.
In all runs, warmup was 10\% of the total steps.
Learning rate warmup is standard practice in LLM
training~\citep{brown2020language,rae2022scaling,biderman2023pythia,dubey2024llama,kosson2024analyzing}.

Note \cref{fig:tema} (and later \cref{tab:celerity_model_arch},
\cref{fig:hptuning}, \cref{fig:prediction}, and
\cref{fig:tuning_tests}) report batch size in \emph{sequences} rather
than \emph{tokens}.

All models in the experiments were trained on a Cerebras CS-3 system.
610M-parameter 20TPP models take roughly 6 hours each to train on a
single CS-3.

\paragraph{Proxy model hyperparameter tuning.}\label{sec:proxy_tuning}

\begin{table}
    \centering
    \caption{Tuned hyperparameters for $\mup$ proxy model for models
      used in \cref{sec:shape} and \cref{sec:tuning}.\label{tab:old_mup_hps}}
    \begin{tabular}{cc}
         \toprule
         $\sigma_{W,\text{base}}$& $8.67$e-$02$ \\
         $\hateta$& $\maxlrdetail$\\
         $\alpha_{\text{input}}$& $9.17$\\
         $\alpha_{\text{output}}$& $1.095$\\
         \bottomrule
    \end{tabular}
\end{table}

To find optimal $\mup$ hyperparameters (HPs), we trained a 39M proxy
model using a width $\dproxy$ of 256, with 24 layers and head size of
64.  We trained this model on 800M tokens with $B$=256 sequences and a
context length 2048.  We randomly sampled 350 configurations of base
learning rates, base initialization standard deviation, and embedding
and output logits scaling factors, and used the top-performing values
as our tuned HPs (\cref{tab:old_mup_hps}).

It is worth noting that the LR values reported in this paper and
shown in figures are base $\mup$ LRs before $\mup$-adjustment.
Calculation of $\tema$ (\cref{sec:background}) requires the adjusted
LR (i.e., multiplying by $\dproxy / \dmodel$).  Also, when LR decay is
used, reported LR values always refer to the peak/max LR of the LR
schedule.

\subsection{Explaining TLC dependence on TPP}\label{sec:app_tpp}

Schedules with decaying LR reach their minimum value only at the final
step (after $D$ tokens).  However, for a constant LR schedule, every
step of training is equivalent to a complete training run ending at
that step.  \citet{qiu2025scaling} make the observation that therefore
the loss at every training fraction $\trainfrac = t/T \in [0,1]$
should respect the same fitted scaling law, but for a training budget
of $\trainfrac \cdot D$ tokens.

Starting from the Chinchilla functional form $L(N,D) = E + \nconst
N^{-\alpha} + \dconst D^{-\beta}$, assume that we train with a
constant LR schedule, training until a certain final
tokens-per-parameter ratio $k = D/N$.  At every fraction of training
$\trainfrac$, we will have trained for an intermediate TPP of
$\trainfrac \cdot k$, i.e., using $\trainfrac \cdot k \cdot N$ total
tokens.
To arrive at scale invariance, we note that
\citet{hoffmann2022empirical} found their fitted model and dataset
exponents $\alpha$ and $\beta$ were roughly equal; this rough equality
has also been repeatedly validated in replication
studies~\citep{besiroglu2024chinchilla,porian2024resolving}.  Using $a
= \alpha = \beta$, and focusing on the reducible loss, we obtain a
final training loss of:
\begin{align}
  L(N,k \cdot N) &= \nconst N^{-a} + \dconst (k \cdot N)^{-a} \notag \\
  &= \nconst N^{-a} + \dconst k^{-a} N^{-a}
\end{align}

Meanwhile, training for training fraction $\trainfrac$, the predicted
loss is
\begin{equation}
  L(N,\trainfrac \cdot k \cdot N) = \nconst N^{-a} + \dconst {\trainfrac}^{-a} k^{-a} N^{-a}
\end{equation}

We now normalize by the final loss to expose the shape of the training
loss curve.  The resulting normalized loss $L(N,\trainfrac \cdot k
\cdot N)/L(N,k \cdot N)$ is independent of model (and dataset) size,
depending only on the training fraction $\trainfrac$ and the target
TPP ratio $k$:
\begin{equation}
  \ell(\trainfrac, k) = \frac{\nconst + \dconst {\trainfrac}^{-a} k^{-a}}
      {\nconst + \dconst k^{-a}}
\end{equation}
In other words, for TLCs using a constant LR schedule, collapse
approximately holds under this normalization.  Scaling of $\ell$ in
$\trainfrac^{-a}$ also motivates our own TLC predictive form
(\cref{eq:prediction}).

As shown in \cref{sec:compression}, $a = \alpha = \beta$ implies there
is a single optimal TPP ratio $r$, and moreover, that the Chinchilla
coefficients obey $B = A r^a$.  For a given training run, suppose that
the TPP at which we train is a multiple of the optimal TPP by the
ratio $v$, e.g., $k = v \cdot r$.  Thus, $v = 1$ corresponds to
optimal TPP, while $v>1$ corresponds to overtraining.
We can reparameterize the $\ell$ equation in terms of $v$ as:
\begin{equation}
  \ell(\trainfrac, v) = \frac{1 + v^{-a} \trainfrac^{-a}}{1 + v^{-a}}
\end{equation}

This simple equation clarifies how the overtraining factor $v$
influences the shape of the TLCs.  When $v$ is small (undertraining),
the power law term dominates, and the TLC gradually decays in
$\trainfrac$.  When $v$ is large (overtraining), the power law only
plays a role for smaller $\trainfrac$, the curve drops quickly and
then flattens to $\ell = 1$.  Intuitively, for overtrained models, we
make gains quickly at the beginning of training and then obtain
diminishing returns as training progresses.

\citet{qiu2025scaling} further show that for non-uniform LR schedules,
the loss curve is deformed by $\eta(\trainfrac)$, but, given
consistent curvature of the loss landscape across model scales under
$\mup$~\citep{noci2024super}, the noise-induced deformation is
invariant to model size, and thus collapse still holds.

\subsection{Explaining TLC dependence on $\tema$}\label{sec:app_tema}
As noted in \cref{sec:shape}, the AdamW timescale $\tema =
1/(\eta\lambda T)$ controls the effective memory length of the
parameter updates: smaller $\tema$ emphasizes recent updates (bias
reduction), while larger $\tema$ averages more broadly (variance
reduction).  In this sense, $\tema$ acts as an implicit batch size.

To provide further insight into the role of $\tema$ in shaping TLCs,
we now derive an analytical expression for training loss under a
constant learning rate, using a simple noisy quadratic model (NQM).
While LLM training minimizes cross-entropy loss, it is common to
perform a local quadratic approximation, i.e., a second-order Taylor
expansion in the parameters, with the constant Hessian replaced by the
instantaneous Hessian along the training
trajectory~\citep{lecun1989optimal}.  Thus conclusions drawn from
quadratic models often generalize to large, realistic
networks~\citep{zhang2019algorithmic}.

\paragraph{Setup.}
Following \citet{zhang2019algorithmic}, we assume the optimizer
dynamics are invariant to rotation and translation, allowing us to
model, without loss of generality, a locally quadratic loss, separable
across dimensions, and having an optimum at zero.  Specifically, we
consider a single quadratic mode with curvature $h>0$, optimum at
$\theta^\star=0$, and parameters $\theta_t$, where $t$ is the step
index:
\begin{equation}
L(t)= \tfrac{1}{2}\,h\,\theta_t^2.
\end{equation}
With AdamW optimization, $\theta_t$ evolves as an exponential moving
average (EMA) of stochastic updates $x_t$ with constant smoothing
$\alpha=\eta\lambda$ (\cref{sec:shape}):
\begin{equation}
\theta_t = (1-\alpha)\,\theta_{t-1} + \alpha\,x_{t-1}.
\end{equation}
Unrolling the recurrence gives the general form
\begin{equation}
\theta_t \;=\; (1-\alpha)^t\,\theta_0 \;+\; \sum_{i=0}^{t-1}(1-\alpha)^{t-1-i}\,\alpha\,x_i.
\end{equation}
The first term is the (decaying) contribution of the initialization,
while the second term reflects the accumulation of stochastic
updates.

\paragraph{Continuous (training-fraction) limit.}
We now switch to fractional time $\trainfrac = t/T$ and define the
AdamW timescale $\tau = 1/(\alpha T)$. Approximating
$(1-\alpha)^{t-1-i} \approx e^{-\alpha(t-1-i)}$ and interpreting the
sum as a Riemann approximation as $T \to \infty$, we obtain
\begin{equation}
\theta(\trainfrac) \;\approx\; e^{-\trainfrac/\tau}\,\theta(0)
\;+\; \frac{1}{\tau}\int_{0}^{\trainfrac}
e^{-(\trainfrac-s)/\tau}\,x(s)\,ds.
\end{equation}
That is, $\theta$ consists of two contributions:
a decaying memory of the initialization, and a convolution of the
update signal $x(s)$ with an exponential kernel of timescale $\tau$
(an EMA filter over updates).

\paragraph{Noise model.}
Following \citet{zhang2019algorithmic}, we model the update signal
$x(\trainfrac)$ as preconditioned white noise: a zero-mean process
with constant variance $\sigma_x^2$ and no temporal correlation,
\[
\E[x(\trainfrac)] = 0,\qquad
\E\!\left[x(\trainfrac)\,x(s)\right] = \sigma_x^2\,\delta(\trainfrac-s).
\]
This idealized assumption isolates the effect of $\tau$ by removing
structure in gradient noise beyond its overall scale.

\paragraph{Mean and variance.}
The EMA filter preserves initialization, which decays exponentially:
\[
\E[\theta(\trainfrac)] \;=\; e^{-\trainfrac/\tau}\,\theta(0).
\]
The variance from stochastic updates is
\begin{equation}
\Var[\theta(\trainfrac)] \;=\;
\frac{\sigma_x^2}{2\tau}\,\Big(1 - e^{-2\trainfrac/\tau}\Big).
\end{equation}
Thus the total second moment is
\[
\E[\theta(\trainfrac)^2] \;=\;
e^{-2\trainfrac/\tau}\,\theta(0)^2 \;+\;
\frac{\sigma_x^2}{2\tau}\Big(1 - e^{-2\trainfrac/\tau}\Big).
\]

In words, the initialization bias decays away on timescale $\tau$,
while variance from noisy updates accumulates toward a floor
proportional to $1/\tau$.

\paragraph{Expected loss.}
The per-mode loss is
\[
L(\trainfrac) = \tfrac{1}{2} h\,\theta(\trainfrac)^2.
\]
Taking expectation, and using the decomposition into bias and variance,
\begin{equation}
\E[L(\trainfrac)] \;=\; \tfrac{1}{2} h \left(
    e^{-2\trainfrac/\tau}\,\theta(0)^2
    + \frac{\sigma_x^2}{2\tau}\Big(1 - e^{-2\trainfrac/\tau}\Big)
\right).
\label{eq:loss_general}
\end{equation}
The first term reflects exponentially decaying initialization bias,
while the second reflects variance accumulation to a floor
proportional to $1/\tau$.

If initialization is zero-mean in expectation
($\E[\theta(0)^2]=0$), the bias term vanishes and the expression
simplifies to
\begin{equation}
\boxed{\;\;
\E[L(\trainfrac)] \;=\; \frac{h\,\sigma_x^2}{4\tau}\,
\Big(1 - e^{-2\trainfrac/\tau}\Big)
\;\;}
\label{eq:constantlr}
\end{equation}
which captures the characteristic \emph{fast-then-flatten} TLC shape
under a constant learning rate.

\paragraph{Interpretation.}
Equation~\ref{eq:loss_general} decomposes the expected loss into
an exponentially decaying \emph{bias} term ($\propto e^{-2\trainfrac/\tau}\theta(0)^2$)
and a \emph{variance} term that rises to a floor ($\propto 1/\tau$).
This yields two opposing effects of $\tau$ on TLCs:
\begin{itemize}
\item Smaller $\tau$ suppresses initialization bias more rapidly
  (via the $e^{-2\trainfrac/\tau}$ decay), but accumulates higher variance,
  yielding a higher asymptotic loss floor ($\propto 1/\tau$).
\item Larger $\tau$ reduces variance more effectively, lowering the final loss,
  but is slower to eliminate bias from initialization.
\end{itemize}
When initialization is zero-mean in expectation, the bias term vanishes
and the expression reduces to Eq.~\ref{eq:constantlr}.

\begin{figure}[ht]
  \centering
  \begin{minipage}{0.33\textwidth}
    \includegraphics[trim={0.3cm 0.42cm 0.264cm 0.3cm}, clip, width=\linewidth]{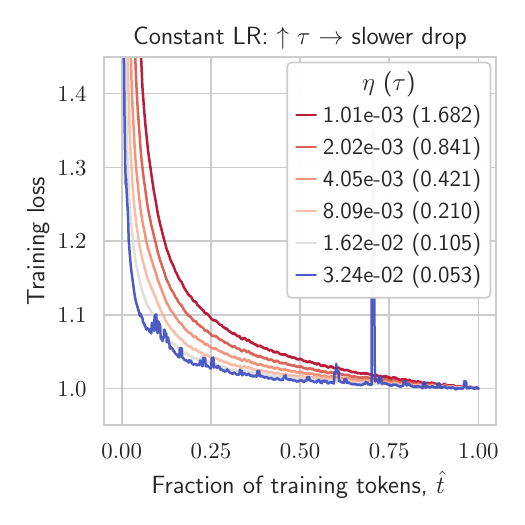}
  \end{minipage}\hfill
  \begin{minipage}{0.33\textwidth}
    \includegraphics[trim={0.3cm 0.42cm 0.264cm 0.3cm}, clip, width=\linewidth]{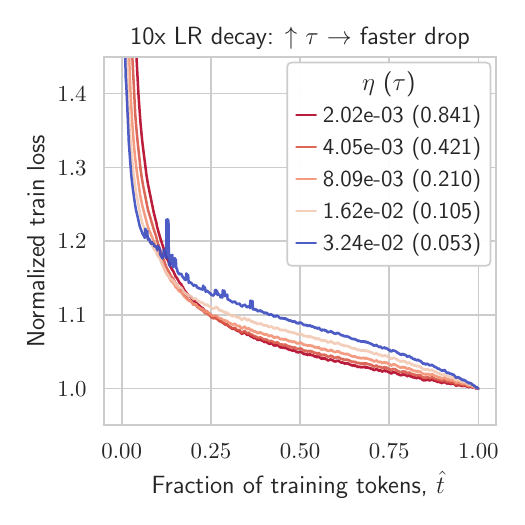}
  \end{minipage}\hfill
  \begin{minipage}{0.33\textwidth}
    \includegraphics[trim={0.3cm 0.42cm 0.264cm 0.3cm}, clip, width=\linewidth]{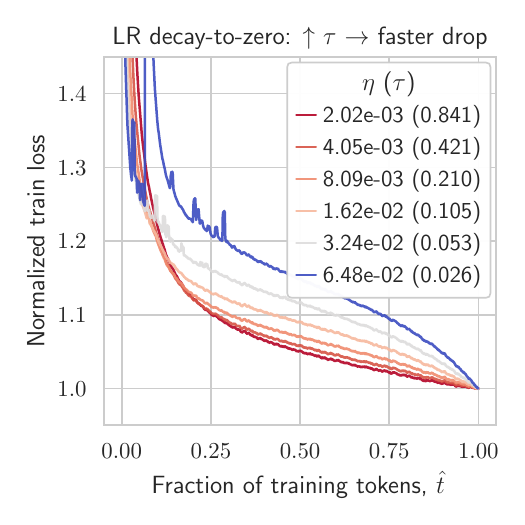}
  \end{minipage}
  \caption{\textbf{Effect of LR schedule on TLC shape (610M, 80TPP).}
    $\Leftfig$: $\constant$ LR, $\Middlefig$: $\linear$
    $10\times$ decay, $\Rightfig$: $\linear$ decay-to-zero.
    Different schedules deform the TLCs in distinct ways, yet in all
    cases the AdamW timescale $\tema$ governs the bias-variance
    trade-off. With a $\constant$ LR, smaller $\tema$ accelerates early
    loss reduction. With $\dtoz$, the effect inverts: smaller $\tema$
    yields a larger late-stage drop. Although here $\tema$ is varied
    by changing LR, equivalent effects arise when varying weight decay
    or batch size (\cref{fig:tema}), confirming $\tema$ as a unifying
    control knob for TLC shape.\label{fig:lrsched}}
\end{figure}

This interpretation matches our empirical findings for normalized
$\constant$-LR TLCs (\cref{fig:lrsched}, $\leftfig$). The situation is
different, however, for LR decay schedules, which we discuss next.

Finally, $\tau$ is a \emph{normalized} timescale and thus invariant to the
absolute number of steps. In the NQM, after normalizing by the final loss
$L(1)$ the curvature factor $h$ cancels exactly. The normalized curve takes
the form
\[
\frac{L(\trainfrac)}{L(1)} \;=\;
\frac{\big(1-e^{-2\trainfrac/\tau}\big)+\kappa\,e^{-2\trainfrac/\tau}}
     {\big(1-e^{-2/\tau}\big)+\kappa\,e^{-2/\tau}},
\qquad
\kappa \;=\; \frac{2\tau\,\E[\theta(0)^2]}{\sigma_x^2}.
\]
Thus, when the initialization contribution is negligible by the end of training
(or when the ratio $\kappa$ is approximately scale-invariant), the normalized
TLC depends only on $(\tau,\trainfrac)$, and curves at matched $\tau$ collapse
across model sizes. If $\kappa$ varies across scales, small early deviations
can appear (bias-dominated regime) but typically diminish as $e^{-2\trainfrac/\tau}$
decays.

\footnotesize\emph{Remark.} \citet{qiu2025scaling} observed collapse without AdamW\@.
Empirically, as $\lambda\!\to\!0$, TLCs approach a limiting shape:
vanilla Adam behaves like AdamW with $\lambda\!=\!0$ (effectively
$\tau\!=\!\infty$).
\normalsize

\paragraph{Extension to decaying LR schedules.}
The constant-LR analysis in Eq.~\ref{eq:constantlr} shows that $\tau$
sets the trade-off: smaller $\tau$ accelerates early bias reduction
but saturates at a higher variance-driven floor, while larger $\tau$
reduces variance more slowly but to a lower asymptote. With a decaying
LR schedule, the smoothing $\alpha_t=\eta_t\lambda$ \emph{decreases}
after warmup, so the instantaneous timescale $\tau_t=1/(\eta_t\lambda
T)$ \emph{increases} as training progresses.  In this setting,
small-$\tau$ runs still make rapid early progress (fast bias
reduction), but during the decay phase they gain additional variance
suppression as $\tau_t$ lengthens, often producing a noticeable
late-stage drop in loss. By contrast, large-$\tau$ runs emphasize
variance reduction throughout, yielding steadier curves without the
same end-of-training acceleration. Equivalently, in the EMA view,
decay flattens the contribution coefficients $c_{t,i}$, averaging over
more (earlier) updates near the end. Thus LR decay effectively
combines the early bias-reducing dynamics of small $\tau$ with the
late variance-reducing dynamics of large $\tau$, inverting the TLC
ordering observed under constant LR (\cref{fig:lrsched}).

This analysis aligns with \citet{bergsma2025straight}, who attribute
the effectiveness of $\dtoz$ schedules to balancing early bias
reduction with later variance suppression (building on
\citealp{dangelo2024why}). Their treatment is primarily conceptual;
here we show how the same bias--variance dynamics manifest directly in
TLC shapes and provide a simple analytical form under the NQM.\@

\subsection{Additional TLC experiments}\label{sec:app_additional}

\begin{figure}[ht]
  \centering
  \begin{minipage}{0.33\textwidth}
    \includegraphics[trim={0.3cm 0.42cm 0.264cm 0.3cm}, clip, width=\linewidth]{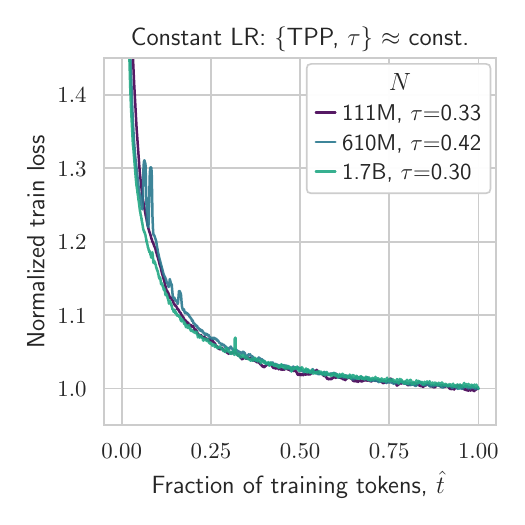}
  \end{minipage}\hfill
  \begin{minipage}{0.33\textwidth}
    \includegraphics[trim={0.3cm 0.42cm 0.264cm 0.3cm}, clip, width=\linewidth]{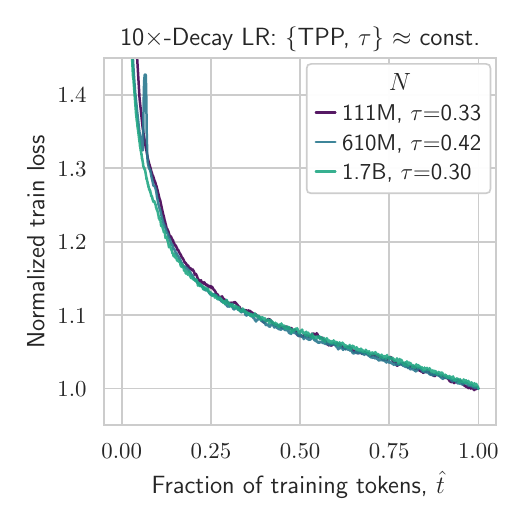}
  \end{minipage}\hfill
  \begin{minipage}{0.33\textwidth}
    \includegraphics[trim={0.3cm 0.42cm 0.264cm 0.3cm}, clip, width=\linewidth]{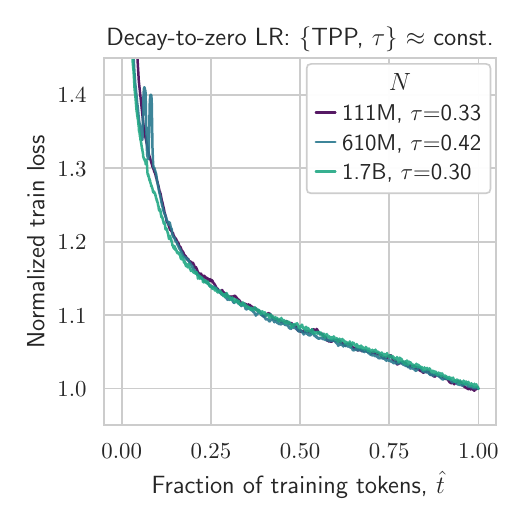}
  \end{minipage}
  \caption{\textbf{Collapse in different LR schedules.}  $\Leftfig$:
    $\constant$ LR, $\Middlefig$: $\linear$ $10\times$ decay,
    $\Rightfig$: $\linear$ decay-to-zero.
    In contrast to \cref{fig:lrsched}, where $\tema$ varies, here
    TPP=20 and $\tema \approx 0.3$: curves collapse across
    scales.\label{fig:lrsched2}}
\end{figure}

\paragraph{Collapse under alternative LR schedules.}
\Cref{fig:lrsched2} shows that normalized TLCs also collapse under a
$\constant$ schedule, a $\tenx$ decay schedule, and our decay-to-zero
schedule (all with 10\% warmup). At corresponding model sizes, we use
the same batch size, peak LR, and weight decay, so same-size results
across schedules differ only in their final LR.\@ Collapse is slightly
looser than in the Celerity runs because the resulting $\tema$ is not
matched exactly across schedules (see plot annotations), but the
qualitative agreement is strong. These results are consistent with our
analysis in \cref{sec:app_tema} and echo the cross-schedule findings
of \citet{qiu2025scaling}.

\paragraph{Collapse across datasets and architectures.}
\begin{figure}[ht]
  \centering
  \begin{minipage}{0.33\textwidth}
    \includegraphics[trim={0.3cm 0.42cm 0.264cm 0.3cm}, clip, width=\linewidth]{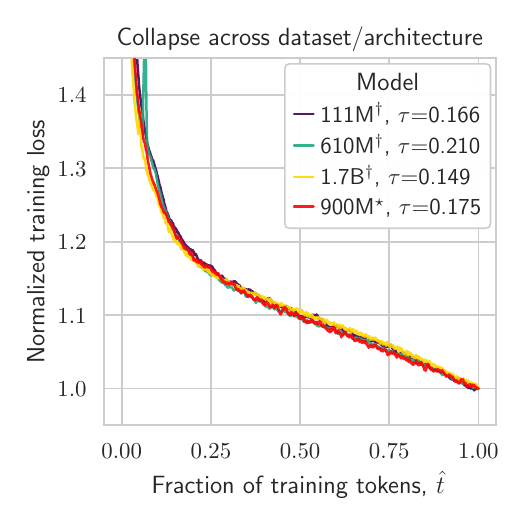}
    \captionsetup{width=.9\linewidth}
    \captionof{figure}{\textbf{Collapse across $^{\dagger}$original, $^{\star}$Celerity setups.} 20 TPP.}
    \label{fig:powercomp}
  \end{minipage}\hfill
  \begin{minipage}{0.33\textwidth}
    \includegraphics[trim={0.3cm 0.42cm 0.264cm 0.3cm}, clip, width=\linewidth]{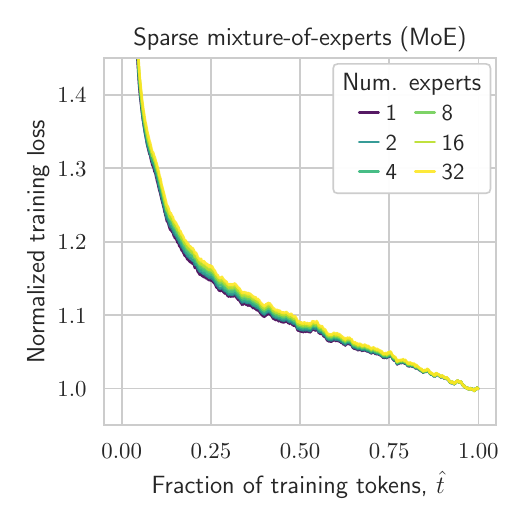}
    \captionsetup{width=.9\linewidth}
    \captionof{figure}{\textbf{Collapse as $E$ varies in a sparse MoE.} 111M, $\tema=0.33$, 20 TPP.}
    \label{fig:moe}
  \end{minipage}\hfill
  \begin{minipage}{0.33\textwidth}
    \includegraphics[trim={0.3cm 0.42cm 0.264cm 0.3cm}, clip, width=\linewidth]{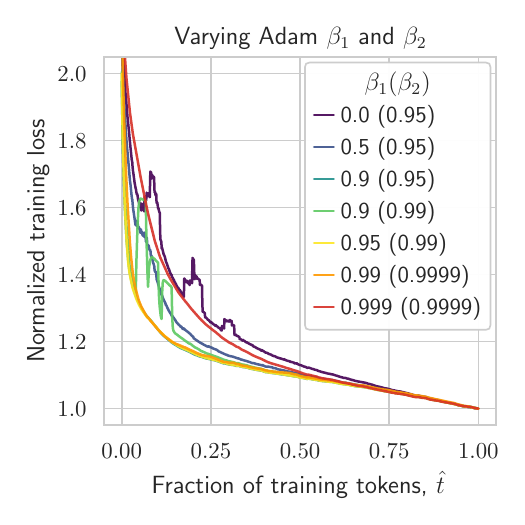}
    \captionsetup{width=.9\linewidth}
    \captionof{figure}{\textbf{Collapse as Adam $\beta_1$ and $\beta_2$
        vary.} 610M, $\tema=0.21$, 20 TPP.}
    \label{fig:betas}
  \end{minipage}
\end{figure}

TLC shape can in principle depend on task, data, and architecture (e.g.,
multi-epoch training on a small corpus can yield faster apparent
improvement than single-epoch pre-training). We therefore ask: how much
does normalized TLC shape change as we vary
parameterization, vocabulary size, architecture, context length, and
dataset mix?

As a first probe, we compare \emph{Celerity} TLCs to our earlier
non-Celerity runs, at the same TPP (20) and similar
$\tema\!\approx\!0.2$, while varying all items above: Celerity uses
CompleteP (vs.\ vanilla $\mup$), a larger vocabulary, different
nonlinearity and FFN multiplier, 4$\times$ longer context, and a
different data mixture
(\cref{sec:experimental_details,sec:celerity_details}). Despite these
differences, the TLCs loosely collapse (\cref{fig:powercomp}). The
Celerity 900M model tracks closer to the 610M model than to the 1.7B
model, although its $\tema$ is intermediate between these
two. Overall, we view this as evidence that the normalized TLC shape
is surprisingly robust when LR schedule, TPP, and $\tema$ are held
(approximately) fixed.

\paragraph{Collapse in sparse mixture-of-experts (MoE).}
We next analyze sparse MoE architectures, where only a subset of
parameters are active per token
\citep{lepikhin2020gshard,fedus2022switch}. Starting from our 111M
dense model (\cref{sec:experimental_details}), we replace each FFN
with a sparse MoE layer and vary the number of experts
$E\!\in\!\{1,2,4,8,16,32\}$.  Tokens are routed to one expert via hash
routing \citep{roller2021hash} so each expert processes a similar
token count. Global training tokens and datasets are identical across
$E$, hence the \emph{effective} TPP per expert decreases from $20$
(dense) to $20/E$ as $E$ grows.
Note also that as the number of experts $E$ increases, and the
effective tokens per expert decrease proportionally, both the expert's
effective batch size $B$ and effective dataset size $D$ are reduced by
a factor of $E$. Since $\tau = B/(\eta \lambda D)$, these reductions
cancel, leaving the overall timescale unchanged (for fixed
$\eta,\lambda$).

\Cref{fig:moe} shows that lower $E$ (higher effective TPP per expert)
yields slightly earlier drops and slightly flatter tails, broadly
obeying the TPP effect characterized in \cref{sec:shape}. Thus, the
observed deformation is explained by effective TPP rather than
differing training dynamics per se.

\paragraph{Collapse across Adam $\beta_1$ and $\beta_2$.}
Finally, we vary $(\beta_1,\beta_2)$ at fixed LR, batch size, and
weight decay ($\tema=0.21$, 610M model, 20~TPP). The default
$(0.9,0.95)$ gives the lowest absolute loss in this experiment, but
several ``standard'' settings---$(0.9,0.95)$, $(0.95,0.99)$, and even
$(0.99,0.9999)$---produce normalized TLCs that collapse
(\cref{fig:betas}). In contrast, runs with $(0.0,0.95)$, $(0.5,0.95)$,
and a noisy instance of $(0.9,0.99)$ exhibit early loss spikes; when
the loss fails to recover promptly, the curves remain elevated and do
not rejoin the main trajectory, breaking collapse (early loss spikes
also distort early collapse for noisy, large-batch-size runs, e.g.,
\cref{fig:hptuning}). We also observe that $(0.999,0.9999)$, which
aggregates gradients over a much longer horizon, follows a
systematically slower (but eventually convergent)
trajectory---consistent with an enlarged momentum timescale
prioritizing variance reduction over bias, akin to increasing $\tema$.

Overall, aside from extreme momentum settings or instability-induced
spikes, setting of $(\beta_1,\beta_2)$ has limited effect on the
\emph{shape} of normalized TLCs. The AdamW timescale $\tema$ remains
the dominant optimization-based control for TLC trajectories.

\takeaway{Normalized TLCs are strikingly robust: they largely collapse
  across diverse datasets and architectures, remain predictable under
  sparse MoE routing (scaling in \emph{effective} TPP as theory
  suggests), and are insensitive to typical Adam $\beta_1,\beta_2$
  settings.  Apart from pathological loss spikes, the dominant factor
  shaping TLCs is still the AdamW timescale $\tau$.}

\section{Celerity models: further details}

\subsection{Compute cost as a function of model compression}\label{sec:compression}

Starting from a compute-optimal model size, we now derive an
expression for the extra compute required ($C/\copt$) to compress a
model to a smaller (less efficient) size, while maintaining the
\emph{same loss}.  We use the resulting equation to plot the
compression vs.\ cost trade-off in \cref{fig:compression}. This analysis
motivated the selection of max TPP in the Celerity model series.

We begin again with the Chinchilla functional form from
\citet{hoffmann2022empirical}, giving loss $L$ as a function of model
size $N$ and data size $D$:
\begin{equation}\label{eq:chinchilla_app}
L(N,D) = E + \nconst N^{-\alpha} + \dconst D^{-\beta}
\end{equation}
where $E$, $\nconst$, $\alpha$, $\dconst$, and $\beta$ are parameters
to be fit on observed training runs.

\citet{hoffmann2022empirical} asked, for a fixed training compute
budget $C$ (in FLOPs), how should we allocate model size $N$ versus
number of training tokens $D$ in order to minimize \emph{loss}?  From
\cref{eq:chinchilla_app}, they derived functions for loss-optimal
$\nopt(C)$ and $\dopt(C)$ (constraining $L(N,D)$ by the common
approximation $C \approx 6 N D$):
\begin{equation}\label{eq:opts}
  \nopt(C) = G\left(\frac{C}{6}\right)^{\frac{\beta}{\alpha+\beta}} \text{and }
  \dopt(C) = G^{-1}\left(\frac{C}{6}\right)^{\frac{\alpha}{\alpha+\beta}},
\end{equation}
where $G = \left(\frac{\alpha \nconst}{\beta \dconst}\right)^{\frac{1}{\alpha + \beta}}$.
Results indicated that $\nopt$ and $\dopt$ scale roughly equally as
$C$ increases. This analysis agreed with their other methods for
estimating compute-optimal scaling, and guided their $N$ and $D$
allocation for training their large-scale Chinchilla model.

\emph{Let $\rstar$ be the optimal $\dopt(C)/\nopt(C)$ ratio.}  If
$\rstar$ is roughly independent of $C$, this implies $\alpha \approx
\beta$.  Using $a = \alpha = \beta$, we obtain:
\begin{equation}
    \rstar = \left(\frac{\dconst}{\nconst}\right)^{\frac{1}{a}},
\end{equation}
or equivalently $\dconst = \nconst \rstar^{a}$.  

Replication studies have found $\alpha \approx \beta \approx 0.35$,
and an optimal TPP of around $\rstar =
20$~\citep{besiroglu2024chinchilla,porian2024resolving} (as noted in
\cref{sec:background}).

Now, suppose $a = \alpha = \beta$ and we obtain a loss of $\lhat$ at the optimal TPP ratio (where $\dopt = \rstar \nopt$):
\begin{align}\label{eq:lhat}
    \lhat &= E + \nconst \nopt^{-\alpha} + \dconst \dopt^{-\beta} \notag \\
    &= E + \nconst \nopt^{-a} + (\nconst \rstar^{a})(\rstar \nopt)^{-a} \notag \\
    &= E + 2 \nconst \nopt^{-a}
\end{align}

We now wish to train a \emph{compressed} model with fraction $k_N$ of
parameters compared to $\nopt$, but obtaining the same loss.  Let $N=
k_N \nopt$.  If $N < \nopt$, we will need $k_D$ extra tokens compared
to $\dopt$ in order to reach the loss target. Let $D = k_D \dopt$.
Rather than training at $\rstar$ TPP, we will train at a higher ratio
$(k_D \dopt)/(k_N \nopt) = (k_D /k_N)\rstar$.  From
\cref{eq:chinchilla_app}, and following a similar derivation
to~\citet{devries2023chinchilla_analysis}, the estimated loss will be:
\begin{equation}\label{eq:lossprime}
L(N, D) = E + \nconst (k_N \nopt)^{-\alpha} + \dconst (k_D \dopt)^{-\beta}
\end{equation}

Again substituting $a = \alpha = \beta$ and $\dconst = \nconst r^a$,
to obtain the target loss $\lhat$, we set the loss in
\cref{eq:lossprime} to equal $\lhat$ in \cref{eq:lhat}, and solve for
$k_D$, finding:
\begin{equation}
    k_D = (2 - {k_N}^{-a})^{\frac{-1}{a}}
\end{equation}

The compute cost $C$ of the compressed training will be $6 (k_N
\nopt)(k_D \dopt)$, from which we can derive the extra compute ratio
compared to $\copt = 6\nopt\dopt$:
\begin{align}\label{eq:c_ratio}
    C/\copt &= k_N k_D \notag \\
    &= k_N (2 - {k_N}^{-a})^{\frac{-1}{a}}
\end{align}
\cref{eq:c_ratio} allows us to vary $k_N$ and obtain the corresponding
compute overhead.  When planning the Celerity training runs, we
assumed $r = 20$ corresponded to the compute-optimal model size
(following the Chinchilla rule-of-thumb) and we tested different
values of $a$ reported in prior work, using $a = 0.35$ in
\cref{fig:compression}.

\subsection{Celerity Recipe Details}\label{sec:celerity_details}

In this section, we provide further details for the techniques that
most impacted Celerity's performance and compute efficiency, including
parameterization, learning rate and weight decay scheduling,
architecture, and dataset construction.

\begin{figure}[ht]
  \centering
  \includegraphics[width=0.55\textwidth]{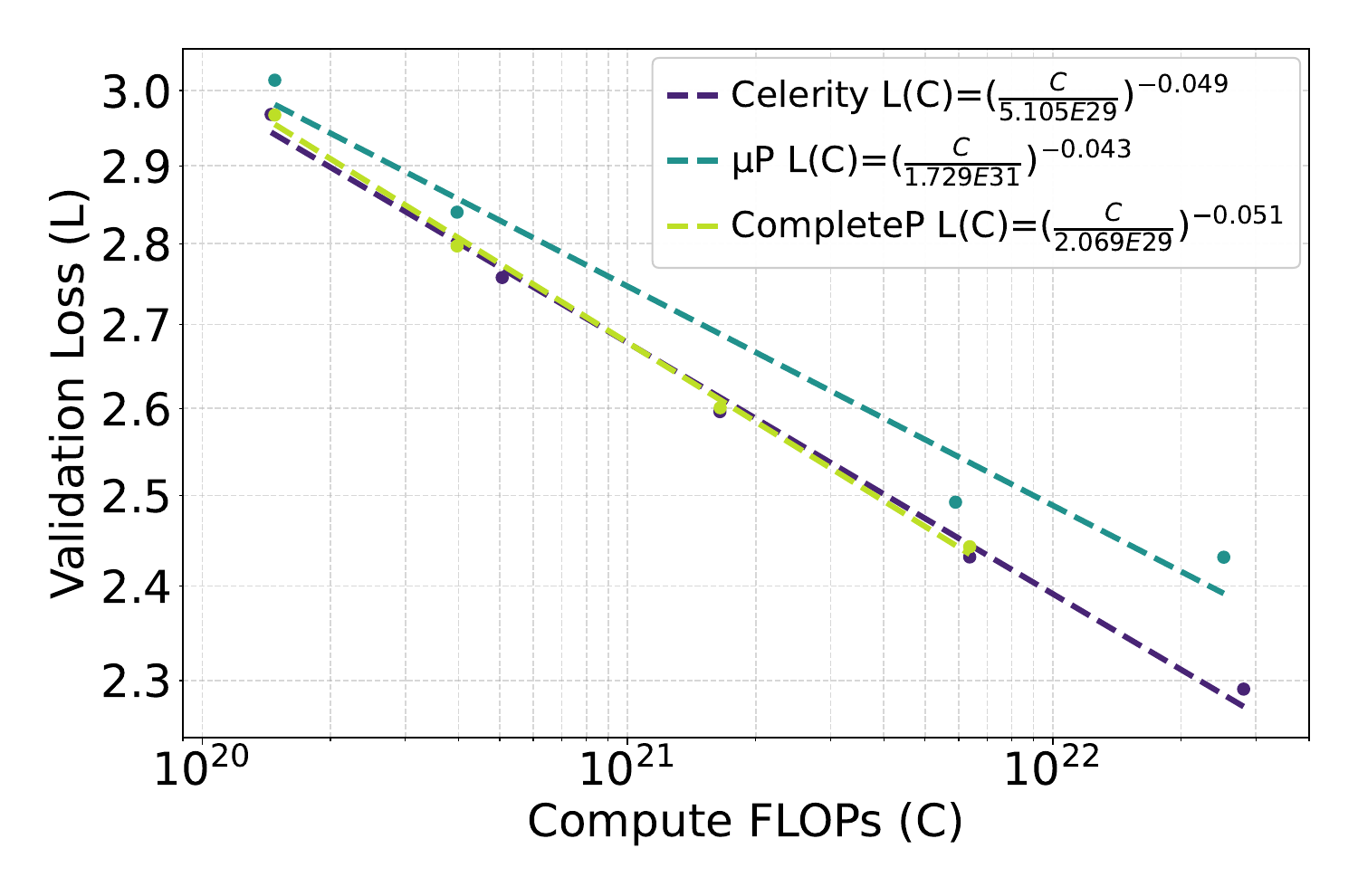}
  \caption{Scaling law comparison between CompleteP and
    $\mu$P. CompleteP scales more predictably: the power law is a
    better fit to the observed points.  CompleteP also exhibits a
    steeper slope, improving loss faster in compute FLOPs, likely due
    to both better HP transfer (across model width \emph{and} depth),
    and better compute efficiency.}
  \label{fig:parameterization}
\end{figure}

\paragraph{Parameterization.} We compare the effect of different
parameterizations and their influence on compute efficiency in
\cref{fig:parameterization}. Specifically, we compare $\mup$
\citep{yang2022mup}, which accounts for scaling in \emph{width}, and
CompleteP \citep{dey2025dont}, which accounts for scaling in
\emph{both width and depth}. For each parameterization, the
hyperparameters (HP), such as learning rate, weight initialization,
and multipliers, are tuned at depth 32 and then directly applied to
the target model training.

Here we note two observations: the first observation is that $\mup$
points do not align well on a standard scaling law.  We attribute this
to HP de-tuning when transferring HPs from proxy model depth to target
model depth.  Such de-tuning is not seen in the scaling laws for
CompleteP and Celerity (which uses CompleteP but a different dataset),
where the points align with minimal error on the scaling law.
\citet{dey2023cerebras} showed poor scaling law fits when scaling
width in the \emph{standard parameterization} vs.\ $\mup$; we believe
a similar phenomenon is now happening when scaling in \emph{depth}.

The second observation is that CompleteP is more compute efficient
than $\mup$, which prior work has explained through the lens of
feature learning~\citep{dey2025dont}.  Indeed, Fig.~4 in
\citet{dey2025dont} suggest that CompleteP models exhibit better
scaling behavior than $\mup$, even when both are comprehensively
tuned.

Based on these observations, we used CompleteP for training the
Celerity series, using the proxy model's tuned HPs across all scales.
Sample code for implementing CompleteP is available at
\url{https://github.com/EleutherAI/nanoGPT-mup/tree/completep}.

\paragraph{Learning Rate and Weight Decay.} We chose the linear decay-to-zero
($\dtoz$) learning rate schedule based on its empirical success and
conceptual motivations in \citet{bergsma2025straight}.  In particular,
\citet{bergsma2025straight} showed that as TPP increases beyond
compute-optimal 20~TPP, the relative benefit of $\dtoz$ also
increases, in a scale-invariant manner.  This makes $\dtoz$
particularly appropriate for parameter-efficient training (e.g.,
Celerity's 234~TPP model band).  All models also train with linear
warmup to the peak LR, over the minimum of 10\%-of-total-tokens or
375M-tokens.

We tuned $\tema$ at a smaller scale and smaller TPP, and transferred
across TPP using the power law fit from \citet{bergsma2025power}.
Given learning rate is determined by CompleteP, and batch size is
optimized according to a separate scaling rule (described below), we
adjusted weight decay in order to obtain the desired $\tema$ setting
at each scale.

\paragraph{Batch Size.} In early experimentation, the batch sizes were
chosen such that they were around the critical batch size
\citep{mccandlish2018empirical}. Later we used the insights from
\citet{bergsma2025power} and started following the rule $\bopt \propto
D^{0.5}$, tuning $B$ at a small scale and then inferring optimal batch
sizes on larger datasets via the power law.

\begin{table}[ht]
\centering
\caption{Composition of the Celerity pre-training dataset.\label{tab:celerity_datasets}}
\begin{tabular}{lccccc}
    \toprule
    Data Subset & Percentage (\%) \\
    \midrule
    FineWeb-Edu \citep{lozhkov2024fineweb-edu} & 64.75 \\
    StarCoder \citep{li2023starcoder} & 10.8 \\
    Cosmopedia \citep{benallal2024cosmopedia} & 4.66 \\
    SlimPajama \citep{cerebras2023slimpajama} & 17.49 \\
    OpenWebMath \citep{paster2023openwebmath} & 1.88 \\
    UltraTextBooks-2.0 \citep{gabarain2024ultra} & 0.42 \\
    \bottomrule
\end{tabular}
\end{table}

\paragraph{Data Selection.} Over the course of experiments, we found that
adding more \emph{refined} data, particularly educational, math, and
coding datasets, generally helps the models score higher on common
benchmarks. In \cref{tab:celerity_datasets}, we break down the
datasets used for Celerity model training, including the proportion
assigned to each subset. A large portion of the datasets are focused
on educational materials, math, and coding.
We use only the \emph{non--web-crawled} subsets of
SlimPajama~\citep{cerebras2023slimpajama}, i.e., excluding C4 and
CommonCrawl, which are effectively replaced by FineWeb-Edu.
As noted in
\cref{sec:celerity}, we do not schedule the data sources, i.e., we do
not employ a data curriculum in the training of Celerity, nor do we
include (benchmark) task-specific data in Celerity training.

\begin{table}[ht]
\centering
\caption{Comparison of Celerity models trained on different datasets.\label{tab:celerity_data_comp}}
\resizebox{\textwidth}{!}
{
\begin{tabular}{lccccccccccc}
    \toprule
    \multirow{1}{*}{Name} & \multicolumn{8}{c}{Downstream Accuracy (Num Shots)}\\
    & arc-c & arc-e & boolq & hellaswag & piqa & siqa & winogrande & Avg.\\
    & (25) & (0) & (0) & (10) & (0) & (0) & (5) & \\
    \midrule
    Celerity 300M & 27.82 & 50.63 & 52.75 & 37.57 & 66.21 & 37.77 & 52.25 & \textbf{46.43} \\
    Celerity 300M SlimPJ & 24.32 & 42.17 & 61.53 & 36.04 & 65.56 & 37.97 & 50.99 & 45.51 \\
    \midrule
    Celerity 900M & 39.68 & 64.52 & 47.92 & 55.02 & 72.03 & 41.97 & 58.48 & \textbf{54.23} \\
    Celerity 900M SlimPJ & 30.89 & 54.67 & 55.47 & 53.74 & 71.00 & 40.89 & 57.46 & 52.02 \\
    \bottomrule
\end{tabular}
}
\end{table}

Comparison in \cref{tab:celerity_data_comp} shows that the same model
configurations trained on a general dataset like SlimPajama result in
worse downstream performance compared to the Celerity data mix. While
dataset optimization was not a focus of Celerity, these results do
underscore the importance of dataset composition in pre-training.
This also makes clear why hyperscalers invest a tremendous amount of
work into data preparation, synthesis, filtering, and refinement.

\begin{table}
  \centering
  \caption{Models, tokens-per-parameter and corresponding dataset
    sizes (in tokens) for Celerity.\label{tab:celerity_data_sizes}}
\begin{tabular}{@{}ccc@{}}
  \toprule
Model & TPP & $D$     \\ \midrule
300M  & 20  & 5.4B   \\
300M  & 80  & 21.7B   \\
300M  & 234 & 63.4B   \\
500M  & 20  & 10.1B   \\
500M  & 80  & 40.2B   \\
500M  & 234 & 117.8B   \\
900M  & 20  & 18.1B   \\
900M  & 80  & 72.5B   \\
900M  & 234 & 212.3B   \\
1.8B  & 20  & 36.2B   \\
1.8B  & 80  & 144.8B   \\
1.8B  & 234 & 424.0B   \\
3.9B  & 20  & 77.6B   \\
3.9B  & 80  & 310.4B   \\
3.9B  & 234 & 909.2B   \\
\bottomrule
\end{tabular}
\end{table}

\cref{tab:celerity_data_sizes} summarizes the dataset sizes for all
models in the Celerity model series.

\paragraph{Model Architecture.} Celerity models use a decoder-only
GPT2-style transformer architecture. \cref{tab:celerity_model_arch}
summarizes the architecture dimensions, hyperparameters, and other
details of the Celerity model family. We trained five Celerity model
sizes from scratch with parameter counts roughly 300M, 500M, 900M,
1.8B, and 3.9B. All models are trained under consistent data and
optimization methods, on public datasets, in order to foster open
science and fair comparison.

\subsection{Celerity further results}

\begin{figure}[ht]
  \centering
    \includegraphics[width=0.58\linewidth]{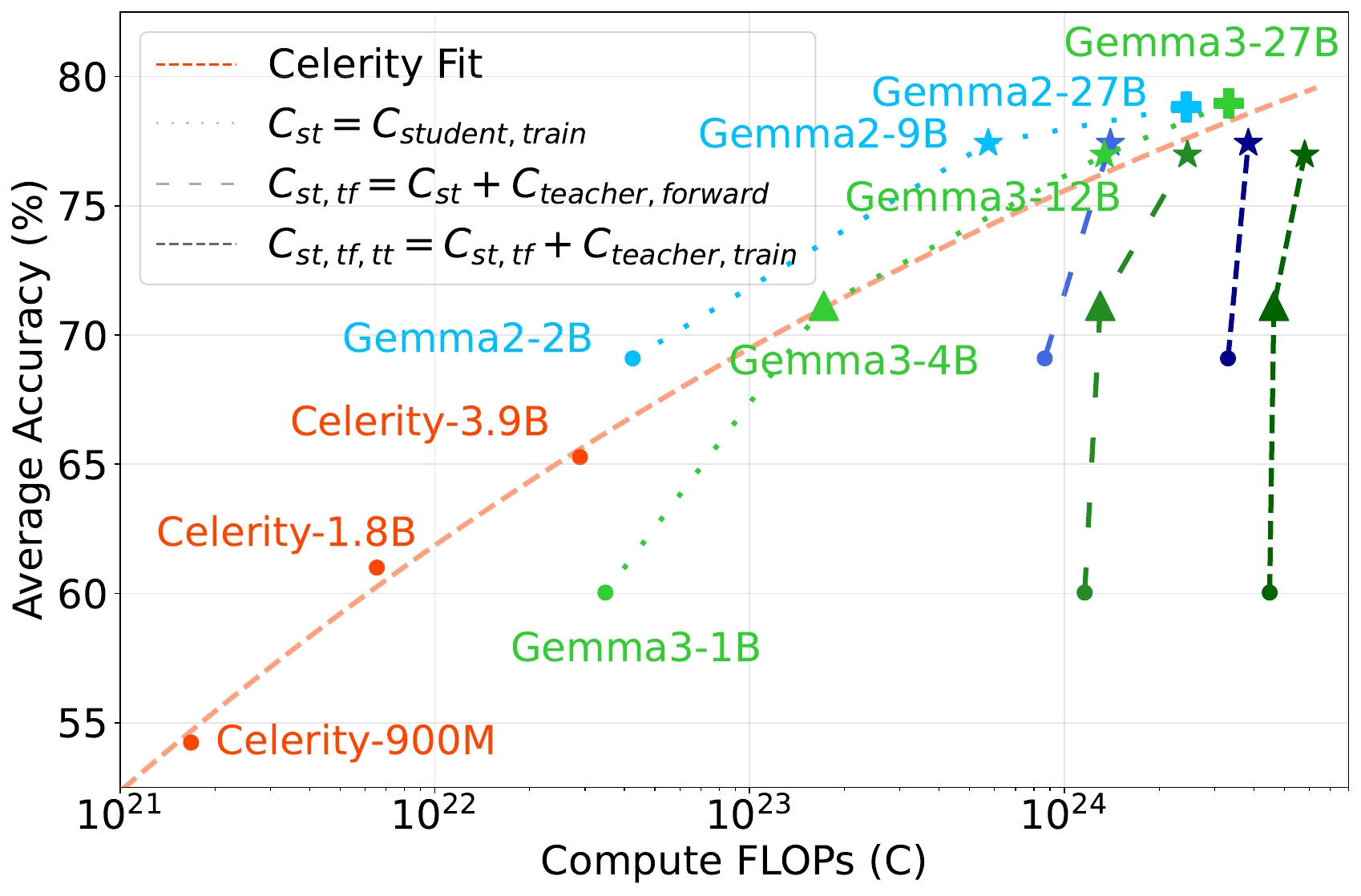}
    \caption{\textbf{Celerity compute efficiency vs.\ distilled
        models:} Downstream accuracy.  Celerity models perform
      similarly to \emph{distilled} Gemma-2/Gemma-3 models, when
      generously only accounting for distillation student FLOPs.  When
      considering \emph{teacher} forward pass FLOPs, Gemma curves
      shift away from Pareto frontier (worse), with a further shift if
      we account for FLOPs to \emph{train} the
      teacher.\label{fig:celerity_compute2}}
\end{figure}

In our empirical evaluation of Celerity, we necessarily only compare
to model families with sufficiently precise and complete training
details, in particular the total training tokens.
E.g., Llama-3.2 reports using ``up to'' 9T tokens
(\href{https://console.cloud.google.com/vertex-ai/publishers/meta/model-garden/llama3-2}{Llama-3.2
  Overview}) while Llama-3.1 and Qwen-3 sizes are reported
``approximately.''  Moreover, it is unclear from the papers whether
the full pre-training datasets (i.e., used to train the flagship
models) were also used for the smaller models.
For an approximate sense of how these models compare, we include
Llama-3 and Qwen models in \cref{tab:all_evals} based on the
assumption the models do use the full (approximately-reported)
datasets.  Note that if these assumptions hold, the smaller Llama-3
and Qwen-3 models would be rather compute-inefficient, e.g., they
would all be beyond the plotting range of \cref{fig:celerity_compute1}
(i.e., $>10^{24}$ FLOPs) and score well below the Celerity extrapolation
(and Gemma results) in \cref{fig:celerity_compute2}.

\paragraph{Compute efficiency.}

\cref{fig:celerity_compute2} provides further downstream results for
Celerity models (and their fitted extrapolation), in comparison to
larger Gemma-2 and Gemma-3 models.  The plot shows how the accuracy
vs.\ FLOPs comparison depends on whether we account for teaching FLOPs
(e.g., generating logits for student training), or the initial cost of
educating the teacher.

\paragraph{Token efficiency.}

\begin{figure}[ht]
  \centering
    \includegraphics[width=0.58\linewidth]{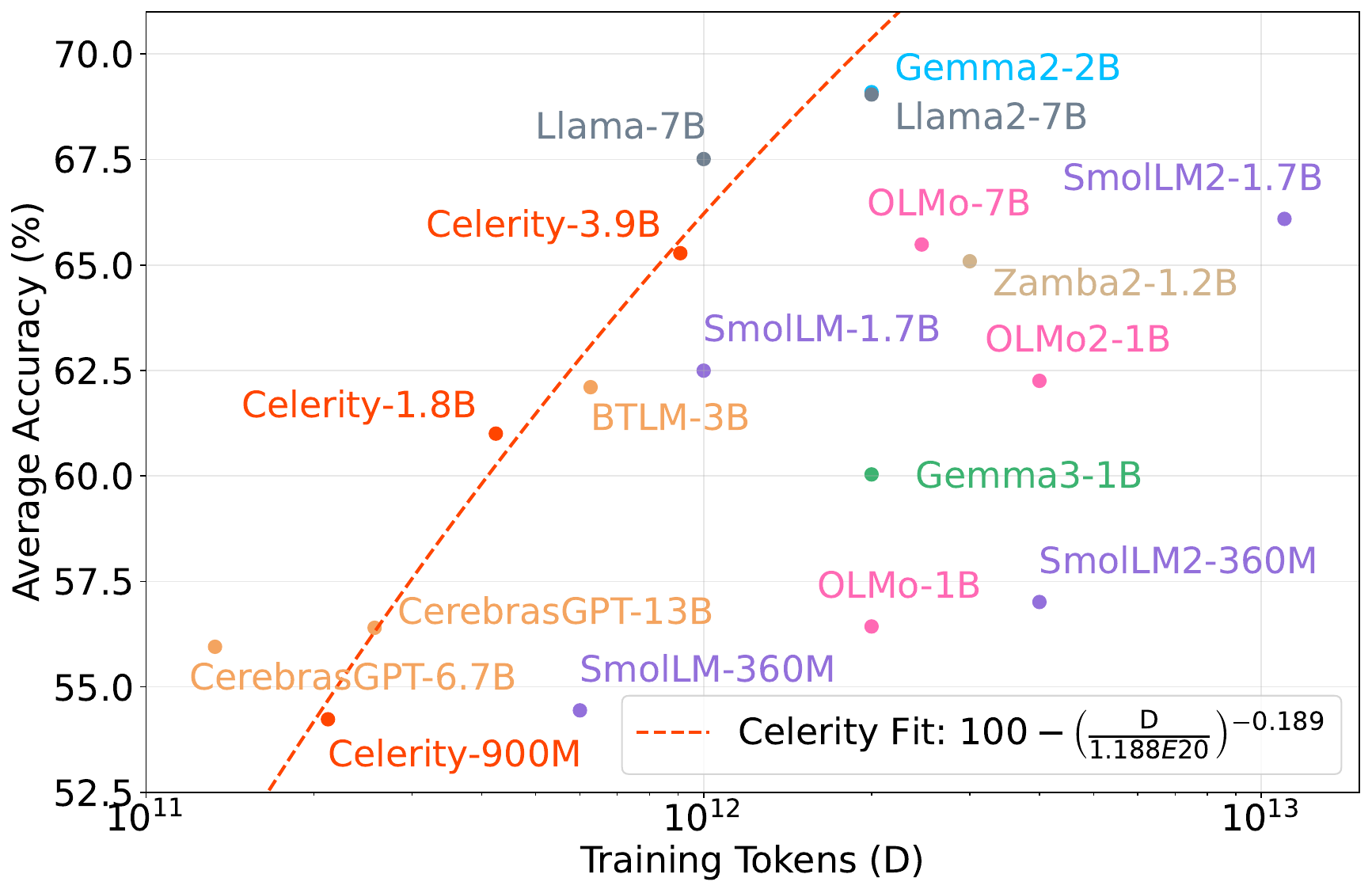}
    \caption{\textbf{Celerity token efficiency:} Downstream
      accuracy. Celerity models are at the Pareto frontier compared to
      other model families.\label{fig:celerity_token1}}
\end{figure}

\begin{figure}[ht]
  \centering
    \includegraphics[width=0.58\linewidth]{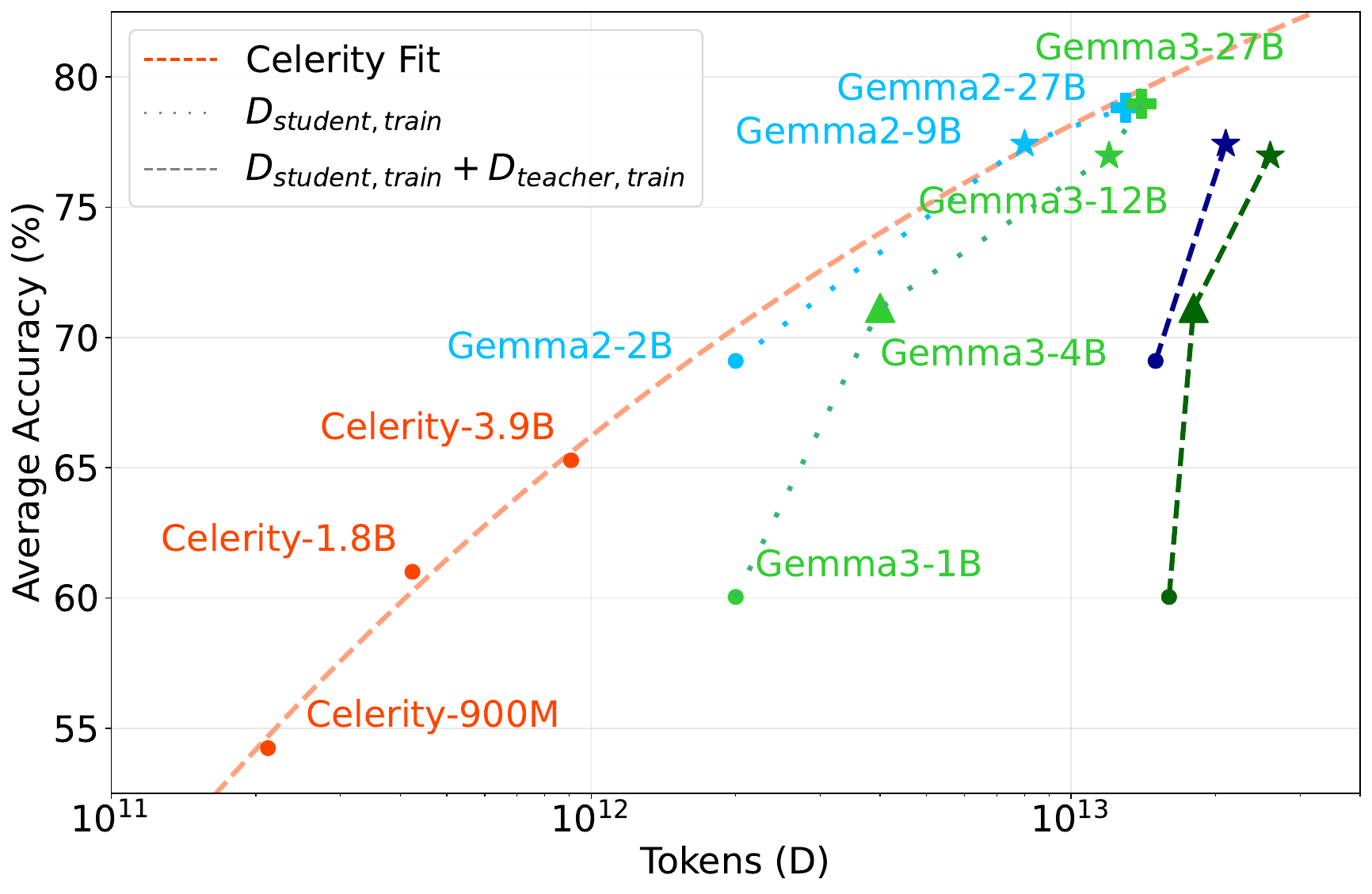}
    \caption{\textbf{Celerity token efficiency vs.\ distilled models:}
      Downstream accuracy. Celerity models are on par with distilled
      models.\label{fig:celerity_token2}}
\end{figure}

\Cref{fig:celerity_token1} compares the \emph{token} efficiency of
Celerity to other model families.
Using the Celerity models trained in the fixed 234~TPP band, we fit a
power law in $D$ and extrapolate token efficiency to larger scales.

Generally, larger models should be more token-efficient for the same
token budget.  Theoretically, distillation should also offer greater
token efficiency---at a given
TPP~\citep{busbridge2025distillation}---but by training small models
to very-high TPP, the distilled models in \cref{fig:celerity_token2}
train mainly in a regime of diminishing returns, and so ultimately end
up without an advantage over Celerity's standard next-token-prediction
training.

There are many interesting questions around token efficiency at scale,
and indeed token efficiency may become more critical as frontier
models reach the limits of high-quality
data~\citep{muennighoff2023scaling}.

\paragraph{Parameter efficiency.}

\begin{figure}[ht]
  \centering
    \includegraphics[width=0.58\linewidth]{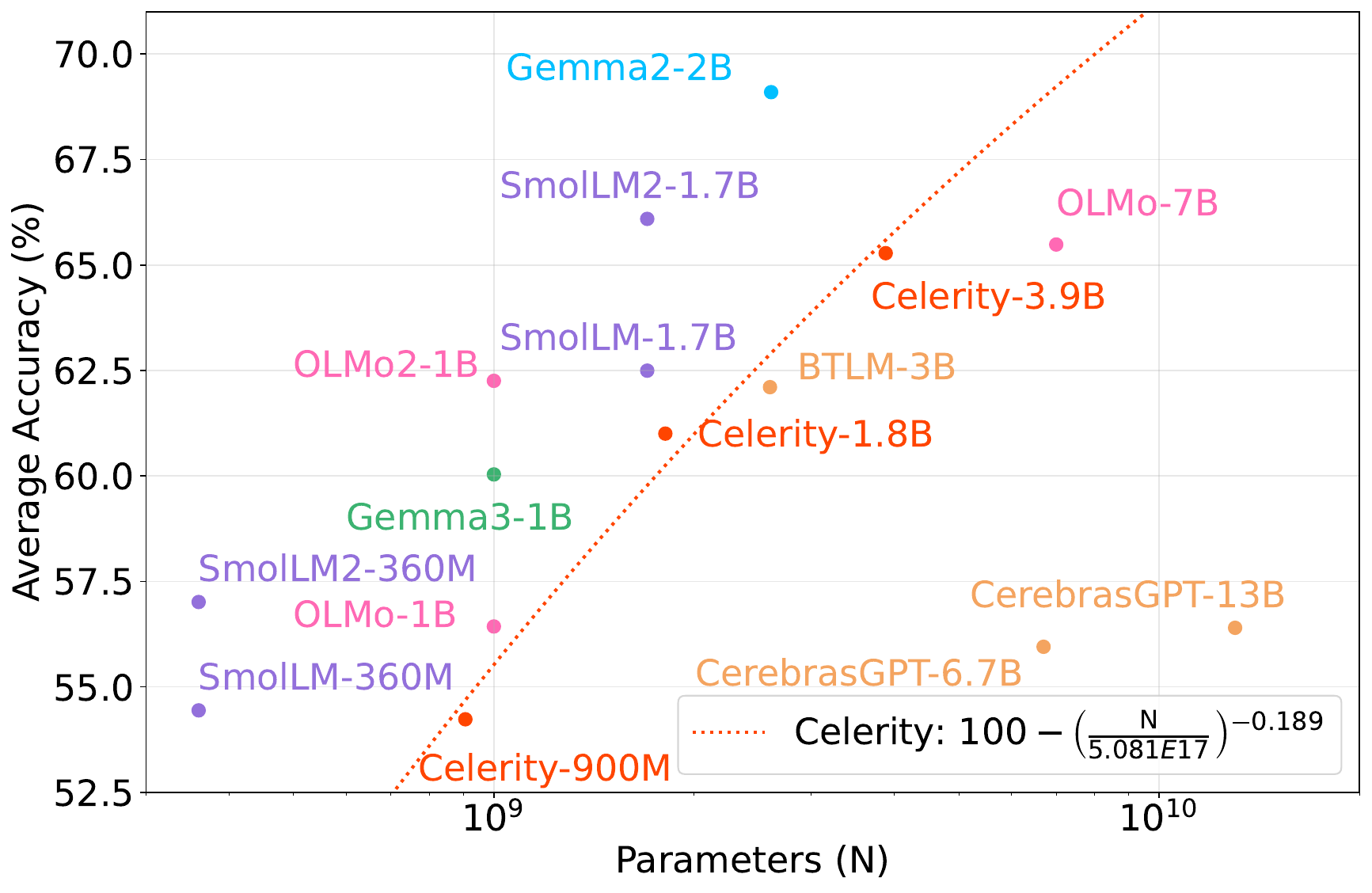}
    \caption{\textbf{Celerity parameter efficiency:} Downstream
      accuracy. Celerity models are less parameter efficient than
      models trained at much higher TPP, while better than prior
      models also aiming for compute efficiency (Cerebras
      GPT).\label{fig:celerity_parameter1}}
\end{figure}

\begin{figure}[ht]
  \centering
    \includegraphics[width=0.58\linewidth]{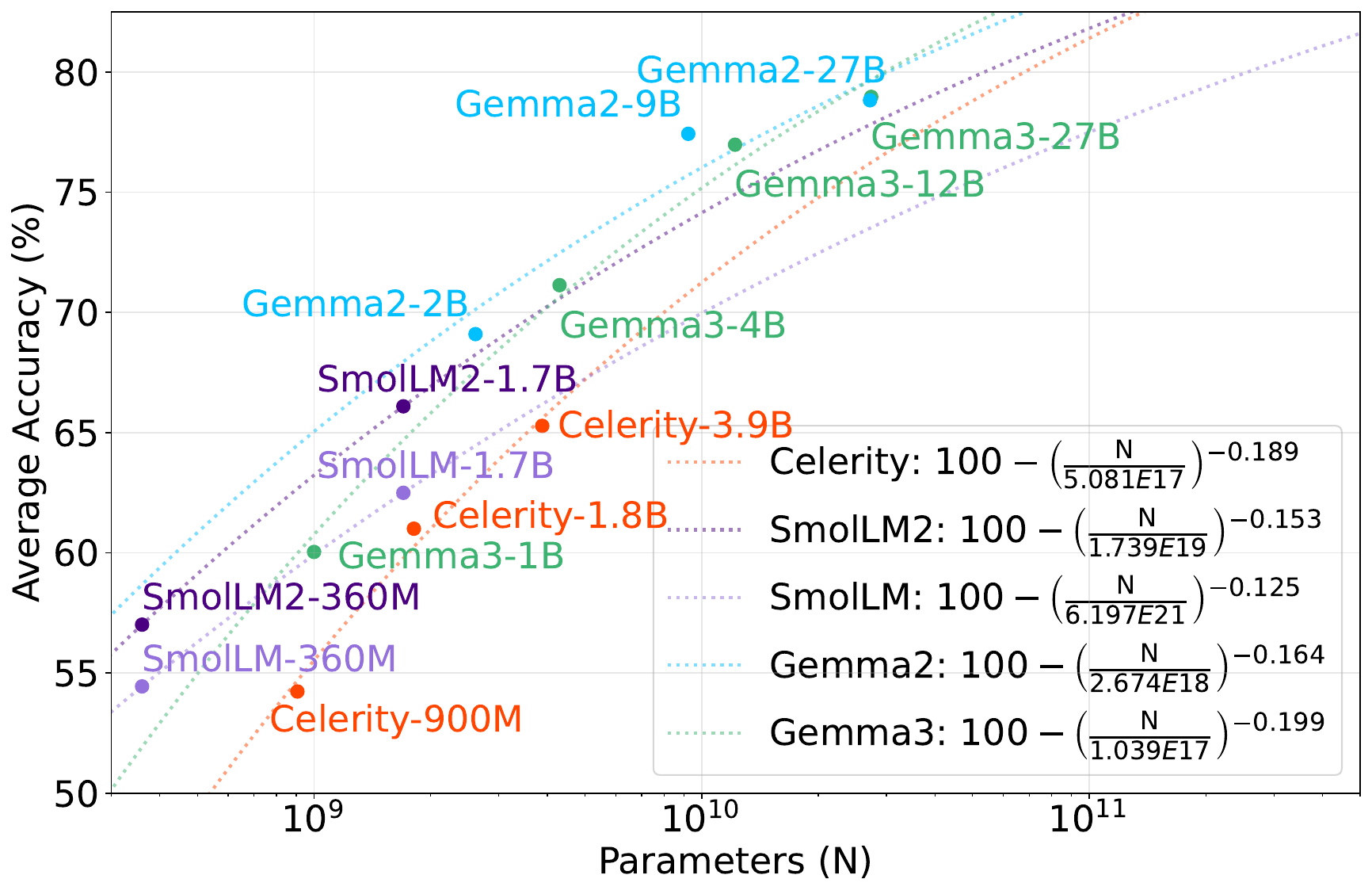}
    \caption{\textbf{Celerity parameter efficiency scaling
        comparison:} Downstream accuracy.  Preliminary accuracy
      vs.\ model size power law comparison between Gemma
      (distillation), SmolLM (refined data), and Celerity models
      (standard pre-training).  Distilled model families have the
      largest scaling exponent, suggesting distillation may scale
      better in parameters.\label{fig:celerity_parameter2}}
\end{figure}

Finally, \cref{fig:celerity_parameter1,fig:celerity_parameter2}
provide the parameter efficiency comparisons for Celerity.  Celerity
models are less parameter efficient than models specifically designed
for parameter efficiency.

\subsection{Open model evaluation and FLOP calculation methods}

All models are obtained from HuggingFace and evaluated using the
Eleuther Eval Harness framework \citep{eval-harness}. The downstream
tasks with number of shots are arc-challenge (25), arc-easy (0), boolq
(0), hellaswag (10), piqa (0), siqa (0) and winogrande (5). These
tasks are chosen as they are the most commonly reported downstream
benchmarks for pre-trained base models, and are appropriate for
Celerity models of the scale that we compare (i.e., tasks where small
models perform above random chance).

\begin{table}[ht]
  \centering
  \caption{Forward FLOPs calculation for self-attention block and
    Mamba-2 block. This table only lists operations that are not covered
    $6*n_{params}*n_{tokens}$, which should take care of all operations
    that involves a matmul with a weight matrix. For Zamba2, the
    training FLOPs can be calculated as $6*n_{params}*n_{tokens} -
    2*V*D_{attn}*n_{tokens} + 3 * L_{attn}*(L*C_{mamba2}+C_{attn})$,
    while the rest of the models analyzed are variations of
    decoder-transformers whose training FLOPs can be estimated as
    $6*n_{params}*n_{tokens} - 2*V*D_{attn}*n_{tokens} + 3 *
    L_{attn}*C_{attn}$. Here 3 represents 1 flop per forward op and 2
    flops per backward op.\label{tab:flops_calc}}
\resizebox{\textwidth}{!}
{
 \begin{tabular}{cll}
    \toprule
    & Operation & FLOPs, given input: \\
    &           & $B \times S \times D$ (or $D_{attn}$) \\
    \midrule
    \multirow{3}{*}{Self-Attention} & Attention: $QK^T$ & $2BS^2D_{attn}$ \\
    & Attention: softmax, scaling, mask & $3BS^2$ \\
    \multirow{2}{*}{$C_{attn}$} & Attention: $V$ matmul & $2BS^2D_{attn}$ \\
    & Attention: $O$ projection & $2BSD_{attn}^2$ \\
    & Feedforward: activation & $BSD_{attn}$ \\
    \midrule
    \multirow{5}{*}{Mamba-2} & dt softplus & $3BSH$ \\
    & xBC conv1d, silu & $BS(ED+2N)K+5BS(ED+2N)$ \\
    \multirow{5}{*}{$C_{mamba2}$} & sampling x, A & $BSED+BSH$ \\
    & SSD, A prefix sum & $BHS$ \\
    & SSD, compute output for each intra-chunk & $4BHSC+BSEDNC$ \\
    & SSD, compute state for each intra-chunk & $2BHS+BSEDN$ \\
    & SSD, compute inter-chunk recurrence & $4BH(Z+1)^2+BN(Z + 1)^2ED$ \\
    & SSD, compute output from state per chunk & $BHS+BSEDN+BSED$ \\
    & y+x*D & $2BSED$ \\
    & z silu, y norm & $6BSED$ \\
    \midrule
    Params & \multicolumn{2}{l}{$B$: batch size, $S$: sequence length, $V$: vocabulary size} \\
    & \multicolumn{2}{l}{$D_{attn}$: attention hidden dim, $L_{attn}$: num attention layers} \\
    & \multicolumn{2}{l}{$D$: mamba2 hidden dim, $L$: num mamba2 layers, $E$: expansion factor} \\
    & \multicolumn{2}{l}{$N$: mamba2 state dim, $H$: mamba2 num heads, $P$: mamba2 head dim} \\
     & \multicolumn{2}{l}{$C$: mamba2 chunk size, $Z$: mamba2 num chunks, $K$: mamba2 conv dim} \\
    \bottomrule
 \end{tabular}
}
\end{table}

\begin{table}[ht]
\centering
\caption{Evaluations, params, tokens, and FLOPs for all models
  evaluated.\label{tab:all_evals}}
\resizebox{\textwidth}{!}  {
\begin{tabular}{l|ccc|cccccccc}
    \toprule
    \multirow{2}{*}{Name} & \multirow{2}{*}{Params} & \multirow{2}{*}{Tokens} & \multirow{2}{*}{FLOPs} & \multicolumn{8}{c}{Downstream Accuracy (Num Shots)}\\
    &&&& arc-c & arc-e & boolq & hellaswag & piqa & siqa & winogrande & Avg.\\
    &&&& (25) & (0) & (0) & (10) & (0) & (0) & (5) & \\
    \midrule
    BTLM-3b-8k-base \citep{dey2023btlm3b8k} & 2.60E+09 & 6.27E+11 & 1.55E+22 & 40.70 & 66.79 & 69.72 & 70.92 & 77.20 & 43.50 & 65.90 & 62.10 \\ 
    \midrule
    Cerebras-GPT-1.3B \citep{dey2023cerebras} & 1.30E+09 & 2.60E+10 & 2.45E+20 & 26.79 & 45.83 & 59.33 & 38.55 & 66.76 & 38.59 & 51.70 & 46.79 \\
    Cerebras-GPT-2.7B & 2.70E+09 & 5.40E+10 & 1.04E+21 & 29.52 & 52.57 & 59.24 & 49.74 & 70.78 & 40.23 & 54.85 & 50.99 \\
    Cerebras-GPT-6.7B & 6.70E+09 & 1.33E+11 & 6.16E+21 & 36.01 & 57.91 & 62.81 & 59.45 & 73.99 & 41.50 & 59.98 & 55.95 \\
    \midrule
    Gemma-2-2b \citep{team2024gemma2} & 2.61E+09 & 2.00E+12 & 4.25E+22 & 53.41 & 80.22 & 73.58 & 74.62 & 79.11 & 51.23 & 71.51 & 69.10 \\
    Gemma-2-9b & 9.24E+09 & 8.00E+12 & 5.73E+23 & 68.34 & 87.88 & 84.22 & 82.76 & 82.97 & 55.48 & 80.35 & 77.43 \\
    Gemma-2-27b & 2.72E+10 & 1.30E+13 & 2.44E+24 & 69.62 & 88.30 & 84.83 & 87.00 & 84.44 & 54.55 & 83.03 & 78.82 \\
    Gemma-2-2b+forward & 2.61E+09 & 2.00E+12 & 8.66E+23 & 53.41 & 80.22 & 73.58 & 74.62 & 79.11 & 51.23 & 71.51 & 69.10 \\
    Gemma-2-2b+forward+teacher & 2.61E+09 & 1.50E+13 & 3.31E+24 & 53.41 & 80.22 & 73.58 & 74.62 & 79.11 & 51.23 & 71.51 & 69.10 \\
    Gemma-2-9b+forward & 9.24E+09 & 8.00E+12 & 1.40E+24 & 68.34 & 87.88 & 84.22 & 82.76 & 82.97 & 55.48 & 80.35 & 77.43 \\
    Gemma-2-9b+forward+teacher & 9.24E+09 & 2.10E+13 & 3.84E+24 & 68.34 & 87.88 & 84.22 & 82.76 & 82.97 & 55.48 & 80.35 & 77.43 \\    
    \midrule
    Gemma-3-1b-pt \citep{team2025gemma} & 1.00E+09 & 2.00E+12 & 3.48E+22 & 39.16 & 71.93 & 66.67 & 62.98 & 74.54 & 42.78 & 62.19 & 60.04 \\
    Gemma-3-4b-pt & 4.30E+09 & 4.00E+12 & 1.72E+23 & 58.28 & 81.69 & 78.96 & 77.78 & 79.87 & 49.13 & 72.22 & 71.13 \\
    Gemma-3-12b-pt & 1.22E+10 & 1.20E+13 & 1.34E+24 & 67.49 & 87.75 & 85.41 & 84.12 & 81.88 & 52.15 & 80.03 & 76.98 \\
    Gemma-3-27b-pt & 2.74E+10 & 1.40E+13 & 3.33E+24 & 70.31 & 88.17 & 87.25 & 86.14 & 83.95 & 53.99 & 82.95 & 78.97 \\
    Gemma-3-1b-pt+forward & 1.00E+09 & 2.00E+12 & 1.16E+24 & 39.16 & 71.93 & 66.67 & 62.98 & 74.54 & 42.78 & 62.19 & 60.04 \\
    Gemma-3-1b-pt+forward+teacher & 1.00E+09 & 1.60E+13 & 4.49E+24 & 39.16 & 71.93 & 66.67 & 62.98 & 74.54 & 42.78 & 62.19 & 60.04 \\
    Gemma-3-4b-pt+forward & 4.00E+09 & 4.00E+12 & 1.30E+24 & 58.28 & 81.69 & 78.96 & 77.78 & 79.87 & 49.13 & 72.22 & 71.13 \\
    Gemma-3-4b-pt+forward+teacher & 4.00E+09 & 1.80E+13 & 4.63E+24 & 58.28 & 81.69 & 78.96 & 77.78 & 79.87 & 49.13 & 72.22 & 71.13 \\
    Gemma-3-12b-pt+forward & 1.20E+10 & 1.20E+13 & 2.46E+24 & 67.49 & 87.75 & 85.41 & 84.12 & 81.88 & 52.15 & 80.03 & 76.98 \\
    Gemma-3-12b-pt+forward+teacher & 1.20E+10 & 2.60E+13 & 5.80E+24 & 67.49 & 87.75 & 85.41 & 84.12 & 81.88 & 52.15 & 80.03 & 76.98 \\
    \midrule
    Llama-7b \citep{touvron2023llama} & 7.00E+09 & 1.00E+12 & 4.82E+22 & 50.77 & 72.90 & 75.05 & 77.84 & 79.00 & 45.91 & 71.11 & 67.51 \\
    Llama-13b & 1.30E+10 & 1.00E+12 & 8.90E+22 & 55.55 & 74.54 & 77.98 & 81.18 & 80.36 & 46.62 & 76.95 & 70.45 \\
    \midrule
    Llama-2-7b-hf \citep{touvron2023llama2} & 7.00E+09 & 2.00E+12 & 1.03E+23 & 52.65 & 74.54 & 77.71 & 78.98 & 79.11 & 46.11 & 74.19 & 69.04 \\ 
    Llama-2-13b-hf & 1.30E+10 & 2.00E+12 & 1.88E+23 & 59.47 & 77.53 & 80.58 & 82.23 & 80.52 & 47.34 & 76.16 & 71.98 \\
    \midrule
    Llama-3-8B \citep{dubey2024llama} & 8.00E+09 & 1.50E+13 & 9.46E+23 & 58.19 & 77.61 & 80.95 & 82.10 & 80.69 & 47.08 & 77.51 & 72.02 \\
    Llama-3.1-8B & 8.00E+09 & 1.50E+13 & 3.85E+24 & 57.85 & 81.19 & 82.05 & 81.91 & 81.01 & 46.98 & 77.19 & 72.60 \\
    Llama-3.2-1B & 1.23E+09 & 9.00E+12 & 5.29E+23 & 39.59 & 60.61 & 63.91 & 65.51 & 74.27 & 42.99 & 62.27 & 58.45 \\
    Llama-3.2-3B & 3.21E+09 & 9.00E+12 & 1.40E+24 & 50.68 & 71.84 & 72.75 & 76.42 & 77.37 & 47.39 & 71.82 & 66.90 \\
    \midrule
    OLMo-1B-hf \citep{muennighoff2024olmoe} & 1.00E+09 & 2.00E+12 & 1.56E+22 & 34.47 & 57.28 & 61.74 & 63.81 & 75.14 & 42.12 & 60.46 & 56.43 \\
    OLMo-7B-hf & 7.00E+09 & 2.46E+12 & 1.26E+23 & 45.14 & 68.77 & 72.45 & 77.13 & 79.43 & 44.52 & 70.96 & 65.49 \\
    \midrule
    OLMo-2-0425-1B \citep{olmo2024} & 1.00E+09 & 4.00E+12 & 3.04E+22 & 45.39 & 73.36 & 63.03 & 68.71 & 75.63 & 43.76 & 65.90 & 62.25 \\
    OLMo-2-1124-7B & 7.00E+09 & 4.00E+12 & 2.03E+23 & 64.51 & 82.87 & 80.00 & 81.93 & 81.01 & 51.33 & 77.03 & 74.10 \\
    OLMo-2-1124-13B & 1.30E+10 & 5.00E+12 & 4.67E+23 & 66.13 & 81.31 & 73.91 & 84.99 & 82.15 & 52.05 & 83.03 & 74.80 \\
    OLMo-2-0325-32B & 3.20E+10 & 6.00E+12 & 1.30E+24 & 69.45 & 85.94 & 82.81 & 87.33 & 82.97 & 54.25 & 83.90 & 78.09 \\
    \midrule 
    Qwen3-0.6B-Base \citep{yang2025qwen3} & 6.00E+08 & 3.60E+13 & 5.31E+23 & 44.80 & 58.00 & 69.82 & 53.46 & 69.80 & 43.30 & 60.46 & 57.09 \\
    Qwen3-1.7B-Base & 1.70E+09 & 3.60E+13 & 1.18E+24 & 55.20 & 68.60 & 79.24 & 67.19 & 75.52 & 48.62 & 65.27 & 65.66 \\
    Qwen3-4B-Base & 4.00E+09 & 3.60E+13 & 2.19E+24 & 64.42 & 75.93 & 82.91 & 75.64 & 77.86 & 50.00 & 72.61 & 71.34 \\
    Qwen3-8B-Base & 8.00E+09 & 3.60E+13 & 3.90E+24 & 67.24 & 79.88 & 83.09 & 79.55 & 79.54 & 54.76 & 77.19 & 74.46 \\
    Qwen3-14B-Base & 1.40E+10 & 3.60E+13 & 6.09E+24 & 69.97 & 81.86 & 86.76 & 82.69 & 82.10 & 55.89 & 79.48 & 76.96 \\
    \midrule
    Qwen2.5-0.5B \citep{yang2024qwen2_5} & 5.00E+08 & 1.80E+13 & 2.04E+23 & 35.24 & 58.54 & 61.47 & 51.83 & 69.80 & 44.17 & 56.59 & 53.95 \\
    Qwen2.5-1.5B & 1.50E+09 & 1.80E+13 & 1.38E+24 & 54.86 & 72.10 & 72.48 & 67.86 & 75.90 & 49.08 & 65.27 & 65.36 \\
    Qwen2.5-3B & 3.00E+09 & 1.80E+13 & 8.51E+23 & 56.31 & 73.02 & 77.43 & 74.54 & 78.67 & 49.80 & 71.67 & 68.78 \\
    Qwen2.5-7B & 7.00E+09 & 1.80E+13 & 3.62E+24 & 63.65 & 77.48 & 84.65 & 80.19 & 79.82 & 54.61 & 76.40 & 73.83 \\
    Qwen2.5-14B & 1.40E+10 & 1.80E+13 & 8.58E+24 & 67.58 & 79.25 & 85.35 & 84.21 & 82.43 & 55.48 & 81.06 & 76.48 \\
    Qwen2.5-32B & 3.20E+10 & 1.80E+13 & 1.29E+25 & 70.65 & 77.99 & 87.49 & 85.16 & 82.43 & 56.29 & 82.08 & 77.44 \\
    \midrule
    SmolLM-135M \citep{allal2024smollm} & 1.35E+08 & 6.00E+11 & 1.51E+21 & 32.00 & 56.14 & 60.09 & 42.92 & 68.01 & 39.56 & 52.25 & 50.14 \\
    SmolLM-360M & 3.60E+08 & 6.00E+11 & 3.16E+21 & 38.65 & 63.59 & 55.05 & 54.24 & 71.44 & 40.99 & 57.14 & 54.44 \\
    SmolLM-1.7B & 1.70E+09 & 1.00E+12 & 1.54E+22 & 49.40 & 73.57 & 66.15 & 67.33 & 75.95 & 43.35 & 61.72 & 62.50 \\
    \midrule
    SmolLM2-135M \citep{allal2025smollm2} & 1.35E+08 & 2.00E+12 & 2.48E+21 & 33.02 & 58.38 & 60.06 & 43.64 & 68.12 & 39.25 & 53.12 & 50.80 \\
    SmolLM2-360M & 3.60E+08 & 4.00E+12 & 1.20E+22 & 40.78 & 68.22 & 61.56 & 57.46 & 71.76 & 40.89 & 58.41 & 57.01\\
    SmolLM2-1.7B & 1.70E+09 & 1.10E+13 & 1.21E+23 & 53.50 & 73.27 & 72.32 & 73.16 & 77.53 & 44.52 & 68.35 & 66.09 \\
    \midrule
    SmolLM3-3B-Base & 3.00E+09& 1.12E+13& 8.56E+23& 59.81& 76.85& 80.49& 77.18& 79.11& 46.78& 73.40& 70.52 \\
    \midrule
    Zamba2-1.2B \citep{glorioso2024zamba2} & 1.20E+09 & 3.00E+12 & 3.86E+23 & 53.92 & 66.71 & 70.18 & 72.21 & 77.20 & 46.42 & 68.98 & 65.09 \\
    Zamba2-2.7B & 2.70E+09 & 3.00E+12 & 4.77E+23 & 60.67 & 73.82 & 78.07 & 77.72 & 79.49 & 45.50 & 76.01 & 70.18 \\
    Zamba2-7B & 7.40E+09 & 2.00E+12 & 7.68E+23 & 68.34 & 80.39 & 83.70 & 83.53 & 80.69 & 49.90 & 79.72 & 75.18 \\
    \midrule
    Celerity-300M & 2.71E+08 & 6.34E+10 & 1.47E+20 & 27.82 & 50.63 & 52.75 & 37.57 & 66.21 & 37.77 & 52.25 & 46.43 \\
    Celerity-500M & 5.03E+08 & 1.18E+11 & 5.15E+20 & 34.39 & 56.06 & 61.22 & 45.96 & 69.31 & 40.23 & 52.64 & 51.40 \\
    Celerity-900M & 9.06E+08 & 2.12E+11 & 1.68E+21 & 39.68 & 64.52 & 47.92 & 55.02 & 72.03 & 41.97 & 58.48 & 54.23 \\
    Celerity-1.8B & 1.81E+09 & 4.24E+11 & 6.54E+21 & 48.55 & 70.29 & 65.17 & 64.34 & 75.46 & 42.99 & 60.22 & 61.00 \\
    Celerity-3.9B & 3.88E+09 & 9.08E+11 & 2.89E+22 & 54.01 & 75.55 & 66.61 & 72.19 & 77.97 & 44.73 & 65.90 & 65.28 \\
    \bottomrule
\end{tabular}
}
\end{table}

For full transparency, our method for counting FLOPs across the
different models families is given in \cref{tab:flops_calc}, while a
table of all the raw downstream evaluation scores are in
\cref{tab:all_evals}.

\subsection{Validation loss collapse}\label{sec:val_loss}

\begin{figure}[ht]
  \centering
    \includegraphics[width=0.33\linewidth]{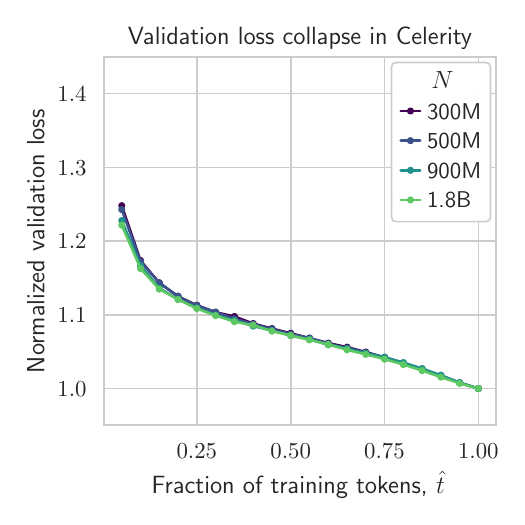}
    \caption{\textbf{Training loss curves also collapse in
        \emph{validation} loss:} Normalized validation loss across 4
      Celerity model sizes, all trained to 80 TPP.  Validation loss
      collected at 5\% intervals of training.\label{fig:val_loss}}
\end{figure}

In \cref{fig:val_loss} we show normalized training loss curves where
we evaluate the \emph{validation loss} of model checkpoints during
training.  For each training run, we evaluate checkpoints at 5\%
intervals, computing loss on the same 493M-token held-out portion of
SlimPajama.  Similar to training, collapse occurs except for the
initial few checkpoints; we attribute the initial differences to
differing LR warmup proportions (\cref{tab:celerity_model_arch}).
Nevertheless, validation collapse is sufficient for deviations to
provide a useful diagnostic of any training issues.  Validation
collapse, measured on fixed datasets, could be a particularly valuable
diagnostic if late-stage annealing or curriculum strategies distort
training curves due to data differences.

\section{Collapse enables early stopping: further details}

\subsection{Predicting normalized training loss curves}\label{sec:predicting}

This section provides further details regarding the development of the
functional form in \cref{eq:prediction}, which we use to predict
normalized TLCs and, through these, extrapolate in-progress TLCs.
Based on \cref{sec:shape}, we know that TLC shape is modulated by LR
schedule, TPP, and $\tema$.  Prior theoretical and empirical work has
mostly focused on how loss proceeds as a function of training steps
and LR
schedule~\citep{defazio2023optimal,tissue2024scaling,schaipp2025surprising,luo2025multi,qiu2025scaling}.
To incorporate these factors into a single functional form, we take
the following approach:
\begin{itemize}
\item Use a functional form that accounts for training fraction and LR schedule
\item Make the \emph{parameters} of this functional form depend on TPP and $\tema$
\end{itemize}
This led to \cref{eq:prediction}.  Our initial aim here is not to
develop the best possible TLC predictor, but to obtain a simple,
effective, and interpretable method for extrapolating TLCs, allowing
us to test the value of this extrapolation for early stopping in
hyperparameter tuning.

We conducted a variety of preliminary experiments at 111M-scale, using
the same data as in \cref{sec:shape} (with details in
\cref{sec:experimental_details}).
As an input to \cref{eq:prediction}, the LR schedule is normalized to
be at 1.0 at its peak.  It's also interpreted over training fraction,
so from 0.0 to 1.0.
Experiments in this data only use fits for linear decay-to-zero
schedules.
To get an initial sense of how the parameters in \cref{eq:prediction}
vary, we did a multi-dimensional grid search to determine optimal
parameters for each individual curve, measuring total macroaveraged
MAE loss over all 111M-scale TLCs.  Over the course of these
experiments, we found total MAE did not change substantially when we
fixed $m = 0.05$, and we subsequently tuned $\epsilon_1$ and
$\epsilon_2$ to small constants in order to avoid boundary effects at
$\trainfrac = 0$ and $\trainfrac = 1$ (when $\eta(\trainfrac)$ goes to
zero).
Prior to fitting, training curves were smoothed using a moving average
filter covering 12288 sequences (equal to the largest batch size in
the dataset), and we ignore error on the first 20\% of each curve
(around LR warmup when curves are noisy).

\begin{figure}[ht]
  \centering
  \begin{minipage}{0.48\textwidth}
    \includegraphics[trim={0.3cm 0.42cm 0.264cm 0.3cm}, clip, width=\linewidth]{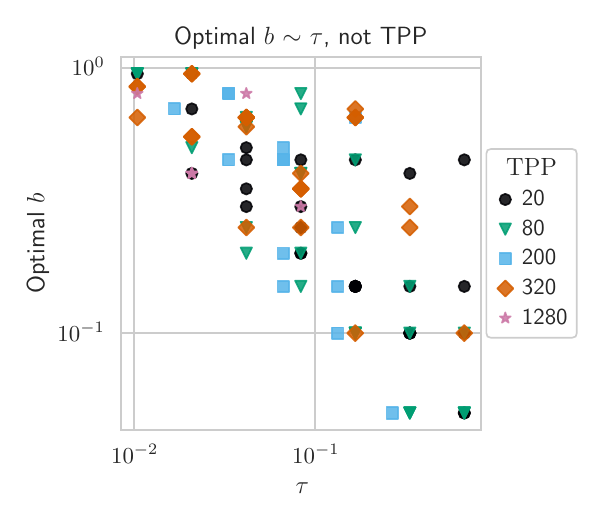}
  \end{minipage}\hfill
  \begin{minipage}{0.48\textwidth}
    \includegraphics[trim={0.3cm 0.42cm 0.264cm 0.3cm}, clip, width=\linewidth]{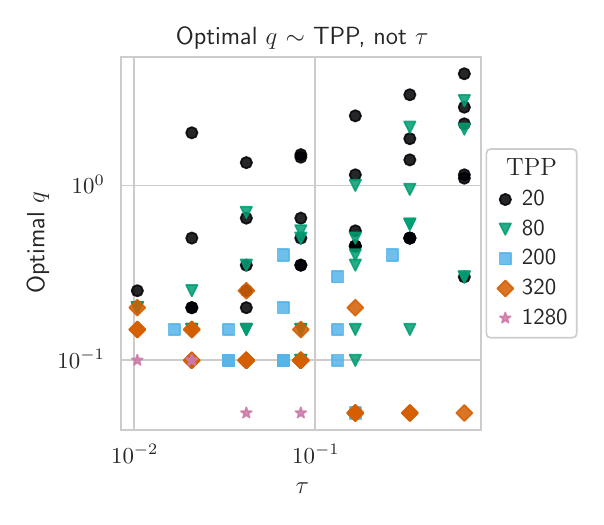}
  \end{minipage}
  \caption{\textbf{Trends in fits for training curve prediction.}
    Optimal per-curve fits (from per-curve grid
    searches) for \cref{eq:prediction}:
    $\ellhat(\trainfrac) \approx 1/m^{0.05} + b \cdot
    \eta(\trainfrac)^q$: $b$ and $q$ parameters.
    $\Leftfig$: Optimal $b$ varies strongly in $\tema$ (Pearson's $r$
    = -0.59), weakly in TPP ($r$ = 0.17).
    $\Rightfig$: Optimal $q$ varies somewhat in TPP ($r$ = -0.30),
    while overall stronger in $\tema$ ($r$ = 0.55), but $\tau$ trends
    reverse at higher TPP.\label{fig:fits}}
\end{figure}

\Cref{fig:fits} shows the optimal fits for $b$ and $q$ when each curve
is fit independently.  For \textbf{optimal $b$}, we found that
correlation in $\tema$ was much stronger than correlation in TPP
(Pearson's $r$ = -0.59 for $\tema$, $r$ = 0.17 for TPP).  On the other
hand, while \textbf{optimal $q$} seems to increase with $\tema$ for
TPP = 20, the relationship with $\tema$ at other TPP appears random.
Furthermore, note larger TPP values do correspond to lower optimal $q$
($r$ = -0.30).  Based on these fits, we hypothesize we could obtain
reasonable predictions by fitting $b$ as a power law in $\tema$, and
$q$ as a power law in TPP:
\begin{equation}
b = \bconst \cdot {\tema}^{\bexp},\quad q = \qconst \cdot \tpp^{\qexp}
\end{equation}
As noted in \cref{sec:tuning}.  Also, as reported in that section, we
developed an alternating greedy optimization procedure to fit these
four parameters, exponentially reducing the cost of the grid search
space.

\paragraph{Results.}

We first note that the fits improve over the iterations of our
alternating grid search procedure, demonstrating that optimal
parameters of the power laws do depend on each other, and can reach
stable fits through iterative alternating fitting.

\begin{table}[ht]
    \centering
    \caption{
      \textbf{Predictions improve with scale}:
      fit at 111M scale, evaluated at larger scales.}
    \label{tab:fit_scaling}
    \begin{tabular}{@{}ccc@{}}
      \toprule
      Evaluation scale       & MAE                                               & Number of evaluation curves \\ \midrule
      111M* (fitting points) & {\color[HTML]{1F2328} 1.37\%}                     & 112                         \\
      266M                   & \multicolumn{1}{l}{{\color[HTML]{1F2328} 0.75\%}} & 40                          \\
      610M                   & 1.07\%                                            & 102                         \\
      1.7B                   & 0.66\%                                            & 21                          \\
      3.3B                   & 0.54\%                                            & 7                           \\ \bottomrule
    \end{tabular}
\end{table}

\begin{table}[ht]
    \centering
    \caption{\textbf{Separate power laws for $b$ and $q$ work well}:
      fit at 111M scale (112 TLCs), evaluation at 610M (102 TLCs).}
    \label{tab:fit_ablation}
    \begin{tabular}{@{}ccc@{}}
      \toprule
      Method for estimating $b$ & Method for estimating $q$ & MAE    \\ \midrule
      Global fixed optimum      & Global fixed optimum      & 3.03\% \\
      Global fixed optimum      & $q=\powerlaw(\tpp)$       & 3.35\% \\
      $b=\powerlaw(\tema)$      & Global fixed optimum      & 2.08\% \\
      $b=\powerlaw(\tema)$      & $q=\powerlaw(\tpp)$       & 1.07\% \\
      $b=\powerlaw(\tema,\tpp)$ & $q=\powerlaw(\tema,\tpp)$ & 1.07\% \\ \bottomrule
    \end{tabular}
\end{table}

\cref{tab:fit_scaling} and \cref{tab:fit_ablation} are the tables
discussed in the main paper, showing how fits obtained at 111M perform
at other scales (\cref{tab:fit_scaling}), and how different fitting
procedures perform on the 610M-scale evaluation data
(\cref{tab:fit_ablation}).  Fitting $b$ and $q$ with the optimum
values \emph{per-curve} (i.e., \emph{oracle} fits) achieves an MAE of
0.504\%, roughly half that of the dual power law extrapolations.

\subsection{Early stopping in tuning: further results}\label{sec:app_early}

In this section we describe some further early stopping experiments,
and present additional evaluation metrics.

\begin{figure}[ht]
  \centering
  \begin{minipage}{0.66\textwidth}
    \begin{minipage}{0.49\textwidth}
      \includegraphics[trim={0.3cm 0.42cm 0.264cm 0.3cm}, clip, width=\linewidth]{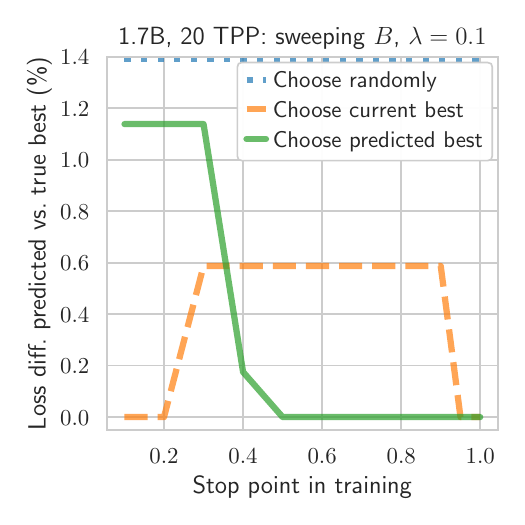}
    \end{minipage}\hfill
    \begin{minipage}{0.49\textwidth}
      \includegraphics[trim={0.3cm 0.42cm 0.264cm 0.3cm}, clip, width=\linewidth]{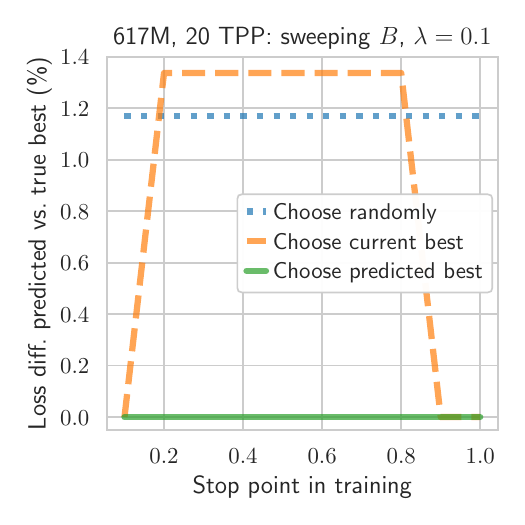}
    \end{minipage}
  \end{minipage}
    \caption{\textbf{Early stopping comparison: further setups.}
      Companion to \cref{fig:tuning_tests}, now comparing early
      stopping accuracy (final loss of predicted vs.\ actual best) for
      $B$ sweeps at 1.7B ($\leftfig$) and 617M ($\rightfig$) (both 20
      TPP).  \emph{Current best} works well very early, but is worse
      for most of training.\label{fig:hptuning3}}
\end{figure}

\cref{fig:hptuning3} evaluates early stopping strategies in batch-size
sweeps at a fixed $\lambda$ value.  \cref{fig:hptuning3}, $\leftfig$,
uses the same data as in \cref{fig:hptuning}, $\leftfig$.  While we do
not advocate keeping $\lambda$ fixed during $B$ sweeps in practice,
this data can nevertheless serve to evaluate prediction of early
winners in tuning.
Both of these plots exhibit the phenomenon also observed in
\cref{fig:tuning_tests}, $\rightfig$: choosing the current best
setting after LR warmup, as was done in
Falcon~\citep{almazrouei2023falcon}, is better than selecting the best
during the middle of training.  However, as seen in
\cref{fig:tuning_tests}, $\leftfig$, this method is not always
successful.  In \cref{fig:hptuning3}, $\leftfig$, choosing the
extrapolated best setting outperforms choosing the current best from
40\% of training, while it picks the correct winner from the beginning
in \cref{fig:hptuning3}, $\rightfig$.

\begin{figure}[ht]
  \centering
  \begin{minipage}{0.66\textwidth}
    \begin{minipage}{0.49\textwidth}
      \includegraphics[trim={0.3cm 0.42cm 0.264cm 0.3cm}, clip, width=\linewidth]{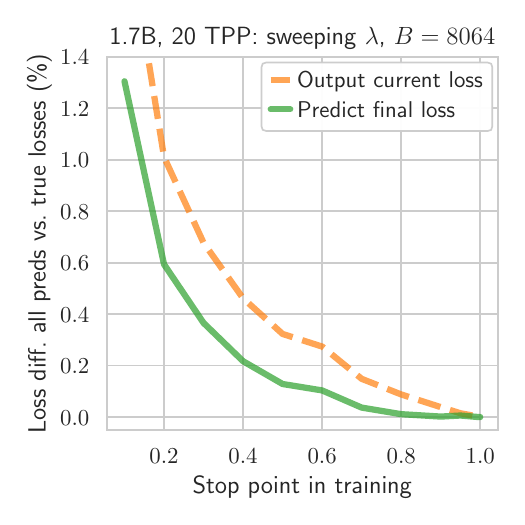}
    \end{minipage}\hfill
    \begin{minipage}{0.49\textwidth}
      \includegraphics[trim={0.3cm 0.42cm 0.264cm 0.3cm}, clip, width=\linewidth]{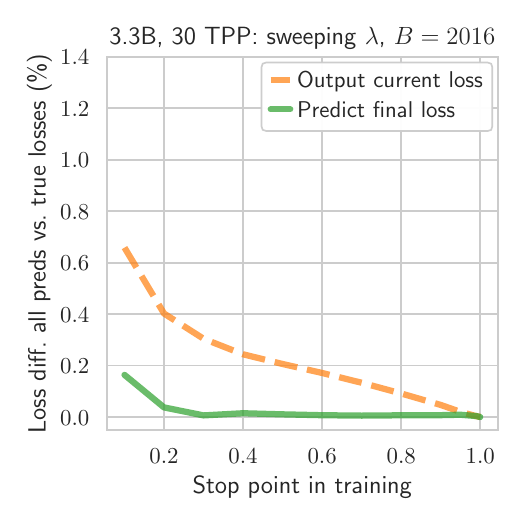}
    \end{minipage}
  \end{minipage}
  \caption{\textbf{Early stopping comparison: MAE at 1.7B, 3.3B:
      $\lambda$ sweeps.} Mean absolute error of all predicted final
    losses, comparing taking current loss vs.\ extrapolating final
    loss.\label{fig:hptuning4}}
\end{figure}

\begin{figure}[ht]
  \centering
  \begin{minipage}{0.66\textwidth}
    \begin{minipage}{0.49\textwidth}
      \includegraphics[trim={0.3cm 0.42cm 0.264cm 0.3cm}, clip, width=\linewidth]{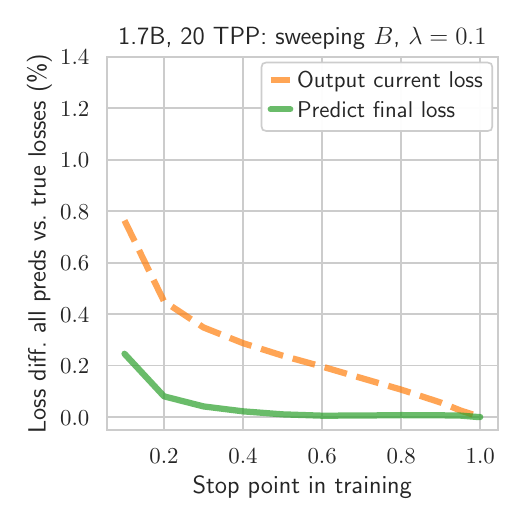}
    \end{minipage}\hfill
    \begin{minipage}{0.49\textwidth}
      \includegraphics[trim={0.3cm 0.42cm 0.264cm 0.3cm}, clip, width=\linewidth]{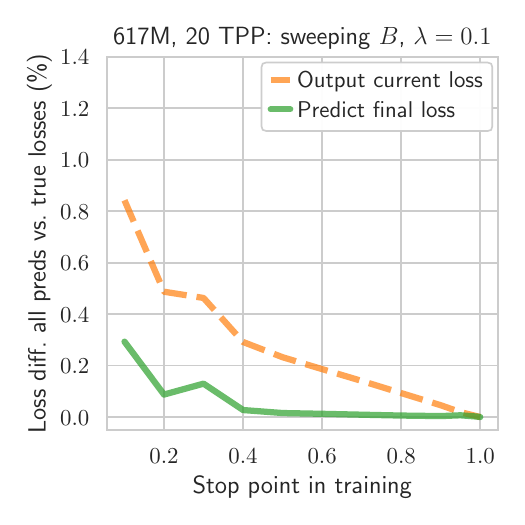}
    \end{minipage}
  \end{minipage}
  \caption{\textbf{Early stopping comparison: MAE at 1.7B, 617M: $B$
      sweeps.} Mean absolute error of all predicted final losses,
    comparing taking current loss vs.\ extrapolating final loss.
    \label{fig:hptuning5}}
\end{figure}

In many cases, rather than caring purely about which setting is best,
we care about the actual projected final loss.  This may be useful for
fitting scaling laws, or for helping practitioners reason about the
trade-offs of, for example, greater throughput from larger $B$
vs.\ suffering higher final loss.
We therefore evaluated the same four hyperparameter sweeps above, but
now evaluating the average loss difference between the predicted final
loss and the true final loss for all curves.  The baseline chooses the
current loss for each curve at the given training fraction, which will
overestimate the final loss.  Results in
\cref{fig:hptuning4,fig:hptuning5} show that in three of four cases,
extrapolating the final loss using our predictive form results in
\emph{much} smaller average error than using the current value.

The only instance where predicting the final loss incurred significant
error was the 1.7B, 20 TPP model with $B=8064$.  We note that the TLCs
are very noisy at this high batch size across almost all $\lambda$
settings and therefore it is evidently challenging to align the
in-progress training runs to the predicted TLC.\@ Increasing smoothing
reduces the predicted error somewhat, but the primary issue is that
the noise affects the TLC mainly in the first 60\% of training, thus
distorting even the smoothed loss from the universal trajectory.
Accurate prediction in the presence of loss spikes is an acknowledged
limitation of our methodology (\cref{sec:limitations}).

\subsection{Optimal and suboptimal TLCs as TPP scales}\label{sec:optimal}

\begin{figure}[ht]
  \centering
  \begin{minipage}{0.33\textwidth}
    \includegraphics[trim={0.3cm 0.42cm 0.264cm 0.3cm}, clip, width=\linewidth]{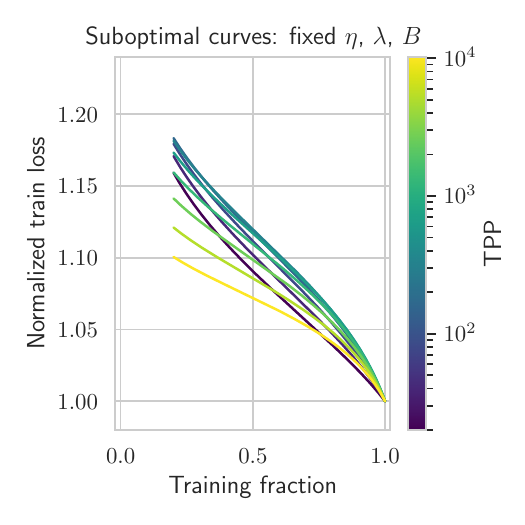}
  \end{minipage}\hfill
  \begin{minipage}{0.33\textwidth}
    \includegraphics[trim={0.3cm 0.42cm 0.264cm 0.3cm}, clip, width=\linewidth]{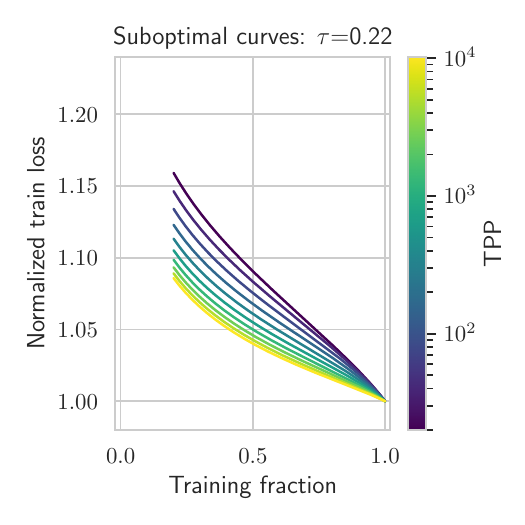}
  \end{minipage}\hfill
  \begin{minipage}{0.33\textwidth}
    \includegraphics[trim={0.3cm 0.42cm 0.264cm 0.3cm}, clip, width=\linewidth]{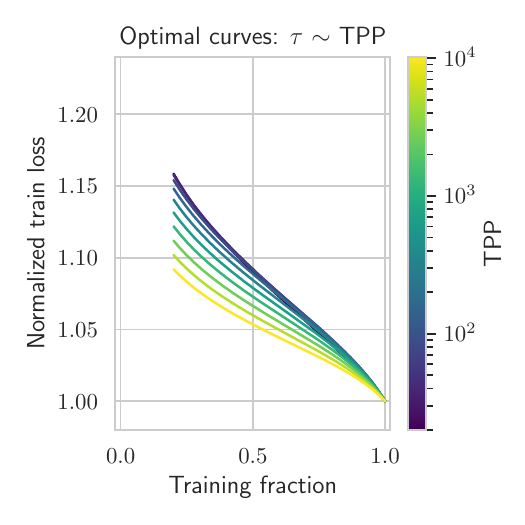}
  \end{minipage}
  \caption{\textbf{Evolution of train curve shape.}  $\Leftfig$: When
    TPP is scaled but $\eta$, $\lambda$ and $B$ are held constant,
    curve shape varies significantly.
    $\Middlefig$: When $\tema$ is instead held constant, shape evolves
    more gradually.
    $\Rightfig$: When $\tema$ scales with TPP according to established
    power laws, curves maintain their concave
    structure.\label{fig:optimalshape}}
\end{figure}

Given a fitted predictive form (\cref{eq:prediction}), it is natural
to ask how TLC shape varies as TPP increases, under various
hyperparameter (HP) scaling strategies.  In this section, we consider
three scenarios:
\begin{enumerate}
  \item No adjustment to any HPs: basically standard practice under
    $\mup$ until very recently.
  \item Maintain constant $\tema$: i.e., following the prescription
    of~\citet{wang2024how}.
  \item Optimize $\tema$: adjust $\tema$ for each TPP setting
    following the $\tema$ power law of \citet{bergsma2025power}.
\end{enumerate}

Results in \cref{fig:optimalshape} demonstrate that, with no HP
adjustments, curve shape changes substantially across TPP
($\leftfig$).\@ Fixing $\tau$ results in more consistent shapes
($\middlefig$), but only when $\tau$ is scaled for TPP do curves
maintain their characteristic concave shape, with a noticeable drop
near the end of training ($\rightfig$).  One may view this final
period as the \emph{annealing} phase of training, or the phase where
variance is reduced and we descend the valley into the
river~\citep{wen2024understanding}.  As TPP increases, we must reduce
$\tau$ correspondingly to prioritize exploration for the majority of
training, enabling this final descent only in the final phases.

\end{document}